\documentclass[10pt,twocolumn,letterpaper]{article}

\usepackage{iccv}
\usepackage{times}
\usepackage{epsfig}
\usepackage{graphicx}
\usepackage{amsmath}
\usepackage{amssymb}

\usepackage{subcaption}
\usepackage{wrapfig}
\usepackage{multirow}
\usepackage{cancel}
\usepackage[final]{changes} 
\usepackage[accsupp]{axessibility}

\usepackage[pagebackref=true,breaklinks=true,letterpaper=true,colorlinks,bookmarks=false]{hyperref}

\iccvfinalcopy 

\def\iccvPaperID{5174} 

\ificcvfinal\fi

\begin{document}

\title{Beyond Object Recognition: A New Benchmark towards \\Object Concept Learning}

\author{
Yong-Lu Li,~
Yue Xu,~
Xinyu Xu,~
Xiaohan Mao,~
Yuan Yao,~
Siqi Liu,~
Cewu Lu\thanks{Corresponding author.}\\
Shanghai Jiao Tong University\\
{\tt\footnotesize 
\{yonglu\_li, silicxuyue, xuxinyu2000, mxh1999, yaoyuan2000, magi-yunan, lucewu\}@sjtu.edu.cn}}


\twocolumn[{%
\renewcommand\twocolumn[1][]{#1}%
\maketitle
\begin{center}
    \centering
    \captionsetup{type=figure}
    \includegraphics[width=.7\textwidth]{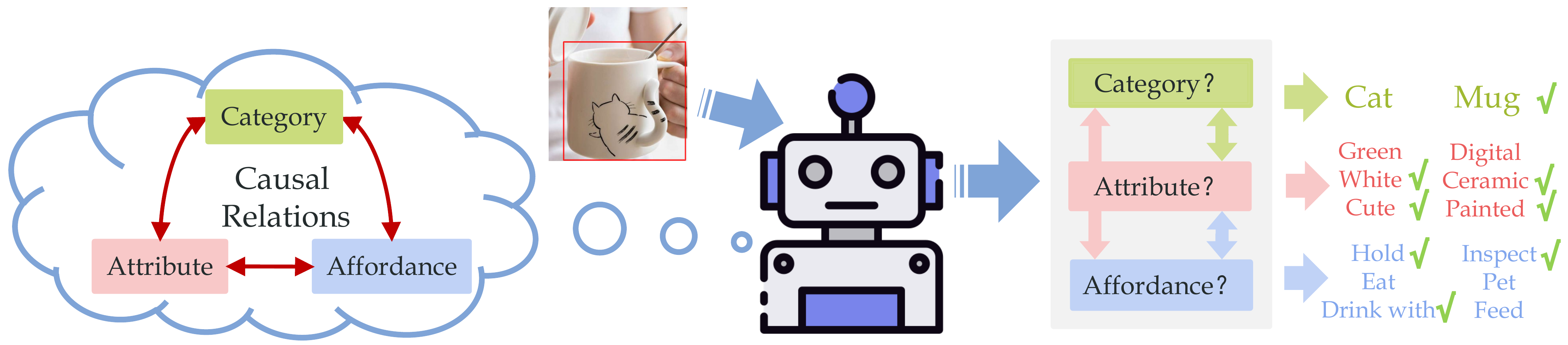}
    \captionof{figure}{For embodied agents, understanding daily objects requires the ability to perceive not only \textbf{category} but also \textbf{attribute} and \textbf{affordance}. In OCL, we try to reveal object concept learning in both three levels and explore their profound causal relations.}
    \label{fig:teaser}
\end{center}
}]
{
  \renewcommand{\thefootnote}%
    {\fnsymbol{footnote}}
  \footnotetext[1]{Corresponding author.}
}

\ificcvfinal\thispagestyle{empty}\fi

\begin{abstract}
    Understanding objects is a central building block of AI, especially for embodied AI. Even though object recognition excels with deep learning, current machines struggle to learn higher-level knowledge, \eg, what attributes an object has, and what we can do with it.
    Here, we propose a challenging \textbf{Object Concept Learning} (OCL) task to push the envelope of object understanding. It requires machines to reason out affordances and simultaneously give the reason: \textit{what attributes make an object possess these affordances}. 
    To support OCL, we build a \textit{densely} annotated knowledge base including extensive annotations for 
    three levels of object concept (category, attribute, affordance), and the clear causal relations of three levels.
    By analyzing the causal structure of OCL, we present a baseline, Object Concept Reasoning Network (OCRN). It leverages concept instantiation and causal intervention to infer the three levels. 
    In experiments, OCRN effectively infers the object knowledge while following the causalities well. 
    \textbf{Our data and code are available at \url{https://mvig-rhos.com/ocl}}.
\end{abstract}


\section{Introduction}
\label{sec:intro}

Object understanding is essential for intelligent robots. Recently, benefiting from deep learning and large-scale datasets~\cite{imagenet,coco}, category recognition~\cite{alexnet,faster} has made tremendous progress. But to close the gap between human and machine perception, machines need to pursue deeper understanding, \eg, recognizing higher-level attributes~\cite{mit} and affordances~\cite{gibson}, which may help it establish object concept~\cite{martin2007representation} when interacting with contexts.

Category \texttt{apple} is a symbol indicating its referent (real apples). In line with symbol grounding~\cite{harnad1990symbol}, machines should learn knowledge beyond category to approach concept understanding. 
According to cognition studies~\cite{memory-and-mind,martin2007representation}, attribute depicting objects from the physical/visual side plays an important role in object understanding.
Thus, many works~\cite{awa1,sun,apy} studied to ground objects with attributes, \eg, a \texttt{hammer} consists of a \texttt{long} handle and a \texttt{heavy} head.
Moreover, attributes can depict object states~\cite{mit}. 
An elegant characteristic of attributes is \textit{cross-category}: objects of the same category can have various states (\texttt{big} or \texttt{fresh apple}), whilst various objects can have the same state (\texttt{sliced orange} or \texttt{apple}). 
If the category is the \textbf{first} level of object concept, the attribute can be seen as the \textbf{second} level closer to the physical fact.

However, recognizing attributes is still far away from concept understanding. 
Given a \texttt{hammer}, we should know it can be \texttt{held} to \texttt{hit} nails, \ie, requiring machines to infer affordance~\cite{gibson} indicating what actions humans can perform with objects. Thus, we refer to affordance as the \textbf{third} level, which is closely related to common sense and causal inference~\cite{gibson}. Though affordance has been studied in robotics~\cite{affordancenet,hermans2011affordance} and vision~\cite{chuang2018learning,yuke} communities for decades, it is still challenging.
First, previous works~\cite{nguyen2017object,fouhey2015defense} often focus on recognizing affordance solely. But we usually infer affordance based on attribute observation.
If we need to knock in a nail without a \texttt{hammer} at hand, we may find other \texttt{hard} or \texttt{heavy} objects instead, \eg, a \texttt{thick} book. This profoundly reveals the \textbf{causality} between attribute and affordance. 
Second, previous works are designed for scale/scene-limited tasks, \eg, in \cite{yuke}, 40 objects and 14 affordances are included; Hermans~\etal.~\cite{hermans2011affordance} collect 375 indoor images of 6 objects, 21 attributes, and 7 affordances; a recent dataset~\cite{nguyen2017object} contains 10 indoor objects and 9 affordances. 
Thus, they cannot afford general affordance reasoning for large-scale applications.
 
To reshape object learning, we believe it is essential to look at the above three levels in a \textbf{unified} and \textbf{causal} way based on an extensive knowledge base.
Hence, we move a step forward 
to propose the object concept learning (OCL) task: given an object, machines need to infer its category, attributes, and further answer ``\textit{what can we do upon it and why}'', as shown in Fig.~\ref{fig:teaser}.
In a nutshell, machines need to reason affordance based on object appearance, category, and attributes.
To this end, we build a large-scale and dense dataset consisting of \textbf{381} categories, \textbf{114} attributes, and \textbf{170} affordances. It contains \textbf{80,463} images of diverse scenes and \textbf{185,941} instances in different states. 
Different from previous works~\cite{dengjia,hermans2011affordance,yuke}, OCL offers a more subtle angle. It includes:
(1) \textbf{category}-level attribute ($A$) and affordance ($B$) labels; 
(2) \textbf{instance}-level attribute ($\alpha$) and affordance ($\beta$) labels.
Besides, we annotate the \textit{causal relations} between three levels to evaluate the reasoning ability of models and keep the follow-up methods from fitting data only. 
Accordingly, based on the causal structure of OCL, we propose a \textit{neuro-causal} method, \textbf{O}bject \textbf{C}oncept \textbf{R}easoning \textbf{N}etwork (\textbf{OCRN}), as the future baseline.
It leverages concept instantiation (from category-level to instance-level) and causal intervention~\cite{pearl2016causal} to 
infer attributes and affordances.
OCRN outperforms a host of baselines and shows impressive performance while following the causal relations well.

In summary, our contributions are threefold:
\begin{enumerate}
\item Introducing the object concept learning task poses challenges and opportunities for object understanding and knowledge-based reasoning.
\item Building a benchmark consisting of diverse objects, elaborate attributes, and affordances, together with their clear causal relations.
\item An object concept reasoning network is introduced to reason three levels with concept instantiation performing well on OCL.
    
\end{enumerate}

\section{Related Work}
{\bf Object Attribute}
depicts the physical properties like color, size, shape, \textit{etc}. It usually plays the role of intermedia between pixels and higher-level concepts, \eg, prompting object recognition~\cite{apy}, affordance learning~\cite{hermans2011affordance}, zero-shot learning~\cite{awa1}, and object detection~\cite{kumar2018dock}. 
Recently, several large-scale datasets~\cite{apy,sun,imagenet150k,cocoattr,mit,visualgenome,hudson2019gqa} are released. 
For attribute recognition, besides direct attribute classification~\cite{awa1,relativeattr,sun,cocoattr} and leveraging the correlation between attribute-attribute and attribute-object~\cite{hwang2011sharing,analogous,mahajan2011joint}, intrinsic properties (compositionality, contextuality~\cite{redwine,operator}, symmetry~\cite{symnetcvpr,symnetpami}) of attribute-object are also proven useful.
\cite{redwine} uses the model weight space to encode the attributes to model the compositionality and contextuality.
\cite{operator} uses the attributes as linear operators to transform object embeddings. 
\cite{symnetcvpr} leverages the symmetry property to model the attribute changes within attribute-object coupling and decoupling. 

{\bf Object Affordance.} 
is introduced by \cite{gibson}. Affordance learning has two canonical paradigms: direct mapping~\cite{fouhey2015defense} or indirect method~\cite{yuke,zhao2013scene,wang2017binge,roy2016multi} with intermediates like object category, attribute, and 3D contents.
Some works learned affordance from human-object interactions (HOI) to encode the relation between object and action~\cite{gupta2007objects,yao2013discovering,kato2018compositional}.
Visual Genome~\cite{visualgenome} provides relations between objects, including actions instead of affordances. 
However, these relations cover limited and sparse affordances.
Differently, we use easily accessible object images as the knowledge source and densely annotate all attributes/affordances for all objects. 
Besides the vision community, the robot community pays much attention to affordance~\cite{pinto2016supersizing,thermos2017deep,pinto2016curious} for grasping and manipulation.
For instance, \cite{pinto2016supersizing} utilized the robot to discover the object affordance via self-supervised learning.
Recently, several datasets~\cite{nguyen2017object,dengjia,chuang2018learning} have been proposed.
IIT-AFF~\cite{nguyen2017object} collected ten daily indoor objects and provided nine common affordances to construct a dataset for robot applications.
Zhu~\etal.~\cite{yuke} built a dataset containing attribute, affordance, human pose, and HOI spatial configuration. 
But labeling pose~\cite{alphapose} and HOI~\cite{hicodet,pastanet} are costly.
Chao~\etal.~\cite{dengjia} proposed a \textit{semantic} category-level affordance dataset including 91 objects~\cite{coco} and 957 affordances.

{\bf Causal Inference.}
There is increasing literature on exploiting causal inference~\cite{pearl2016causal} in machine learning, especially with causal graphical models~\cite{spirtes2000causation,pearl2016causal}, including feature selection~\cite{guyon2007causal} and learning~\cite{chalupka2014visual}, video analysis~\cite{pickup2014seeing,lebeda2015exploring}, reinforcement learning~\cite{nair2019causal,dasgupta2019causal}, \etc.
Recently, Wang~\etal.~\cite{vcrcnn} studied the causal relation between objects in images and used intervention~\cite{pearl2016causal} to alleviate the observation bias.
Atzmon~\etal.~\cite{causal-czsl} analyze the causal generative model of compositional zero-shot learning and disentangle the representations of attributes and objects.
Here, we explore the causal relations between three object levels and apply backdoor adjustment~\cite{pearl2016causal} to alleviate the existing bias. 

\begin{figure*}
    \centering
    \includegraphics[width=\linewidth]{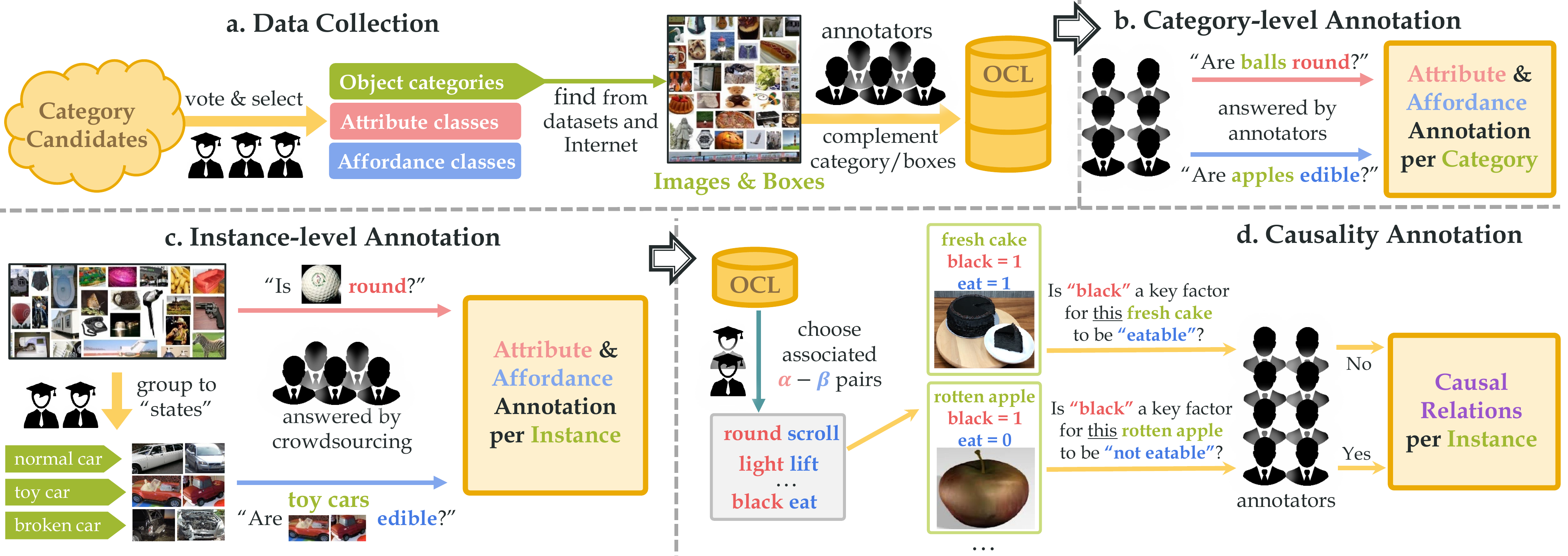}
    \caption{OCL construction. a) Data collection. b) Annotating category-level attributes and affordances. c) Annotating instance-level attributes and affordances. d) Finding direct and clear instance-level causal relations.}
    \label{fig:labelling}
\end{figure*}

\section{Constructing OCL Benchmark} 
\label{sec:annotation}
We construct a benchmark to characterize abundant object knowledge following Fig.~\ref{fig:labelling}.

\subsection{Fine-Grained Object Knowledge Base}

{\bf Data Collection.}
We briefly introduce the collection of affordances, categories, attribute classes, and image sources here.
\begin{enumerate}
 \item \textbf{Affordance}: 
We collect 170 affordances out of 1,006 candidates from widely-used action/affordance datasets~\cite{dengjia,AVA,hicodet,vcoco,yuke,nguyen2017object} given generality and commonness.

 \item \textbf{Category}: 
Considering the taxonomy (WordNet~\cite{wordnet}) and diversity, we collect 381 objects out of 1,742 candidates from object datasets~\cite{apy,sun,coco,cocoattr,imagenet,imagenet150k}.

 \item \textbf{Attribute}: 
We manually filter the 500 most frequent attributes from attribute datasets~\cite{apy,sun,coco,cocoattr,imagenet,imagenet150k,visualgenome} and choose 114 attributes, covering colors, deformations, supercategories, surface, geometrical and physical properties. 

 \item \textbf{Image}: 
We extract 75,578 images from object datasets~\cite{apy,sun,coco,cocoattr,imagenet,imagenet150k,visualgenome}, together with Ground Truth (GT) boxes.
We manually collected 4,885 Internet images of selected categories to ensure diversity. 
Then, we annotate the missing box and category labels for all instances.
Finally, \textbf{185,941} instances of \textbf{381} categories from \textbf{80,463} images are collected: an average of 488 instances per category and 2.31 boxes per image. 
Details are given in the supplementary.
OCL is long-tail distributed, where the head categories have over 5,000 instances each, but the rarest categories have only 9 instances, which challenges the robustness of machines greatly.

\end{enumerate}

\begin{table}[t]
\begin{center}
    \resizebox{\linewidth}{!}{
    \begin{tabular}{lcccccc}
	\hline  
    Dataset & \# Image & \# Instance & \# Object & \# Attribute & \# Affordance\\
    \hline 
    APY~\cite{apy} & 15,339 & 15,339 & 32 & 64 & / \\
    SUN~\cite{sun} & 14,340 & 14,340 & 717 & 102 & / \\
    COCO-a~\cite{cocoattr} & 84,044 & 188,426 & 29 & 196 & / \\
    ImageNet150k~\cite{imagenet150k} & 150,000 & 150,000 & 1,000 & 25 & /\\
    Chao~\etal.~\cite{dengjia} &/ & /& 91 & / & 957 ($B$) \\
    Hermans~et.al.~\cite{hermans2011affordance} & 375 & - & 6 & 21 & 7 \\
    Zhu~\etal.~\cite{yuke} & 4,000 & 4,000 & 40 & 57 & 14 \\
    \hline
    OCL & 80,463 & 185,941 & 381 & 114 & 170\\
    \hline
	\end{tabular}
	}
\end{center}
\caption{\textit{Dense annotated} datasets. OCL provides category- and instance- level attributes ($A$, $\alpha$), affordances ($B$, $\beta$).} 
\label{table:OKB-stat}
\end{table}

{\bf Annotating Attribute} in two levels of granularity: 
(1) \textbf{Category-level} attribute ($A$) contains common sense.  
For each category, we annotate its \textit{most common} attributes. 
In concept learning, the usage of the category-level labels as common knowledge can date back to \cite{osherson1991default}. 
Following \cite{osherson1991default}, to avoid bias, annotators are given \textit{category-attribute pairs} (category names instead of images) and multiple annotators vote to build the binary $A$
matrix $M_{A}$ in size of $[381,114]$.
(2) \textbf{Instance-level} attribute ($\alpha$) is the individual attributes of \textit{each instance}. 
The annotation unit is an \textit{attribute-instance pair} and each pair is labeled by multiple annotators.

\begin{figure*}[t]
    \centering
    \includegraphics[width=0.8\linewidth]{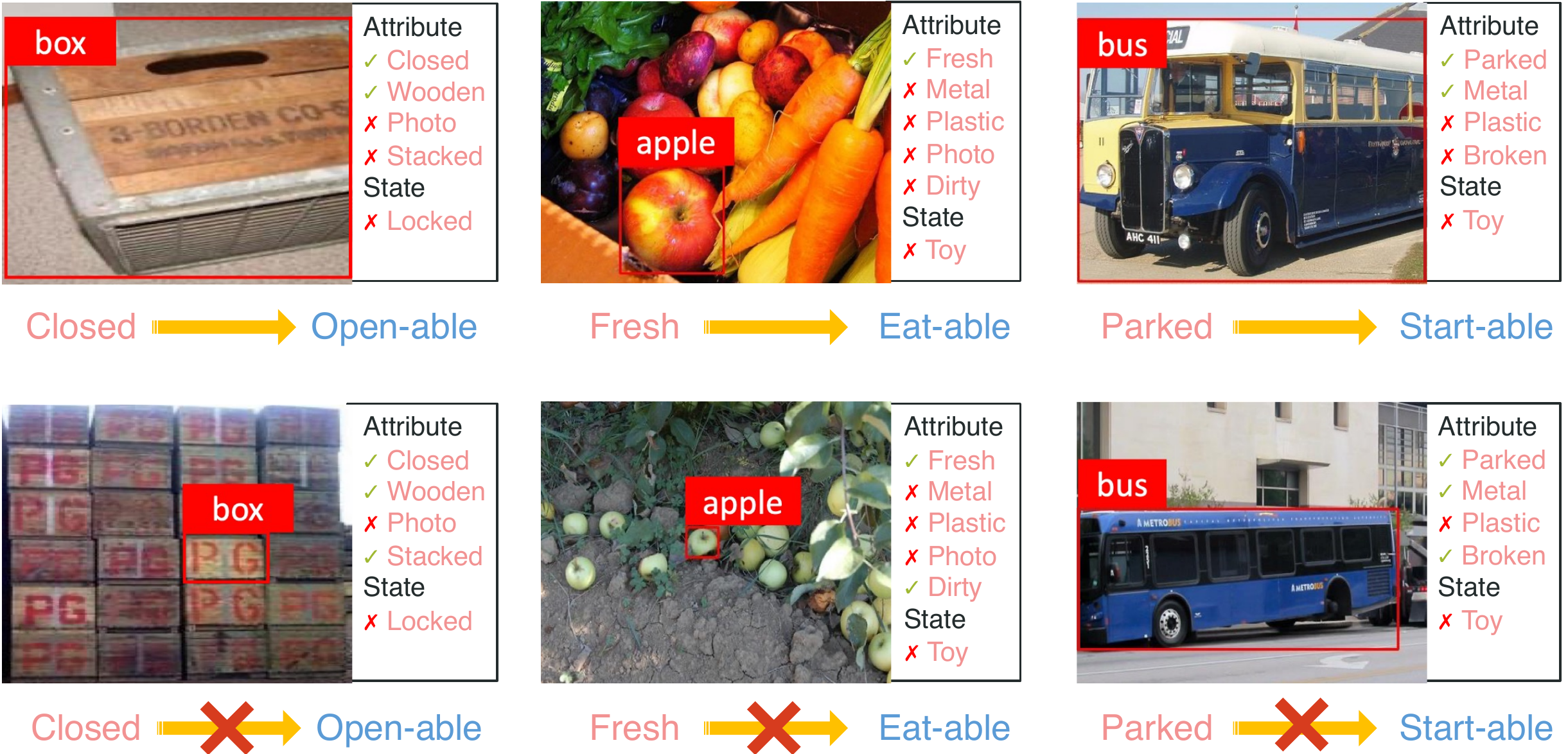}
    \caption{OCL samples including category, $\alpha$ (red), $\beta$ (blue), and their causal relations in various contexts.}
    \label{Figure:data-sample}
\end{figure*}

{\bf Annotating Affordance} in two levels of granularity:
(1) \textbf{Category-level} affordance $B$, similar to $A$, is annotated in \textit{category-affordance pairs}, indicating the common affordances of each category.
Following \cite{dengjia}, the annotators label $B$ matrix $M_{B}$ in size of $[381,170]$.
(2) \textbf{Instance-level} affordance $\beta$ is annotated for \textit{each instance} with the help of object \textit{state}. 
As $B$ is defined by common states, objects in specific states may have different affordances from $B$: 
if a service robot finds a \texttt{broken cup}, it may infer that the \texttt{cup} can still hold water as it is trained with $B$ labels. Thus, we need detailed $\beta$ beyond $B$. 
$\beta$ exhibits evident similarities for objects in similar status forming ``state'' aligning with commonsense, thus we use them to streamline annotation and reduce the annotator discrepancy.
A state is defined as an [\texttt{category, description} (\eg, a set of attributes)] pair, and instances in a state usually possess similar affordances, \eg, \texttt{fresh, juicy, clean oranges} are \texttt{eatable}.
First, six experts conclude the states by scanning \textit{all} instances of a category and listing all states according to affordance. Then these states were merged manually.
In total, \textbf{1,376} states are defined, and each category has 3.6 states on average. 
Next, $\beta$ is annotated for \textit{each state}, and the instances are first assigned with the state-level $\beta$.  
Bext, the instance-level $\beta$ is \textbf{detailed} based on the state-level $\beta$ according to the visual content of each instance.
Note that the state is category-dependent and can not be transferred among object categories, which is different from attribute and affordance. 
Besides, the composition of attributes makes the state space huge and there can be many \textit{unseen} states.
Thus, we only use them in annotation but not in our method.

Fig.~\ref{Figure:data-sample} shows some examples of OCL.
We compare OCL with previous dense datasets in Tab.~\ref{table:OKB-stat}. 
More details, figures, and tables are given in the supplementary.

\subsection{Causal Graph Definition} 
\begin{figure}[t]
\begin{center}
    \includegraphics[width=\linewidth]{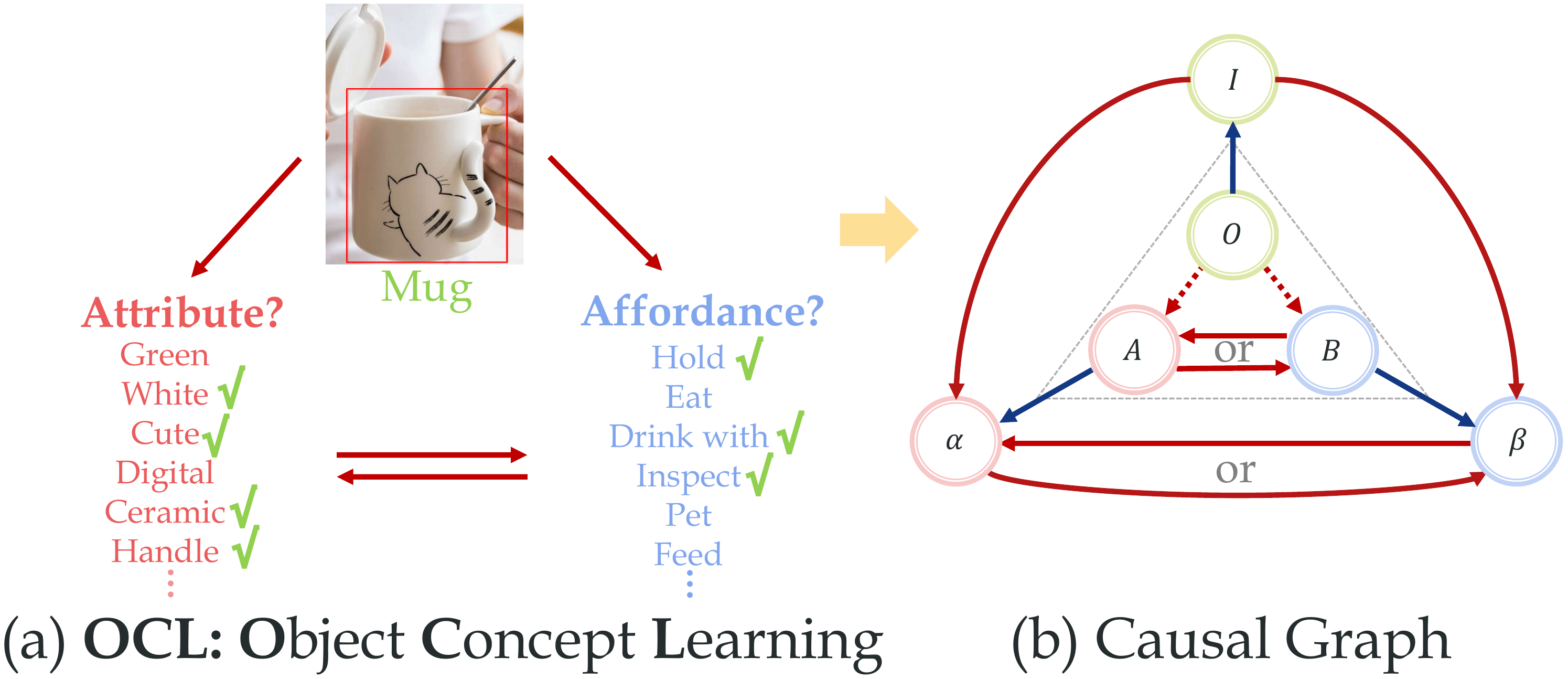} 
    \caption{Causal graph of our OCL task. ``or'' indicates that either $A \leftarrow B$ or $A \rightarrow B$ ($\alpha \rightarrow \beta$ or $\alpha \leftarrow \beta$) exists.}
    \label{fig:OCL-causal-graph}
\end{center}
\end{figure}

We use a causal graph to shed light on the subtle causalities of our knowledge base in Fig.~\ref{fig:OCL-causal-graph}. Causal graph~\cite{pearl2016causal} indicates the underlying causalities based on components:

\begin{itemize}
\item $O$: object category
\item $I$: object instance in an image
\item $A$: category-level attribute
\item $\alpha$: instance-level attribute
\item $B$: category-level affordance
\item $\beta$: instance-level affordance
\end{itemize}

According to the prior knowledge about the causalities between three levels, a hierarchical structure is depicted: 
\textbf{(a)} the \textbf{inner} triangle with dotted lines is the \textbf{category}-level: object category $O$, category-level attributes $A$, and affordances $B$;
\textbf{(b)} the \textbf{outer} triangle is the \textbf{instance}-level: instance visual appearance $I$, instance-level attributes $\alpha$, and affordances $\beta$. 
Each directed \textit{possible} arc in the graph indicates the \textit{possible} causality between two nodes.

Here, besides the red arcs indicating the common causal relations (\eg, $I \rightarrow \alpha$, $I \rightarrow \beta$ as attribute/affordance recognition from images), we define some special arcs given our category-level attribute and affordance settings:
(1) $O \rightarrow A$, $O \rightarrow B$ (dotted arcs): Given $O$, $A, B$ are strictly \textit{determined} within labels.
(2) $O \rightarrow I$, $A \rightarrow \alpha$, $B \rightarrow \beta$ (blue arcs): The category-level $O$, $A$, and $B$ are direct causes of instance-level $I$, $\alpha$, and $\beta$ during the concept \textit{instantiation}.
Note that, according to the previous analysis, we focus on the $A \rightarrow B$ and $\alpha \rightarrow \beta$ but sometimes the opposite can also happen:
$A \leftarrow B$ and $\alpha \leftarrow \beta$ (``or'' in Fig.~\ref{fig:OCL-causal-graph}).
\replaced{In annotation and experiments, we observe that $\alpha\rightarrow\beta$ is stronger and more common and natural to human perception, so we focus more on $\alpha\rightarrow\beta$ in our causal benchmark (Sec.~\ref{sec:task-overview}).}

In this work, we focus on $\alpha, \beta$ perception ($I \rightarrow \alpha$, $I \rightarrow \beta$) and visual reasoning (with $I$, inferring $\beta$ given $\alpha$) for embodied AI.
Thus, Fig.~\ref{fig:OCL-causal-graph} is simplified. 
Our knowledge base can support more tasks such as attribute/affordance conditioned image generation ($\alpha \rightarrow I$, $\beta \rightarrow I$)~\cite{stabledifussion}. However, they are beyond the scope of this paper (Suppl.~Sec.~\textcolor{red}{3}).

\subsection{Causal Inference Benchmark on $\alpha\rightarrow\beta$}
\label{sec:causal_labeling}
We annotate \textit{instance}-level (considering the context of each instance) causality of $\alpha\rightarrow\beta$ to answer ``\textit{which attribute(s) are the critical and direct causes of a certain affordance?}'' in two phrases:

\textbf{Filtering:}
Initially, we need to make binary decisions on all \textit{instance}-$\alpha$-$\beta$ triplets, which is far beyond handleable. 
Fortunately, we find that \textbf{most} $\alpha$-$\beta$ classes (\eg, \texttt{shiny} and \texttt{kick}) are meaningless and always of no causality.
Thus, we exclude the most impossible pairs and only annotate existing rules without ambiguity, meanwhile, guaranteeing the completeness of causality.
For each of the 114×170 $\alpha$-$\beta$ pairs, we attach 10 samples for reference and 3 experts vote \textit{yes/no/not-sure}. We take the majority vote and the \textit{not-sure} and controversial pairs are rechecked. The \textit{not-sure} and \textit{no} pairs are removed, and so do the ambiguous pairs.
Finally, we obtain about 10\% $\alpha$-$\beta$ classes as candidates. The left 90\% pairs may hold value, we plan to use LLMs to mine new rules in future work, especially from ambiguous pairs.

\textbf{Instance-level Causality}: we adopt object states as a reference.
Multiple annotators have been involved for each \textit{state}-$\alpha$-$\beta$ triplet and are asked whether the specific attribute is the \textit{clear} and \textit{direct} cause of this affordance in this state. The answers are combined and checked for all instances of a state.
Finally, we obtain about 2 M \textit{instance-$\alpha$-$\beta$} triplets of causal relations.
As we have labeled all $\alpha$ and $\beta$ for all instances, the causal relations would be in four situations: [0,0], [1,1]; [0,1], [1,0]. 
The former two are ``positive'', \eg, \texttt{fresh}(1/0)$\rightarrow$\texttt{eat}(1/0) for an \texttt{apple}. While the last two are ``negative'', \eg, \texttt{broken}(1/0)$\rightarrow$\texttt{drive}(0/1) for a \textit{car}.

Fig.~\ref{Figure:data-sample} shows some causal examples.
These causalities are not thoroughly studied in previous datasets~\cite{yuke,nguyen2017object,fouhey2015defense,hermans2011affordance}. 
For more details, please refer to the supplementary.

\subsection{Task Overview} 
\label{sec:task-overview}
Here, we formulate the OCL task formally.
Given an instance $I$ (content in box $b_o$ representing an object instance), OCL aims to infer attribute $\alpha$ and affordance $\beta$ while following the causalities. 
Formally, OCL can be described as:
\begin{eqnarray}
\label{eq:OCL}
    <P_{\alpha}, P_{\beta}>=\mathcal{F}(I, P(O|I)),
\end{eqnarray}
where $P_{\alpha}, P_{\beta}$ are the probabilities of $\alpha, \beta$, $P(O|I)$ is the predicted category probability  from an object detector~\cite{faster}.

We aim at benchmarking the reasoning ability of machines, causal relations in Fig.~\ref{fig:OCL-causal-graph} can all be candidates.
However, annotating causal relations is usually ambiguous and it is impractical to cover all relations. In a user study, experts met significant divergence when annotating different arcs.
For embodied AI, affordance $\beta$ is more important in robot-world interactions.
Moreover, both the causal relation annotation and the ablations support that the causal effect of ${\alpha}\rightarrow{\beta}$ is more significant than the other alternatives.
Thus, we only annotate the unambiguous $\alpha\rightarrow\beta$ (Sec.~\ref{sec:causal_labeling}) and mainly measure the learning of $\alpha\rightarrow\beta$ here. 
Formally, the evaluation of $\alpha\rightarrow\beta$ learning follows
\begin{eqnarray}
\label{eq:OCL-ITE}
    \Delta P_{\beta} = ITE[\mathcal{F}(I, P(O|I))],
\end{eqnarray}
where $\Delta P_{\beta}$ is the Individual Treatment Effect~\cite{rubin2005causal} of \textbf{affordance prediction change} after we operate $ITE[\cdot]$ on a model $\mathcal{F}(\cdot)$. $\Delta P_{\beta}$ is expected to follow the GT causal relation between $\alpha, \beta$ from humans.
For example, when the attributes of an object change, then the causal-related affordances should also change accordingly.
We will detail the ITE evaluation in Sec.~\ref{sec:experiment}.
Note that $A, B$ are decided by $O$. Given $O$, we can get $A, B$ via querying the prior $M_{A}, M_{B}$ (Sec.~\ref{sec:annotation}). Thus, we do not evaluate $A \rightarrow B$ here.

We split images into the train, validation, and test sets with 56K:14K:9K images.
The validation and test sets cover 221 of the 381 categories, and the train set covers all categories. 
OCL is a long-tailed recognition task~\cite{gupta2019lvis,xu2022constructing} and requires generalization to cover the whole object category-attribute-affordance space with imbalanced information. 
Thus, it is challenging for current machines without the reasoning ability to understand the causalities.

\section{Object Concept Reasoning Network}
\label{sec:OCRN}
\begin{figure*}[!t]
    \includegraphics[width=0.99\linewidth]{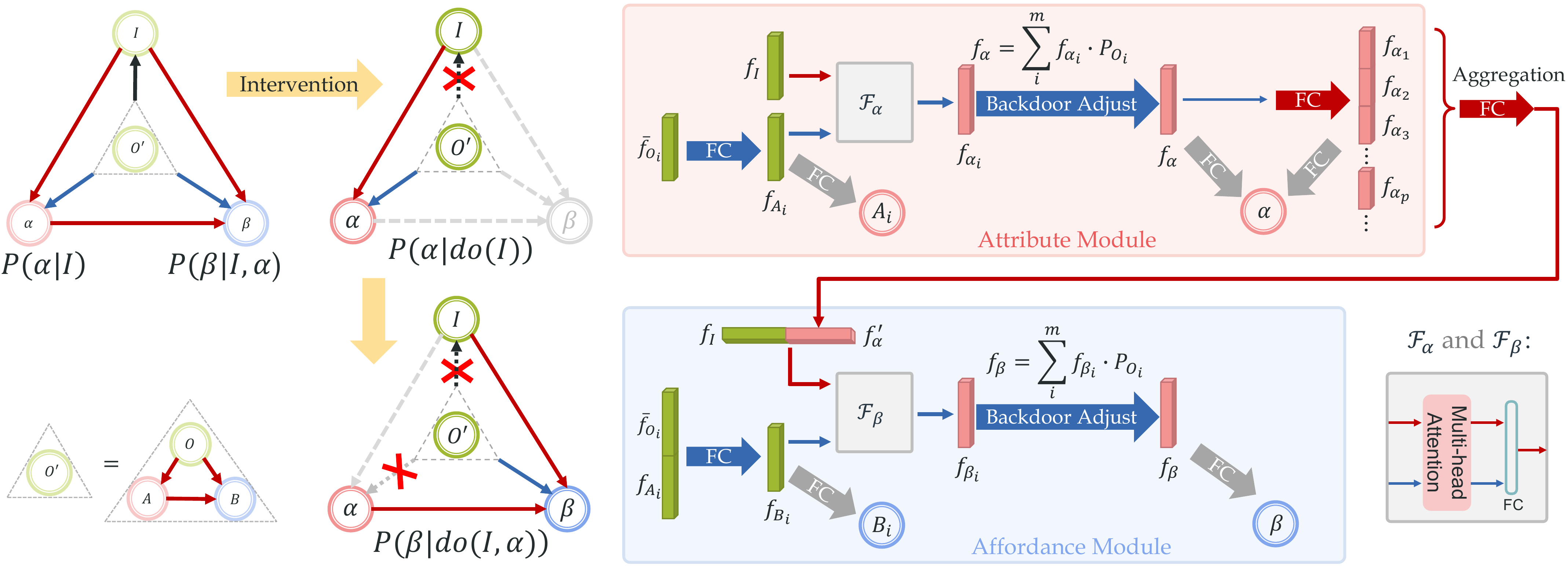} 
	\centering
	\caption{OCRN overview. 
	The arc from $O$ to $I$ is deconfounded. Thus, we can eliminate the bias from the $O$ imbalance. Equations below the graphs are the original or deconfounded estimations of $\alpha, \beta$. 
	Attribute and affordance modules are the \textbf{instantiations} of category-level features: categorical features $f_{A_i}$ or $f_{B_i}$ are obtained following the left-bottom-most causal graph and then instantiated via $\mathcal{F}_{\alpha}$ or $\mathcal{F}_{\beta}$ conditioned by the instance representations.
	$f_{\alpha}$ and $f_{\beta}$ after intervention are the expectations of instantiated $f_{\alpha_i}$ and $f_{\beta_i}$ w.r.t \textbf{prior} $P_{O_i}$. At last, linear-Sigmoid classifiers give the final predictions.} 
	\label{fig:graph-do}
\end{figure*}

Before proposing the OCRN, we first simplify the causal graph in Fig.~\ref{fig:OCL-causal-graph} to facilitate the implementation. 
We focus on $\alpha \rightarrow \beta$ and omit $\beta \rightarrow \alpha$. Similarly, we omit $B \rightarrow A$.
Besides, {$I, \alpha, \beta$} are the \textit{instantiations} of {$O, A, B$} respectively and we use a $O'$ node to represent {$O, A, B$}.
The adapted causal graph is shown in Fig.~\ref{fig:graph-do}.
OCRN implements the \textbf{instantiation} of attribute and affordance, corresponding to $A \rightarrow \alpha$, $B \rightarrow \beta$. Thus the model can propose a coarse estimation of attribute and affordance at category-level, then tune the results with the image patterns as a condition for a more accurate prediction.
Besides, we exploit \textbf{intervention} to remove the causal relation between $I$ and $O$ to construct a category-agnostic model. It suffers less from category bias and is more capable of learning uncommon cases.

{\bf Object Category Bias.} 
OCL can be depicted as $P(\alpha|I)$ and $P(\beta|I,\alpha)$. 
As the samples of different categories are usually imbalanced, conventional methods may suffer from severe \textit{category bias}~\cite{vcrcnn}, \eg, animal accounts for 22\% instances in OCL, and home appliance only accounts for 3\%. 
In $P(\alpha|I)$, category bias is imported following 
\begin{equation}
    \begin{aligned}
    \label{eq:attr-bias}
        P(\alpha|I)=\sum^m_{i} P(\alpha|I,O_i)P(O_i|I),
    \end{aligned}
\end{equation}
where $P(O_i|I)$ is the predicted category probability. 
That is, $O$ is a confounder~\cite{pearl2016causal} and pollutes attribute inference, especially for the \textit{rare} categories.

{\bf Causal Intervention.}
To tackle this, we propose OCRN using intervention~\cite{pearl2016causal} to deconfound the confounder $O$ for $\alpha$ (Fig.~\ref{fig:graph-do}).
In $\alpha$ estimation, we use $do(\cdot)$ operation~\cite{pearl2016causal} to eliminate the arc from $O$ to $I$: $P(\alpha|do(I))$ is
\begin{equation}
    \begin{aligned}
    \label{eq:iat-do}
        &\sum_{i}^m P(\alpha|I,O_i)P(O_i)
        \\=&\sum_{i}^m P(O_i)\sum_{j}^m P(\alpha|I,A_j)P(A_j|O_i)
        \\=&\sum_{i}^m P(\alpha|I,A_i)P(O_i),
    \end{aligned}
\end{equation}
where $m=381$. 
$A_j$ is the category-attribute vector of $j^{th}$ category.
As $A$ is decided by $O$, $P(A_j|O_i)=1$ if $i=j$ and $P(A_j|O_i)=0$ if $i \neq j$, 
where $O_i$ is the $i^{th}$ category and $A_j$ is the category-attribute of $j^{th}$ category.
$P(O_i)$ is the \textbf{prior} probability of the $i$-th category (frequency in our train set).
We apply the intervention to reduce the bias from $O$ recognition for an \textbf{category-agnostic} model.

Similar to $\alpha$, in $\beta$ estimation, category bias also exists:
\begin{equation}
    \begin{aligned}
    \label{eq:aff-bias}
        P(\beta|I,\alpha)=\sum_{i}^m P(\beta|I,\alpha,O_i)P(O_i|I,\alpha).
    \end{aligned}
\end{equation}
With Eq.~\ref{eq:iat-do}, $\alpha$ is beforehand estimated and thus can be seen as ``enforced'' and deconfounded.
For $I$, we again use the intervention~\cite{pearl2016causal}:
\begin{equation}
    \begin{aligned}
    \label{eq:iaf-do}
        P(\beta|do(I,\alpha))=\sum_{i}^m P(\beta|I,\alpha,B_i)P(O_i).
    \end{aligned}
\end{equation}
Similar to Eq.~\ref{eq:iat-do}, $P(B_j|O_i)=1$ if $i=j$, $P(B_j|O_i)=0$ if $i \neq j$, we omit the process for clarity.

\subsection{Model Implementation}
We represent nodes $\{I, A, B, \alpha, \beta\}$ as $\{f_I$, $f_{A}$, $f_{B}$, $f_{\alpha}$, $f_{\beta}\}$ respectively in latent space.
$f_I$ is the RoI pooling feature of an instance extracted by a COCO pre-trained ResNet-50~\cite{resnet}. 
Following Eq.~\ref{eq:iat-do}, we represent category-level attribute $A$ based on the \textit{mean} object category feature $\bar f_{O_i}$, which is the mean of $f_I$ of all \textbf{training} samples in category $O_i$.
We map $\bar f_{O_i}$ to the attribute latent space $f_{A_i}$ with fully-connected layers (FC) (Fig.~\ref{fig:graph-do}).
$f_{A_i}$ stands for the category-attribute representation for $\text{i}^{\text{th}}$ category.

{\bf Attribute Instantiation.}
Next, we obtain $\alpha$ representation following Eq.~\ref{eq:iat-do}: 
\begin{eqnarray}
\label{eq:iat}
    f_{\alpha_i}=\mathcal{F}_{\alpha}(f_I, f_{A_i}),\quad f_{\alpha}=\sum_{i}^m f_{\alpha_i}\cdot P_{O_i},
\end{eqnarray}
where $P_{O_i}$ is the \textit{prior} category probability ($P(O_i)$ in Eq.~\ref{eq:iat-do}). 
Eq.~\ref{eq:iat} indicates the attribute \textit{instantiation} from $A$ to $\alpha$ with $I$ as the \textit{condition}. Hence, we can equally translate the $\alpha$ estimation problem into a \textbf{conditioned instantiation problem}.
$\mathcal{F}_{\alpha}(\cdot)$ is implemented with multi-head attention~\cite{attention} with two entries (Fig.~\ref{fig:graph-do}). 
The attention output is compressed by a linear layer to the instantiated representation $f_{\alpha_i}$.
The debiased representation $f_{\alpha}$ is the expectation of $f_{\alpha_i}$ w.r.t $P_{O_i}$ according to back-door adjustment in Eq.~\ref{eq:iat-do}.

We also get the feature for specific attributes for ITE operation (Sec.~\ref{sec:experiment}).
$f_{\alpha}$ is first separated to $f_{\alpha_p}$ for each attribute $p$ ($p\in[1,114]$) by multiple independent FCs, then we can manipulate specific attributes by masking some certain $f_{\alpha_p}$. 
Next, the features are aggregated via concatenating-compressing by an FC to $f'_{\alpha}$ as shown in Fig.~\ref{fig:graph-do}.

{\bf Affordance Instantiation.}
Similarly, FCs are used to obtain $f_{B}$ from $\bar f_{O_i}$ and $f_{A_i}$ and Eq.~\ref{eq:iaf-do} is implemented as:
\begin{eqnarray}
    \label{eq:caf-iaf}
    f_{\beta_i}=\mathcal{F}_{\beta}(f_I, f'_{\alpha}, f_{B_i}), \quad f_{\beta}=\sum_{i}^m f_{\beta_i}\cdot P_{O_i}.
\end{eqnarray}
$\mathcal{F}_{\beta}(\cdot)$ operates instantiation with conditions $\{f_I, f'_{\alpha}, f_{B_i}\}$.

\subsection{Learning Objectives.} 
To drive the learning, we devise several objectives:

\textbf{Category-level loss $L_C$}. 
We input category-level $f_{A}, f_{B}$ to two linear-Sigmoid classifiers to classify $A, B$. The binary cross-entropy losses are $L_{A}$ and $L_{B}$.
The total category-level loss is $L_C = L_{A} + L_{B}$.

\textbf{Instance-level loss $L_I$}.
We input instance-level $f_{\alpha}$, $f_{\beta}$, together with $f_{\alpha_i}$, $f_{\beta_i}$ to linear-Sigmoid classifiers. The separated $f_{\alpha_p}$ are also sent to independent binary classifiers.
The binary cross-entropy losses are represented as $L_{\alpha}, L_{\beta}$.
The total instance-level loss is $L_I = L_{\alpha} + L_{\beta}$.

The total loss is $L=\lambda_C L_C + L_I$.
We adopt a two-stage policy: first inferring attributes, then reasoning affordances.

\section{Experiment}
\label{sec:experiment}

\subsection{Metrics}

$\alpha, \beta$ \textbf{Recognition}: we measure the correctness of model prediction $\hat\alpha$ and $\hat\beta$. For multi-label classification tasks, we use the mean Average Precision (mAP) metric.

\textbf{Reasoning}: we use \textbf{Individual Treatment Effect (ITE)}~\cite{rubin2005causal}. $ITE_i=Y_{i,T=1}-Y_{i,T=0}$ measures the causal effect $T\rightarrow Y$ of $\text{i}^{\text{th}}$ individual with the difference between outcomes ($Y$) with or without receiving the treatment ($T$).
In OCL, we discuss the causal relation between $\text{p}^\text{th}$ attribute and $\text{q}^\text{th}$ affordance: $\alpha_p\rightarrow\beta_q$. 
So we interpret the treatment $T$ as the \textbf{existence of $\alpha_q$} and the outcome $Y$ as the $\beta_q$ output. 
We measure the difference of $\beta_q$ output when the whole $\alpha_q$ feature is wiped out or not, which should be non-zero when the causal relation $\alpha_p\rightarrow\beta_q$ exists.

\begin{figure}[t]
\begin{center}
    \includegraphics[width=\linewidth]{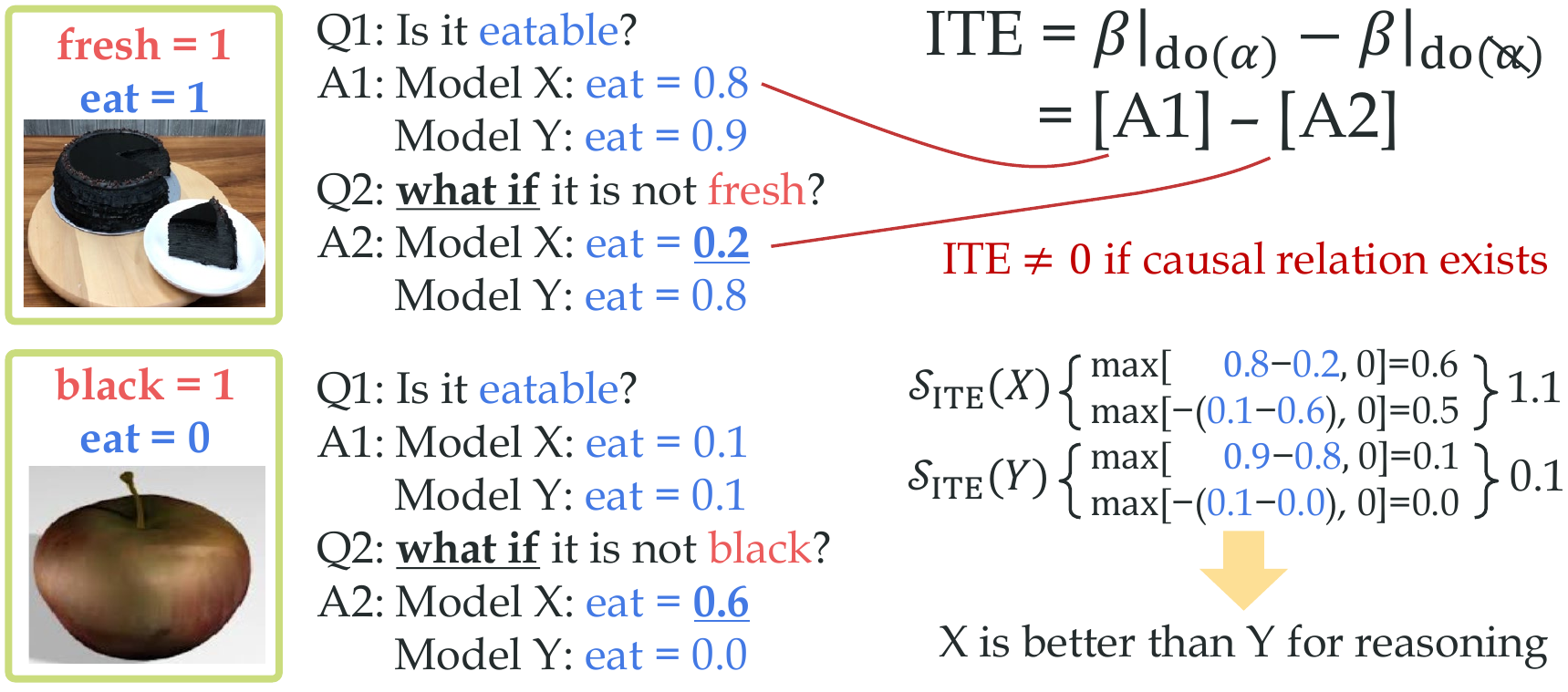}
    \caption{Example of ITE reasoning benchmark.} 
    \label{fig:ite-intro}
\end{center}
\end{figure}

In detail, given a model, for an instance with causal relation $\alpha_p\rightarrow\beta_q$ ($p\in[1,114], q\in[1,170]$), we first formulate ITE as the \textbf{affordance probability change} following Eq.~\ref{eq:OCL-ITE}: 
\begin{equation}
    ITE=\Delta\hat{\beta_q}=\hat{\beta_q}|_{do(\alpha_p)}-\hat{\beta_q}|_{do(\bcancel{\alpha_p})}.
\end{equation}
$\hat{\beta_q}|_{do(\alpha_p)}$ is the factual output of the affordance probability. $\hat{\beta_q}|_{do(\bcancel{\alpha_p})}$ is the counterfactual output when the $\alpha_p$ is wiped out, which can be got by assign zero-mask~\cite{tang2020unbiased} to the feature of $\alpha_p$ (\eg, $f_{\alpha_p}$ in OCRN) and keep the other features.

Then, based on ITE, we benchmark instances following:

\textbf{ITE}:
If the causality $\alpha_p\rightarrow\beta_q$ exists on the instance, ITE should be non-zero when eliminating the effect of $\alpha_p$. And the direction of ITE depends on the affordance ground-truth $\beta_q$: if $\beta_q=0$, the predicted $\hat{\beta_q}$ tend to be $1$ after wiping out $\alpha_p$ so ITE should be a negative value; contrarily, ITE should be positive if $\beta_q=1$.
Hence we compute the ITE score as:
\begin{equation}
    \begin{aligned}
        \mathcal{S}_\text{ITE}=\left\{
            \begin{aligned}
                & \max(\Delta\hat{\beta_q},~0), & \beta_q=1, \\
                & \max(-\Delta\hat{\beta_q},~0), & \beta_q=0,
    \end{aligned}
    \right.
    \end{aligned}
\end{equation}
so that larger $\mathcal{S}_\text{ITE}$ indicates the model infers more accurate ITE directions and has better reasoning performance. An example is given in Fig.~\ref{fig:ite-intro}.

\textbf{$\alpha$-$\beta$-ITE}:
we combine recognition and reasoning performances. We multiply $\mathcal{S}_\text{ITE}$ with $P(\hat{\alpha_p}=\alpha_p)$ and $P(\hat{\beta_q}=\beta_q)$ as a unified metric $\mathcal{S}_{\alpha\text{-}\beta\text{-ITE}}$. 

For all metrics, we compute AP for each $[\alpha_p,\beta_q]$ and average them to mAP. Non-existing pairs are not considered.

\subsection{Baselines}
Different methods exploit different causal paths including the sub-graphs with $\alpha\rightarrow\beta$ or $\alpha\leftarrow\beta$ based on Fig.~\ref{fig:OCL-causal-graph}. 
We implement a series of baselines following different sub-graphs to fully exert the potential of OCL and divide them into 3 folds w.r.t. $\alpha-\beta$ causal structure.
We briefly list them here and detail them in the supplementary:

{\bf Fold \uppercase\expandafter{\romannumeral 1}.} 
No arc connecting $\alpha$ and $\beta$:

{(1)} Direct Mapping from $f_I$ to $P_{\alpha}, P_{\beta}$ via an MLP (DM-V): feeding $f_I$ into MLP-Sigmoids to predict $P_{\alpha}, P_{\beta}$.

{(2)} DM Linguistic feature (DM-L): replacing the $f_I$ of DM-V with linguistic feature $f_L$, which is the expectation of Bert~\cite{bert} embeddings of category names w.r.t $P(O_i|I)$.

{(3)} Visual-Linguistic alignment, \ie, Multi-Modality (MM): mapping $f_I$ to a latent space and minimizing the distance to $f_L$, feeding it to an MLP-Sigmoids to get $\alpha, \beta$.

{(4)} Linguistic Correlation of $O$-$\alpha$, $O$-$\beta$ (LingCorr): measuring the correlation between object and $\alpha$ or $\beta$ classes via their Bert~\cite{bert} embedding cosine similarities. $P_{\alpha}$, $P_{\beta}$ are given by multiplying $P(O|I)$ to correlation matrices.

{(5)} Kernelized Probabilistic Matrix Factorization (KPMF)~\cite{kpmf}: calculating feature similarity to all training samples as weights. Taking the weighted sum of GT $\alpha$ or $\beta$ of training samples as predictions.

{(6)} $\mathbf{A\&B}$ Lookup: getting $P_{A}, P_{B}$ from $M_{A}, M_{B}$.

{(7)} Hierarchical Mapping (HMa): mapping $f_I$ to category-level attribute or affordance space by an MLP, then feeding it to an MLP-Sigmoids to predict $P_{\alpha}$ or $P_{\beta}$.

{\bf Fold \uppercase\expandafter{\romannumeral 2}.}
$\beta \rightarrow \alpha$:

{(8)} DM from $\beta$ to $\alpha$ (DM-$\beta\rightarrow\alpha$): same as DM-V but using $f_{\beta}$ to infer $\alpha$.

{(9)} DM from $\beta$ and $I$ to $\alpha$ (DM-$\beta I \rightarrow\alpha$): same as DM-V but using both $f_I$ and $f_{\beta}$ to infer $\alpha$.

{\bf Fold \uppercase\expandafter{\romannumeral 3}.} 
$\alpha \rightarrow \beta$:

{(10)} DM from $\alpha$ to $\beta$ (DM-$\alpha\rightarrow\beta$): same as DM-V but using both $f_I$ and $f_{\alpha}$ to infer $\beta$.

{(11)} DM from $\alpha$ and $I$ to $\beta$ (DM-$\alpha I \rightarrow\beta$): same as DM-V but using both $f_I$ and $f_{\alpha}$ to infer $\beta$.

{(12)} Retrieving $\alpha$-$\beta$ relation by Ngram~\cite{ngram} (Ngram): adopting Ngram to retrieve the relevance of $\alpha$ \& $\beta$. Then we use DM predicted $\alpha$ and the relevance to estimate $\beta$.

{(13)} Markov Logic Network~\cite{mln} (MLN-GT): using \textbf{GT} $\alpha$ to infer $\beta$ with MLN.

{(14)} Instantiation with attention (Attention): feeding $[f_\alpha, f_I]$ to an MLP-Sigmoid to generate attentions and predicting $P_\beta$ by multiplying the attentions with $P_B$.

{(15)} DM with multi-head attention (DM-att): the $\alpha$ and $\beta$ features are sent to multi-head attention to learn their interaction, then use MLP-Sigmoids to get predictions.

{(16)} Vanilla CLIP: CLIP~\cite{clip} trained from scratch.

\subsection{ITE loss}
Though machines are expected to learn the causalities given $\alpha, \beta$ labels only. 
We wonder how it would perform given \textit{causal supervision}. We adopt an extra Hinge loss to maximize the ITE score of all [$\alpha_p, \beta_q$]. 
In detail, we intend the ITE of causal relations larger than a margin $\tau$ (= 0.1 in experiments), so the loss term is:
\begin{equation}
    \begin{aligned}
    \label{eq:ite-loss}
        \left\{
            \begin{aligned}
                &\max\{0,\tau - \Delta\hat{\beta_q}\}, & \beta_q=1, \\
                &\max\{0,\tau + \Delta\hat{\beta_q}\}, & \beta_q=0.
    \end{aligned}
    \right.
    \end{aligned}
\end{equation}
We enumerate all \textit{annotated} [$\alpha_p, \beta_q$] of an instance to obtain $L_{ITE}$.
Different from the default, the total loss here is $L=\lambda_CL_C + L_I + \lambda_{ITE}L_{ITE}$.

\subsection{Implementation Details}
For a fair comparison, all methods adopt a shared COCO~\cite{coco} pre-trained ResNet-50~\cite{resnet} (frozen) to extract $f_I$ and use the same object boxes in training and inference.
In OCRN, the dimension of $f_I$ and all $f_{A_i},f_{B_i},f_\alpha,f_\beta$ is 1024. The individual features of each attribute category are 512d and aggregated to 1024d by an FC. 
We train the attribute module with a learning rate of 0.3 and batch size of 1024 for 470 epochs. Then the attribute module is frozen, and the affordance module is trained with a learning rate of 3.0e-3 and batch size of 768 for 20 epochs. In training, $\lambda_C=0.03$, $\lambda_{ITE}=3$. 

\subsection{Results}
Tab.~\ref{table:res-main} presents the results. 
We can find that the causal structure of the models matters in OCL. 
Comparing DM methods implementing different causal graphs (including $\alpha\rightarrow\beta$, $\alpha\leftarrow\beta$),
$\alpha$ as intermediate knowledge (DM-$\alpha \rightarrow \beta$ and DM-$\alpha I\rightarrow \beta$) could advance $\beta$ perception (DM-V). 
But when $\beta$ serves as intermediate (DM-$\beta\rightarrow \alpha$ and DM-$\beta I\rightarrow \alpha$), $\beta$ perception is comparable or even worse than DM-V. 
So the causal relation $\alpha\rightarrow\beta$ is more evident than $\beta\rightarrow\alpha$ in the realistic dataset, which supports our choice in Sec.~\ref{sec:task-overview} that we focus more on the $\alpha\rightarrow\beta$ arc and implement our model with only $\alpha\rightarrow\beta$.

\begin{table}[t]
  \centering
  \resizebox{0.5\textwidth}{!}{
          \begin{tabular}{l|l|cccc}
          \hline
          Fold & Method & $\alpha$ & $\beta$ & $\mathcal{S}_\text{ITE}$ & $\mathcal{S}_{\alpha\text{-}\beta\text{-ITE}}$ \\ 
          \hline
          \multirow{7}{*}{{\romannumeral 1} N/A} 
          & DM-V            & \underline{29.9} & 51.8 & - & - \\ 
          & DM-L            & 21.2 & 47.5 & - & - \\ 
          & MM              & 23.8 & 48.9 & - & - \\ 
          & LingCorr        &  7.9 & 25.9 & - & - \\ 
          & KPMF            & 25.4 & 49.1 & - & - \\ 
          & $A\&B$-Lookup   & 18.9 & 30.9 & - & - \\ 
          & HMa             & 28.6 & 51.7 & - & - \\ 
          & DM-att          & 21.9 & 49.2 & - & - \\ 
          & Vanilla CLIP    & 23.6 & 49.6 & - & - \\ 
          \hline
          \multirow{2}{*}{{\romannumeral 2}: $\beta \rightarrow \alpha$} 
          & DM-$\beta \rightarrow \alpha$     & 30.0  & 52.0  &   -   &   -   \\
          & DM-$\beta I \rightarrow \alpha$   & 29.5  & 51.8  &   -   &   -   \\
          \hline          
          \multirow{6}{*}{{\romannumeral 3}: $\alpha \rightarrow \beta $} 
          & DM-$\alpha\rightarrow\beta$           & 28.7 & \underline{52.6} & 7.6 & 6.7 \\
          & DM-$\alpha I\rightarrow\beta$         & 29.0 & \underline{52.6} & 8.1 & 7.0 \\
          & Ngram                                 & 22.6 & 50.8 & \underline{8.3} & 7.6 \\
          & MLN-\textit{GT}                       &  -   & 33.4 & \textbf{9.5} & \underline{9.1} \\
          & Attention                             & 24.1 & 48.9 & 8.1 & 7.1 \\
          & OCRN                                  & \textbf{31.6}& \textbf{53.3}& \textbf{9.5}& \textbf{9.2} \\
          \hline
          \hline
          \multirow{6}{*}{$\alpha \rightarrow \beta $} 
          & DM-$\alpha\rightarrow\beta$ w/ $L_{ITE}$           & 28.8 & 52.4 & 15.5 & 14.0 \\ 
          & DM-$\alpha I\rightarrow\beta$ w/ $L_{ITE}$          & \underline{29.0} & \underline{52.5} & 15.4 & 13.6 \\
          & Ngram w/ $L_{ITE}$           & 22.2 & 49.9 & 14.1 & 12.9 \\ 
          & MLN-\textit{GT} w/ $L_{ITE}$          &  -   & 33.7 & 12.3 & 11.8 \\ 
          & Attention w/ $L_{ITE}$       & 23.9 & 49.0 & \underline{17.8} & \underline{15.5} \\
          & OCRN w/ $L_{ITE}$            & \textbf{31.5} & \textbf{53.6} & \textbf{20.3} &  \textbf{16.9} \\
          \hline
          \end{tabular}
      } 
      \captionof{table}{
      OCL results. w/ $L_{ITE}$ means that training with ITE loss. The baselines in the upper block cannot operate ITE due to the model structure. 
      Different $\alpha$-$\beta$ relations are exploited for causal graph comparison.
      } 
      \label{table:res-main}
\end{table}

OCRN outperforms the baselines and achieves decent improvements on all tracks.
In terms of $\alpha$ recognition, with or without $L_{ITE}$, OCRN outperforms the second-best method with 1.7 and 2.5 mAP respectively.
As for $\beta$ recognition, the improvements are 0.7 and 1.1 mAP with or without $L_{ITE}$.
Comparatively, HMa utilizes the supervision of $A, B$, but it performs much worse. 
$A\&B$ Lookup directly uses GT $A, B$ to infer $\alpha, \beta$, but its poor performance verifies the significant difference between $A, B$ and $\alpha, \beta$. 
Moreover, we find that all methods perform better on $\beta$ than $\alpha$, and the improvement of OCRN on $\alpha$ is larger too.
This may be because $\alpha$ are more diverse than $\beta$, \eg, we can \texttt{eat} lots of \texttt{food}s, but \texttt{food}s usually have various attributes (\texttt{fruit} vs. \texttt{pizza}). And OCL also has fewer attribute classes than affordance classes (114 vs. 170).
Another reason is that the positive samples in $\beta$ labels (23.2\%) are much more than the positives in $\alpha$ labels (9.4\%). The different pos-neg ratio affects learning a lot and results in the above gap.

In ITE evaluation, without the guidance of $L_{ITE}$, all methods achieve unsatisfactory performances. However, OCRN still has an advantage. Only MLN-\textit{GT} adopting the first-order logic and \textit{GT} $\alpha$ labels is comparable with OCRN.
If trained with $L_{ITE}$ and direct causality labels, all methods perform much better to learn the causalities, \eg, on OCRN, the ITE loss brings 10.8 and 7.7 mAP improvements on the two ITE tracks.
Particularly, the typical deep learning model Attention performs best in baselines, but MLN-\textit{GT} no longer holds the advantage. 
Relatively, OCRN shows more improvements and outperforms Attention with 2.5 and 1.4 mAP on the two ITE tracks. 

We provide more visualizations and discussions in the supplementary.
In particular, we also apply OCRN to \textbf{Human-Object Interaction Detection}~\cite{hicodet}, where OCRN boosts the performances of multiple HOI models and verifies the generalization and application potential of OCL.

\subsection{Ablation Study}
We verify the components of OCRN on the validation set in Tab.~\ref{table:res-abl}.

\textbf{(1) Deconfounding.}
OCRN w/o deconfounding is implemented following Eq.~\ref{eq:attr-bias} and \ref{eq:aff-bias}, where $P(O|I)$ and $P(O|I,\alpha)$ are the category predictions of pre-trained detectors~\cite{swin}. All the $\alpha$, $\beta$, and ITE performances drop due to the object bias. 
For more bias analyses please refer to the supplementary.

\textbf{(2) Losses.}
The performances slightly drop after removing category-level $L_{A_i},L_{B_i}$, but significantly drop without instance-level $L_{\alpha},L_{\beta}$ by over 20 mAP.

\textbf{(3) Feature dimension.} 
We compare different dimentionality for feature $f_{A_i},f_{B_i},f_{\alpha},f_{\beta}$.
Smaller and larger feature sizes than 1024 all have degrading effects. 

\textbf{(4) ITE-related implementations.}
We probe some different methods:
(a) Mean aggregation: $f'_{\alpha}=\sum_i f_{\alpha_p}$;
(b) Max-pooling aggregation: $f'_{\alpha}$ is the max value of $f_{\alpha_p}$ as each component;
(c) Random counterfactual feature: assigned random vector as the counterfactual attribute feature (instead of zero vector) during ITE. 
These methods perform worse than the chosen setting on ITE performance but are comparable on $\alpha$ and $\beta$ performance.

\begin{figure}[t]
  \begin{minipage}{\hsize}\centering
      \resizebox{0.95\textwidth}{!}{
          \begin{tabular}{l|cccc}
          \hline
          Method & $\alpha$ & $\beta$ & $\mathcal{S}_\text{ITE}$ & $\mathcal{S}_{\alpha\text{-}\beta\text{-ITE}}$\\ 
          \hline
          OCRN                & {\bf 32.4} & {\bf 52.2} & {\bf 20.5} & {\bf 17.0}\\
          \hline 
          w/o deconfounding          & 32.1 & 51.8 & 18.2 & 16.1 \\
          w/o $L_{A_i},L_{B_i}$      & 32.1 & 51.8 & 19.8 & 16.7 \\
          w/o $L_{\alpha},L_{\beta}$ & 10.0 & 27.0 & 16.6 & 16.4 \\
          \hline
          128 Dims                   & 31.7 & 51.5 & 18.0 & 16.0 \\
          512 Dims                   & 32.3 & 52.1 & 19.9 & 16.7 \\
          2048 Dims                  & 32.2 & 51.5 & 19.1 & 16.3 \\
          \hline
          Mean aggregation           & 32.2 & 51.3 & 18.9 & 16.7 \\
          Max-pooling aggregation    & 32.1 & 49.1 & 19.0 & 16.8 \\
          Random counterfactual      & \textbf{32.4} & 51.8 &  5.1 &  5.1 \\
          \hline
          \end{tabular}
      } 
      \captionof{table}{Ablation study results (validation set).}
      \label{table:res-abl}
  \end{minipage}
\end{figure}

\subsection{Discussion}
Overall, OCL poses extreme challenges to current AI systems. It expects representative learning to accurately recognize attributes and affordances from raw data meanwhile causal inference to capture the causalities within diverse instances and contexts, \ie, both the \textit{intuitive System 1 and logical System 2}~\cite{system12}.
From the experiments, we find that models struggle to achieve satisfying results on all tracks \textbf{simultaneously}. Notably, it is difficult to achieve a satisfying ITE score via data fitting. There is much room for improvement.
For future studies, a harmonious performance on $\alpha, \beta$, and causality learning are encouraged to better capture object knowledge. 
Potential directions may include causal representation learning~\cite{causalrepresent}, neural-symbolic reasoning~\cite{neuralsymbolic}, and Foundation Models~\cite{instructgpt}. etc.

\section{Conclusion}
In this work, we introduce object concept learning (OCL) expecting machines to infer affordances and explain what attributes enable an object to possess them. 
Accordingly, we build an extensive dataset and present OCRN based on casual intervention and instantiation. OCRN achieves decent performance and follows the causalities well.
However, OCL remains challenging and would inspire a line of studies on reasoning-based object understanding.

\noindent{\textbf{Acknowledgment}}: Supported by the National Key R\&D Program of China (No.2021ZD0110704), Shanghai Municipal Science and Technology Major Project (2021SHZDZX0102), Shanghai Qi Zhi Institute, Shanghai Science and Technology Commission (21511101200).

{\small
\bibliographystyle{ieee_fullname}
\bibliography{egbib}
}

\clearpage

\appendix

We report more details and analyses here:

Sec.~\ref{sec:selection}: Category/Attribute/Affordance Selection

Sec.~\ref{sec:annotation-detail}: Annotation Details

Sec.~\ref{causal_graph}: Causal Graph

Sec.~\ref{sec:dataset}: OCL Characteristics

Sec.~\ref{sec:detailed-tde-metric}: ITE Metric Details

Sec.~\ref{sec:baseline-detail}: Baseline Details

Sec.~\ref{sec:result-analysis}: Detailed Result Analysis

Sec.~\ref{sec:application-hoi}: Application on HOI Detection 

Sec.~\ref{app:sec:debias}: Comparison on Debiasing

Sec.~\ref{app:sec:state}: Discussion about States

Sec.~\ref{app:sec:causal-graph}: Discussion about Causality and Causal Graph

Sec.~\ref{sec:classes-list}: Detailed Lists

\section{Category/Attribute/Affordance Selection}
\label{sec:selection}
\begin{figure*}[ht]
    \centering
    \vspace{-10px}
    \begin{subfigure}{0.33\textwidth}
        \centering
        \includegraphics[width=\textwidth]{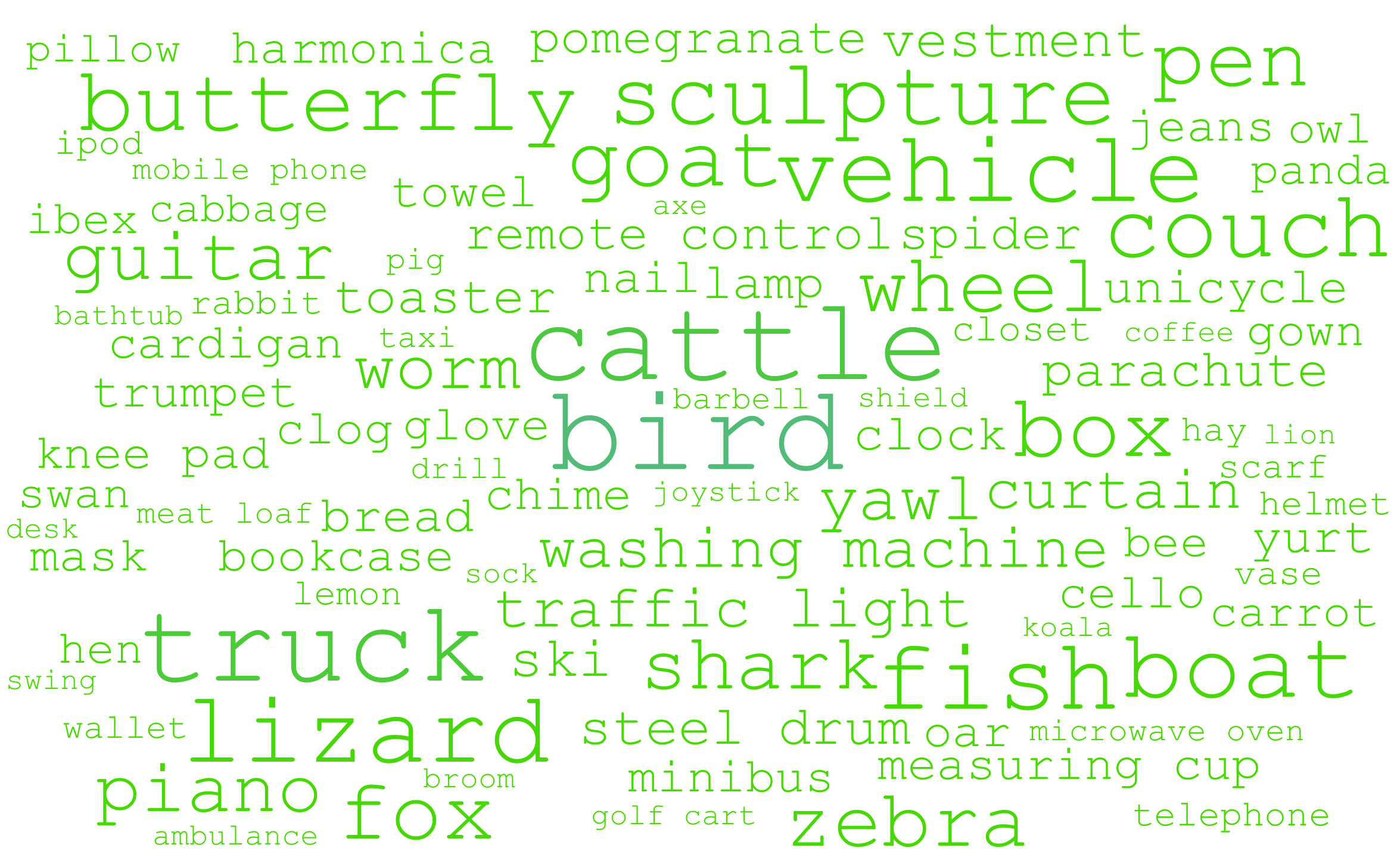}
        \caption{Category}
    \end{subfigure}%
    \begin{subfigure}{0.33\textwidth}
        \centering
        \includegraphics[width=\textwidth]{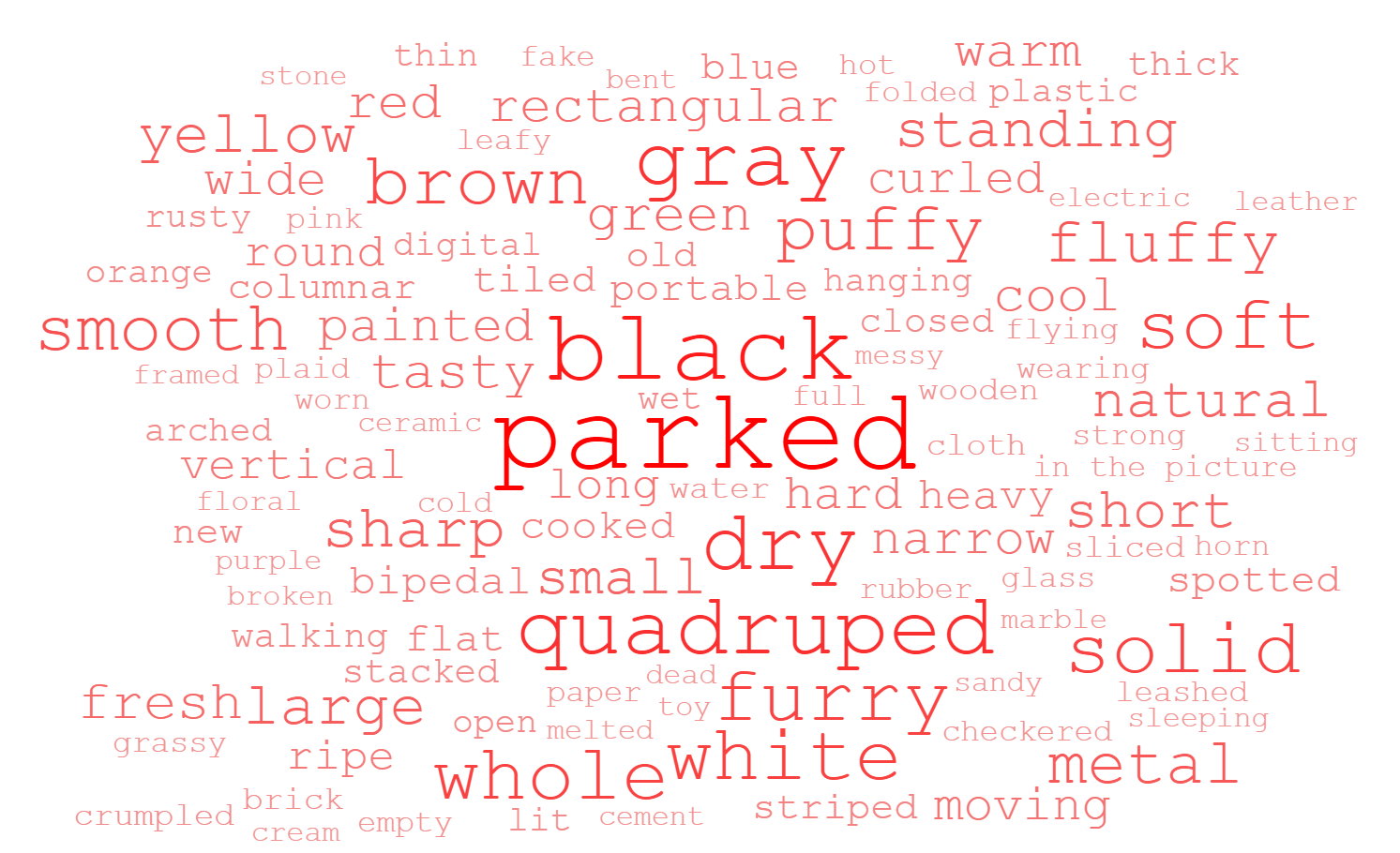}
        \caption{Attribute}
    \end{subfigure} %
    \begin{subfigure}{0.33\textwidth}
        \centering
        \includegraphics[width=\textwidth]{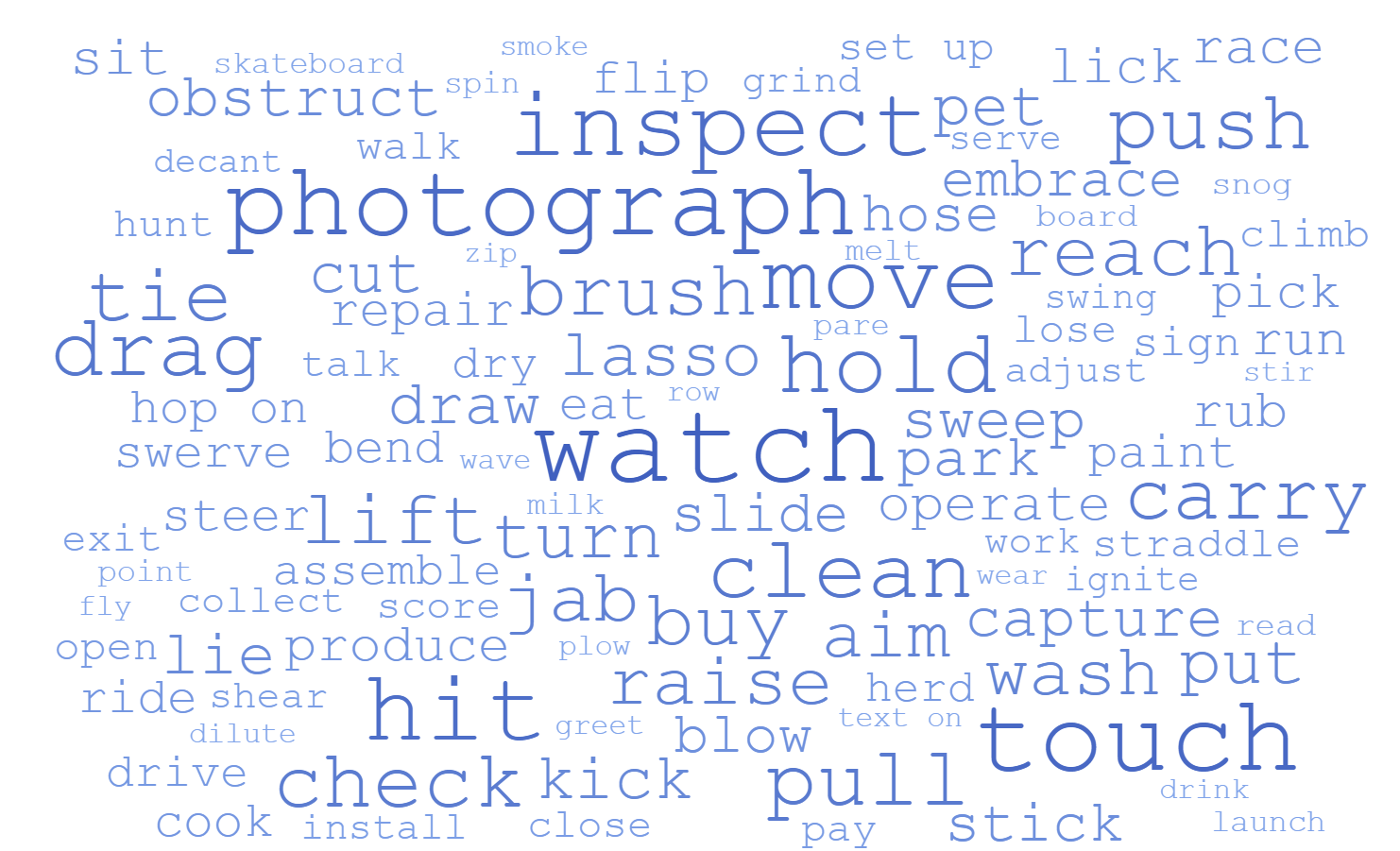}
        \caption{Affordance}
    \end{subfigure}
    \vspace{-10px}
    \caption{Word clouds of object categories, attributes, and affordance (by positive frequencies in OCL).}
    \label{fig:word-cloud}
    \vspace{-10px}
\end{figure*}

\begin{figure*}[ht]
    \centering
    \includegraphics[width=0.5\textwidth]{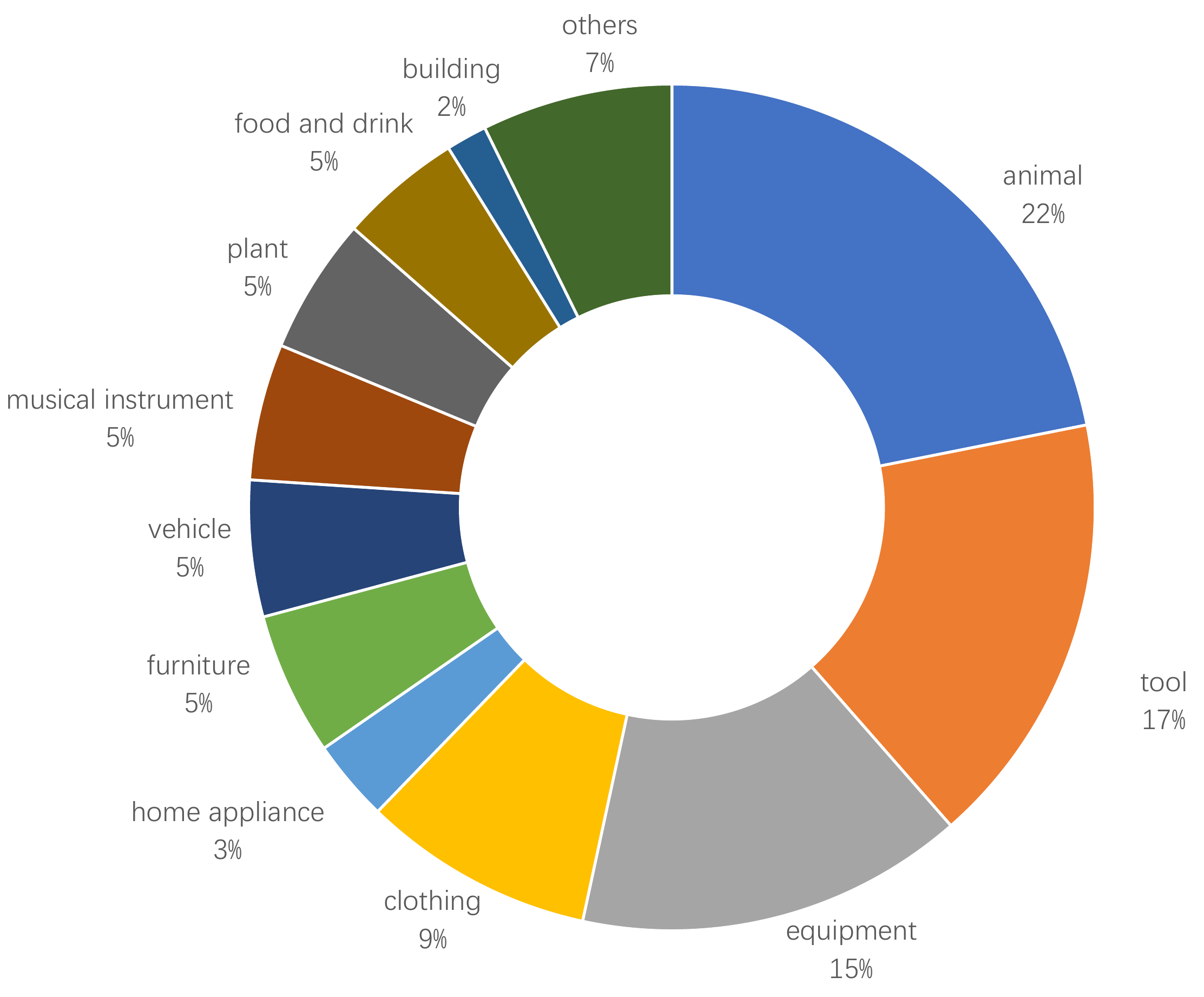}
    \caption{Super-categories of objects in OCL.}
    \label{fig:super-obj-class}
    \vspace{-10px}
\end{figure*}

We choose affordances, categories, and attributes, considering their causal relations. 
Their word clouds are shown in Fig.~\ref{fig:word-cloud}.
The complete lists can be found in Suppl. Sec.~\ref{sec:classes-list}. 

(1) {\bf Affordance}: 
To build a general and applicable knowledge base, we collect 1,006 affordance candidates from several widely-used action/affordance datasets: 957 from \cite{dengjia}, 160 from \cite{AVA}, 146 from \cite{hicodet}, 97 from \cite{vcoco}, 41 from \cite{yuke}, 21 from \cite{nguyen2017object} (with \textit{overlaps}). 
We find that not all affordances are in common use and some of them are difficult for visual recognition, \textit{e.g.}, \texttt{accept} (consider right and proper). So each candidate is scored by 5 human experts from 0.0 to 5.0 according to generality and commonness. We keep \textbf{170} top-scored affordances in our base (134 from \cite{dengjia}, 78 from \cite{AVA}, 127 from \cite{hicodet}, 53 from \cite{vcoco}, 13 from \cite{yuke}, 11 from \cite{nguyen2017object}, with \textit{overlaps}).

(2) {\bf Category}: 
Considering the taxonomy (WordNet~\cite{wordnet}), we collect a pool with over 1,742 object categories from previous datasets: 32 from \cite{apy}, 28 from \cite{cocoattr}, 717 from \cite{sun}, 1,000 from \cite{imagenet150k} (with \textit{overlaps}). 
Then we merge the similar categories according to WordNet~\cite{wordnet} and filter out the categories which are not common daily objects (\texttt{man}, \texttt{planet}), unrelated to the above 170 affordances (\texttt{skyscraper}) or too uncommon (\texttt{malleefowl}). Finally, our database has \textbf{381} common object categories. These object categories are divided into \textbf{12} super categories, shown in Fig.~\ref{fig:super-obj-class}.

(3) {\bf Attribute}: 
We extract the attributes from several large-scale attribute datasets: 64 from \cite{apy}, 203 from \cite{cocoattr}, 66 from \cite{sun}, 25 from \cite{imagenet150k}, top 500 from \cite{visualgenome}), and manually filter the 500 most frequent attributes. 
Five experts give 0 to 5 scores based on their relevance to human actions and the selected 170 affordances to better explore the causal relations between attributes and affordances. Some attributes (\texttt{cloudy}, \texttt{competitive}) that are not useful for affordance reasoning are discarded. Finally, \textbf{114} attributes are kept, covering colors, deformations, supercategories, surface, geometrical, and physical properties.

\section{Annotation Details}
\label{sec:annotation-detail}

\subsection{Attribute Annotation}

(1) \textbf{Category-level attribute} ($A$).
Following \cite{osherson1991default}, to avoid bias, annotators are given \textit{category-attribute pairs} (category \textit{names}, not images). They propose a 0-3 score according to the category concept in their minds (0: No, 1: Normally No, 2: Normally Yes, 3: Yes). Each pair is annotated by three annotators and takes the plurality as the $A$ label. If the range of 3 proposals exceeds 1, another three annotators will re-annotate this pair until achieving consensus. We binarize the annotations (0: No, 1: Yes) with a threshold of 2 and get a category-level attribute matrix $M_{A}$ ($[381,114]$).

(2) \textbf{Instance-level attribute} ($\alpha$).
Two annotators label each pair with 0 (No) and 1 (Yes). 
If they give different labels, this pair will be handed over to another two annotators until meeting consensus.

\subsection{Affordance Annotation}
\label{sec:beta-annotation}

(1) \textbf{Category-level affordance} $B$.
Following \cite{dengjia}, the annotators are given category-affordance pairs. 
The pairs are annotated in four bins (0-3) and normalized (same as $A$) to describe the possibility of an affordance in a category. 
Each pair is annotated by three annotators and makes consensus the same as $A$. The 0-3 scores are binarized (1: Yes, 0: No) with a threshold of 2.
The final category-level affordance matrix $M_{B}$ is $[381,170]$.

(2) \textbf{Instance-level affordance} $\beta$ is annotated for \textbf{every instance} with the help of \textit{object states}~\cite{mit}. 
As $B$ is determined by common states, objects in specific states may have different affordances from $B$, \textit{e.g.}, we cannot \texttt{board} a \texttt{flying} plane.
As the instances in the same state should have similar $\beta$ (all \texttt{rotten apple}s cannot be \texttt{eaten}),
six experts first conclude the states. The experts scan all instances of each category and use their knowledge of affordance to define all the existing states. 
Then all 186 K instances are dispatched to the concluded states via crowdsourcing. If some instances do not belong to any predefined states, they will be returned to the experts to add more states. In total, \textbf{1,376} states are defined, and each category has 3.6 states on average. 
Next, $\beta$ is annotated for each state. Given a \textit{state-affordance pair} and example images, two annotators mark it with 0 (No) and 1 (Yes). The results are combined in the same way as $\alpha$. Thus, each instance would have a state and the corresponding $\beta$.
An annotator would recheck each instance together with its state and $\beta$ labels to ensure the quality. 
If its state is inaccurate or the state $\beta$ labels are unsuitable, this annotator would correct them.

\begin{figure*}[ht]
    \centering
    \includegraphics[width=0.7\textwidth]{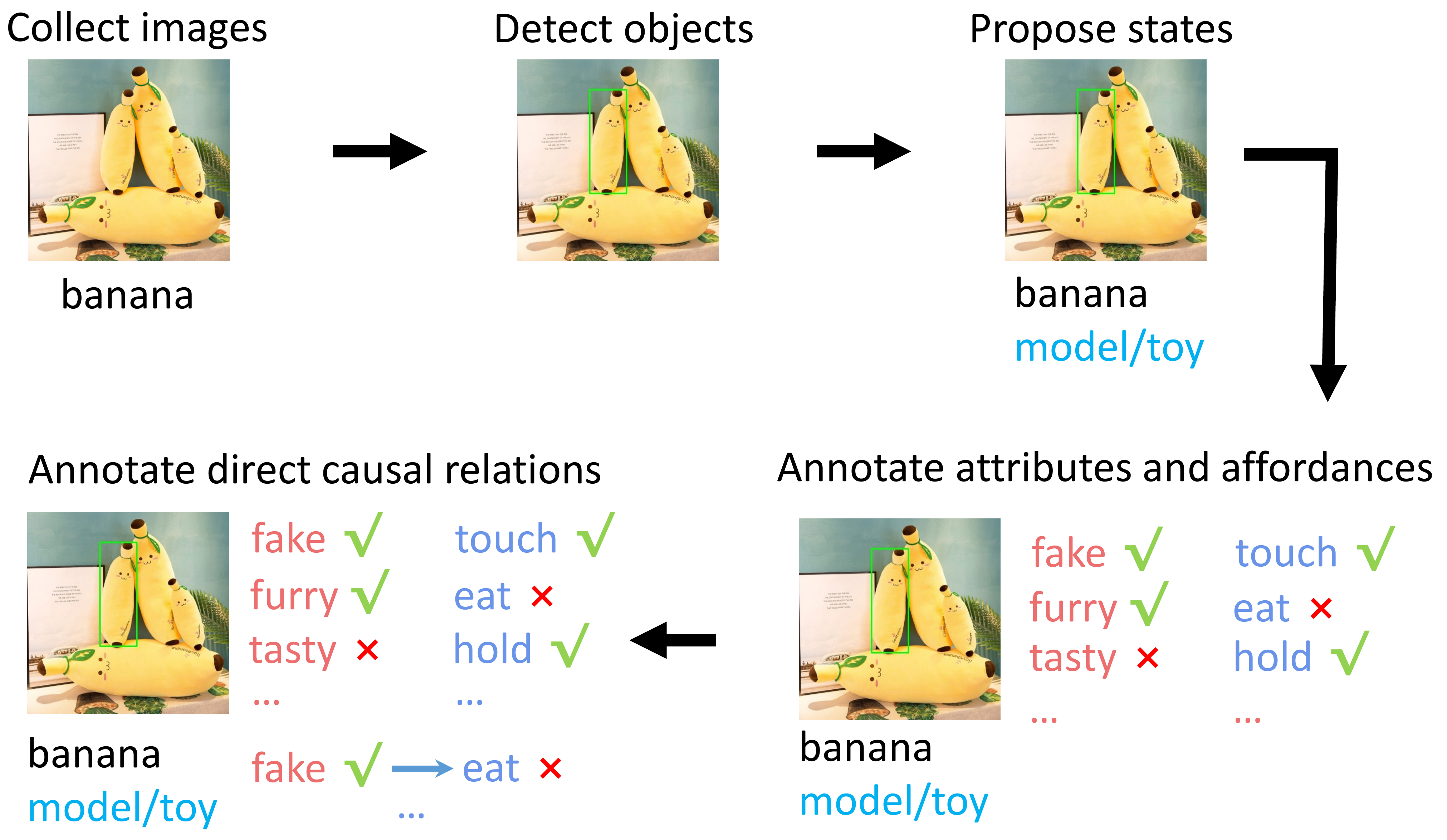}
    \vspace{-10px}
    \caption{A running example of dataset construction.}
    \label{fig:OCL_running_anno_sample}
    \vspace{-10px}
\end{figure*}

\begin{figure}[ht]
\begin{center}
    \includegraphics[width=0.6\linewidth]{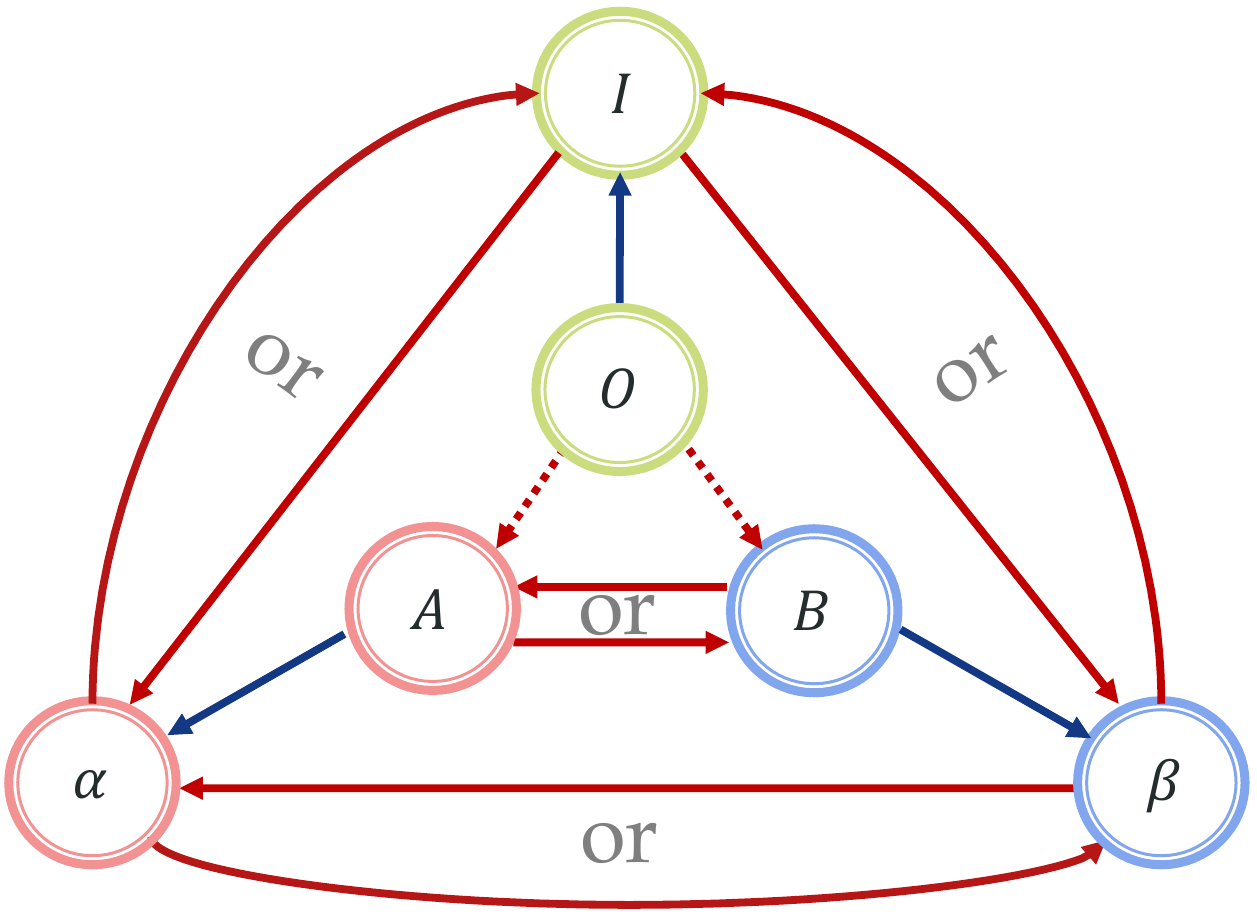}
    \vspace{-10px}
    \caption{A more complex causal graph of our knowledge base. $A, B, O$ are the object category and category-level attribute and affordance. $I$ is the object appearance, $\alpha, \beta$ are the instance-level attribute and affordance.
    Note that ``or'' indicates that the arcs between $A, B$, $\alpha, \beta$, $I, \alpha$, and $I, \beta$ indicate that either $A \leftarrow B$ or $A \rightarrow B$ (the others are similar) is considering in the setting.
    } 
    \label{fig:full-causalgraph-suppl}
\end{center}
\vspace{-18px}
\end{figure}

\subsection{Causal Relation Annotation}
\label{sec:detail_causal_anno}
\textbf{(1) Filtering}.
As exhaustive annotation is arduous, we only annotated existing rules without ambiguity.
Starting from the [114,170] matrix of $\alpha$-$\beta$ classes, we ask three experts to vote on the causal relation of each class. They scan all instances to answer whether the relationship exists in any case. 
That is, we just annotate the \textit{least} pairs with the \textit{largest} possibility to be casually related. Some causal pairs may be excluded. 
In detail, for each of the 114$\times$170 $\alpha$-$\beta$ pairs, we attach 10 samples for reference and 3 experts vote \texttt{yes/no/not sure}. We take the majority vote and the \texttt{not sure} and controversial pairs are rechecked. The \texttt{not sure} and \texttt{no} pairs are removed, and so do the \textbf{ambiguous} pairs. 
The pairs we selected are checked carefully to ensure the causalities and we only evaluate models on them. Thus, the missed causal pairs or non-causal pairs would not affect the results.
Finally, we obtain about 10\% $\alpha$-$\beta$ classes as candidates. 
The left 90\% pairs may hold value and we will mine new rules with LLMs in future work, especially from ambiguous pairs.

\textbf{(2) Instance-level causality}: we also adopt object states as a reference.
For each \textit{state-$\alpha$-$\beta$} triplet, two annotators are asked whether the specific attribute is the \textit{direct} and \textit{unambiguous} cause of this affordance in this state and gives their binary answer. We use the same method in annotating $\beta$ to combine results and assign \textit{state-level} labels to instances. 
Next, for all instances of a state, an expert decides whether the state-level relations are reasonable for each \textit{instance} in specific contexts and correct the inaccurate ones.
Finally, we obtain about 2 M \textit{instance-$\alpha$-$\beta$} triplets of causal relations.

\subsection{A Running Example of Dataset Construction.}
A running example is shown in Fig. \ref{fig:OCL_running_anno_sample} to show the process of annotations clearly.

\section{Causal Graph}
\label{causal_graph}
In this section, we first briefly introduce the causal graph model and causal intervention. Then we introduce the details of the causal graph our knowledge base can support. Then, we detail the implementation of the causal graphs used by different methods.

\subsection{Basics of Causal Inference and Causal Graph}

\begin{figure}[ht]
    \centering
    \includegraphics[width=0.6\linewidth]{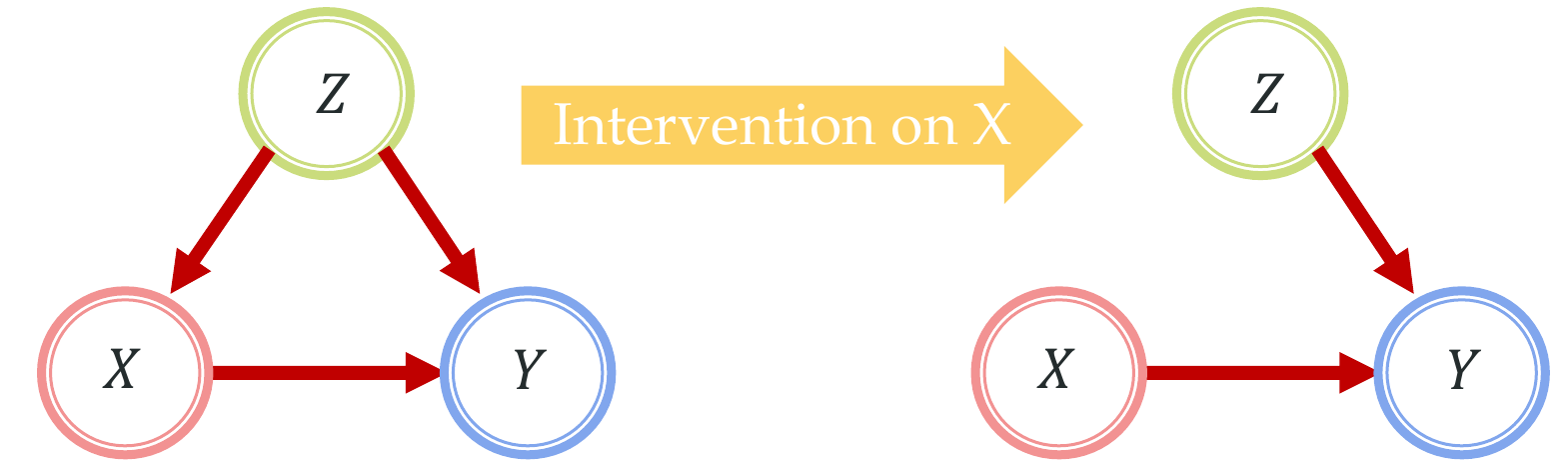}
    \caption{An example of the causal graph and causal intervention. We study the causal relation $X\rightarrow Y$ while confounder $Z$ exists and brings bias. After the intervention on variable $X$, the poisonous relation $Z\rightarrow X$ is eliminated.} 
    \label{fig:causal-primer}
\end{figure}

A causal graph is a DAG that describes the causal relations between multiple factors. Each directed edge points from the ``cause'' to its ``effect'', \eg in Fig.~\ref{fig:causal-primer}, node $X$ is the cause of node $Y$. Under the scenario that causal variables and causal graphs are known, \textbf{causal inference} studies how to infer the strength of causal edges given observations, or infer the outcomes given some of the causal variable values.

However, the causal relation in the real world is sophisticated. The causal relation that we observed may have been polluted by spurious variables. 
For example, let $X$ in Fig.~\ref{fig:causal-primer} be ice cream sales and $Y$ be drownings, one may observe that more ice cream sales lead to more drownings and infer that they are causally related. Actually, the observed relation is due to another factor $Z$: weather temperature. 
These variables are called \textbf{confounders}, which is the common cause of two causal variables that we are studying, \eg in the left graph in Fig.~\ref{fig:causal-primer}, $Z$ is a confounder when we focus on the causal edge $X\rightarrow Y$.

In causal inference, confounders should be eliminated to avoid biases on causal learning, by applying \textbf{intervention} on the cause variables (\eg $X$ in our example) to ``control'' its distribution to block the effect of confounder.
Traditional scientific research on causality adopts Randomized Controlled Trial (RCT) to completely remove the confounder, but it is not applicable when we only have observational data.
Pearl.~\cite{pearl2016causal} \textit{et al.} propose \textit{do-calculus} to systematically analyze the causal graph and alleviate the confounder bias in a probabilistic view. In the simple case in Fig.~\ref{fig:causal-primer}, the confounder $Z$ can be eliminated with \textbf{Back-door Adjustment}:
\begin{equation}
    P(Y|do(X))=\sum_zP(Y|X, Z=z)P(Z=z),
\end{equation}
where $z$ is the specific value of the random variable $Z$. The causal graph of our OCRN also meets the back-door criterion so we apply the back-door adjustment to alleviate bias from the confounder $O$.

\subsection{Causal Graph of Our Knowledge Base}
A more complicated causal graph considering more arcs between nodes is shown in Fig.~\ref{fig:full-causalgraph-suppl}.
The causal relations between nodes or arcs in Fig.~\ref{fig:full-causalgraph-suppl} are determined as follows:

Firstly, we introduce two kinds of special arcs.

$O \rightarrow A$, $O \rightarrow B$ (dotted arcs):
in OCL, $A$ and $B$ are defined as the category-level annotations. Given $O$, $A$, and $B$ are strictly determined. In Fig.~\ref{fig:full-causalgraph-suppl}, we use two dotted arrows from $O$ to $A, B$ respectively to indicate this deterministic relation to distinguish them from the other causal relations.

$O \rightarrow I$, $A \rightarrow \alpha$, $B \rightarrow \beta$ (blue arcs):
we see the category-level $O$, $A$, and $B$ are direct causes of instance-level $I$, $\alpha$, and $\beta$ during the concept \textit{instantiation} according to OCL definition. 
Because the visual representation $I$ and properties $\alpha$, $\beta$ of an instance are derived from the concept-level categorical ones.
The reversed arcs ${O}\leftarrow{I}$, ${A}\leftarrow{\alpha}$, ${B}\leftarrow{\beta}$ mean that $O$, $A$, $B$ are the \textit{aggregations} of instances and would be marginally affected by one specific instance, thus we do not include these arcs here for clarity.

Next, we illustrate the regular causal arcs as follows.

$I \rightarrow \alpha$, $I \rightarrow \beta$: the recognition process of $\alpha$ and $\beta$. As $I$ indicates the \textit{physical noumenon}, it is the source of semantic and functional properties and decides/causes $\alpha, \beta$. 

$\alpha \rightarrow I$, $\beta \rightarrow I$: the generation of visual pattern from attribute or affordance descriptions and can be utilized in image generation/manipulation tasks~\cite{he2019attgan}.

$A \leftarrow B$ or $A \rightarrow B$, $\alpha \leftarrow \beta$ or $\alpha \rightarrow \beta$: the causal direction between attribute and affordance can be reversed sometimes. The arc from $\alpha$ to $\beta$ is evident, \textit{e.g.}, a \texttt{broken cup} is not \texttt{useable}. Sometimes, the reverse arc causal effect from $\beta$ to $\alpha$ also exists, \textit{e.g.}, an \texttt{eatable banana} would not be \texttt{unripe}. 

\begin{figure*}[ht]
    \centering
    \includegraphics[width=0.8\textwidth]{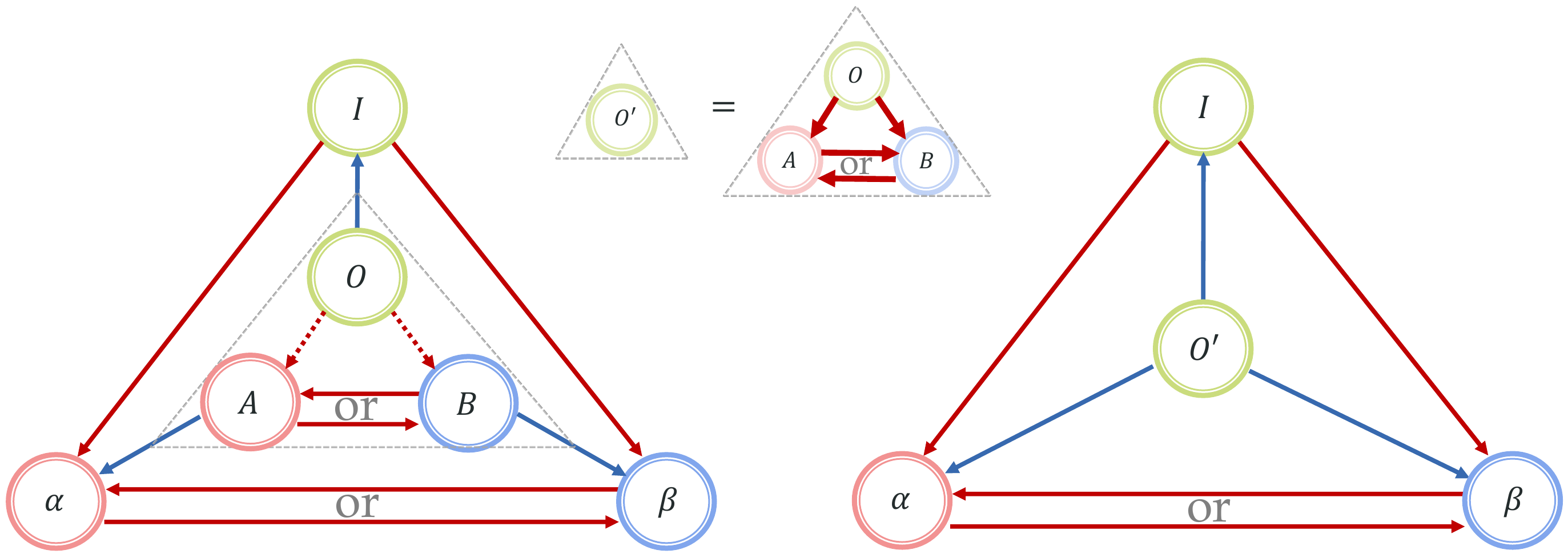} 
    \caption{Simplified causal graph for OCL task. Note that ``or'' indicates that the arcs between $A, B$ and $\alpha, \beta$ are either $A \leftarrow B$ or $A \rightarrow B$ ($\alpha \leftarrow \beta$ or $\alpha \rightarrow \beta$), instead of concurrence.}
    \label{fig:reduce-graph}
\end{figure*}

\begin{figure*}[ht]
\centering
\begin{subfigure}{0.3\textwidth}
    \centering
    \includegraphics[width=\textwidth]{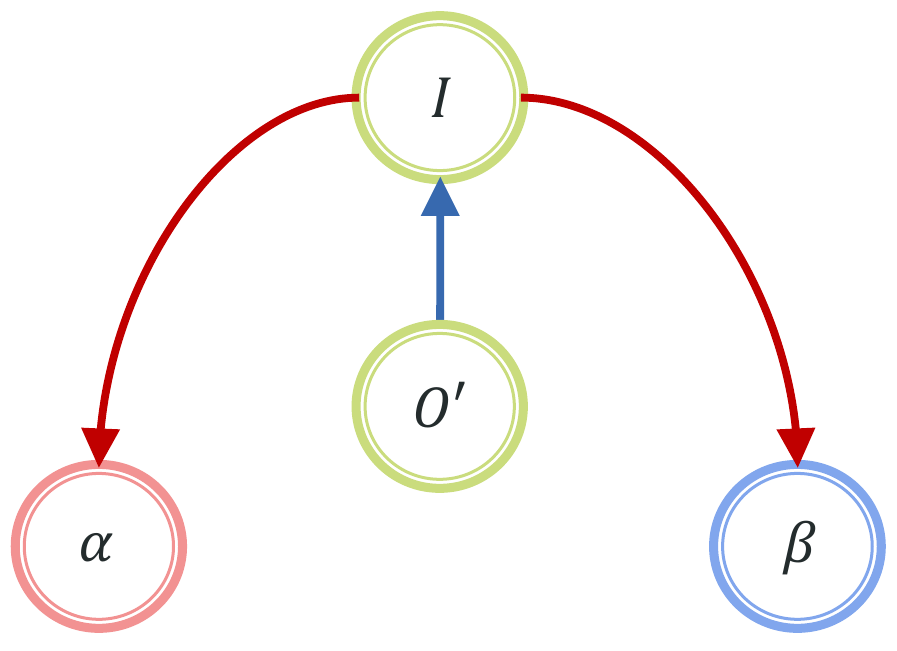}
    \caption{DM-V}
\end{subfigure}
\hfill
\begin{subfigure}{0.3\textwidth}
    \centering
    \includegraphics[width=\textwidth]{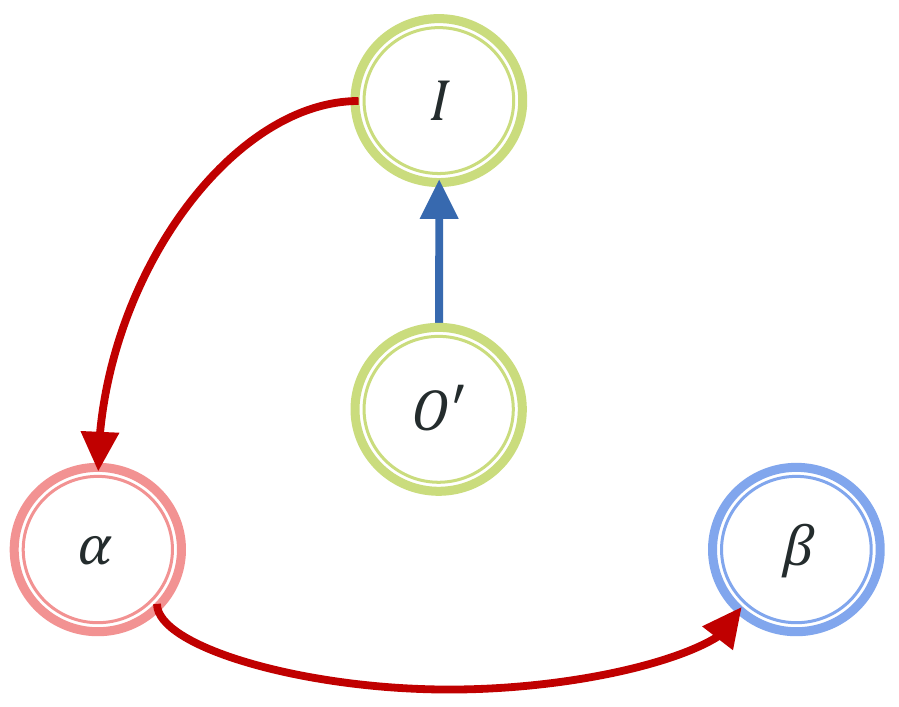}
    \caption{DM-$\alpha\rightarrow\beta$}
\end{subfigure}
\hfill
\begin{subfigure}{0.3\textwidth}
    \centering
    \includegraphics[width=\textwidth]{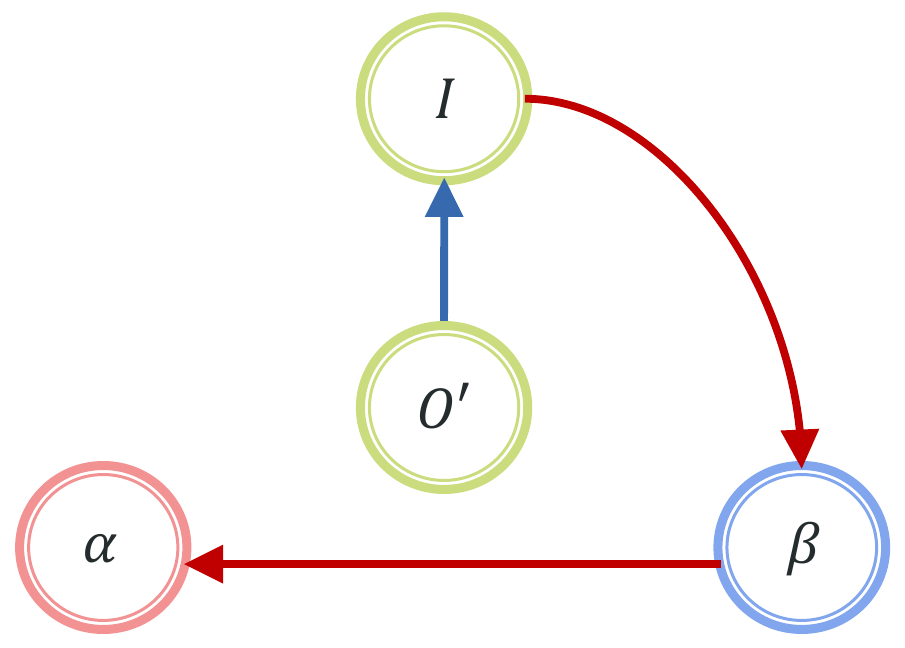}
    \caption{DM-$\beta\rightarrow\alpha$}
\end{subfigure}
\vfill
\begin{subfigure}{0.3\textwidth}
    \centering
    \includegraphics[width=\textwidth]{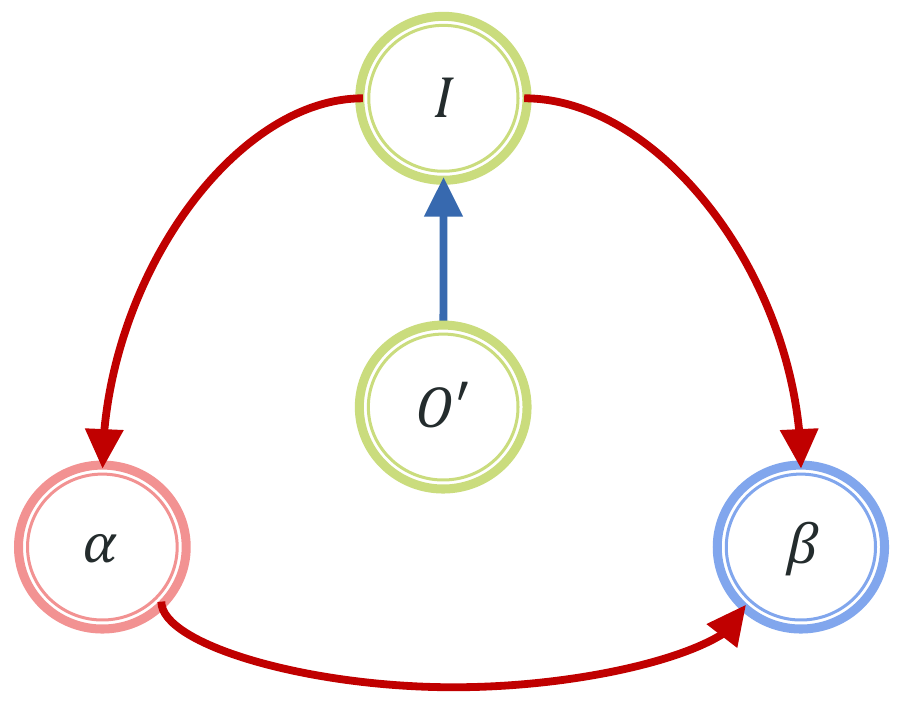}
    \caption{DM-$\alpha I\rightarrow\beta$}
\end{subfigure}
\hfill
\begin{subfigure}{0.3\textwidth}
    \centering
    \includegraphics[width=\textwidth]{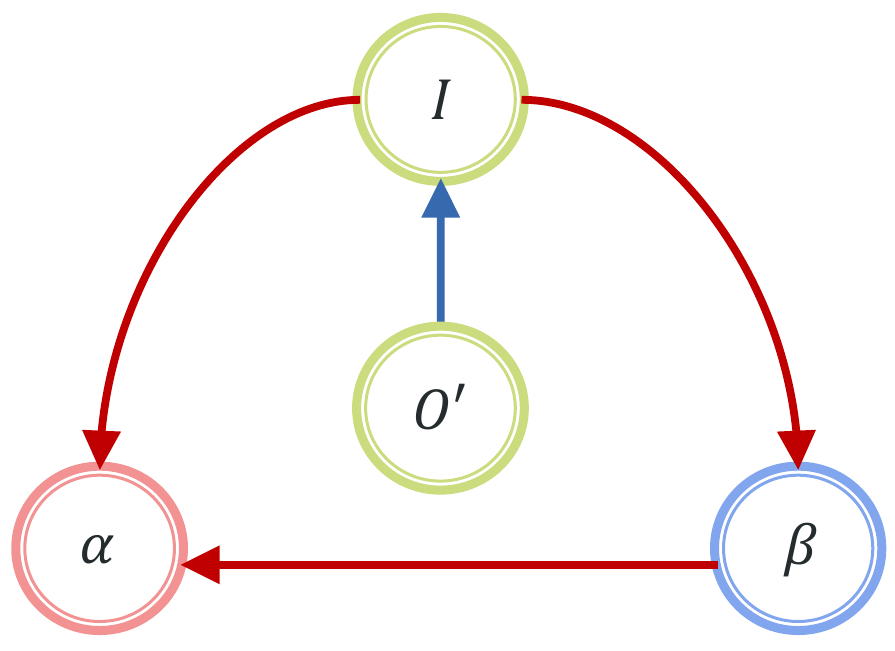}
    \caption{DM-$\beta I\rightarrow\alpha$}
\end{subfigure}
\caption{Causal graphs of the baselines.}
\label{fig:baseline-graph}
\end{figure*}

\begin{figure}[ht]
    \centering
    \includegraphics[width=0.3\textwidth]{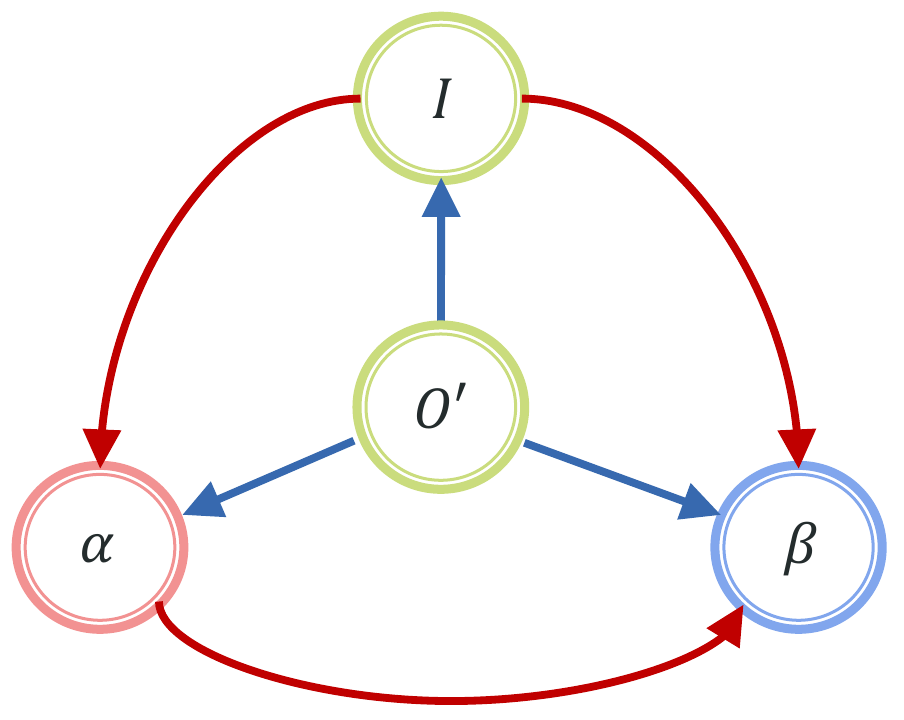}
    \caption{Causal graph of OCRN.}
    \label{fig:ocrn-graph}
\end{figure}

\subsection{Causal Graph Implementation}
\label{app:sec:graph-implm}

\begin{figure}[ht]
    \begin{center}
    \begin{minipage}{0.4\textwidth}
        \centering
        \includegraphics[width=\textwidth]{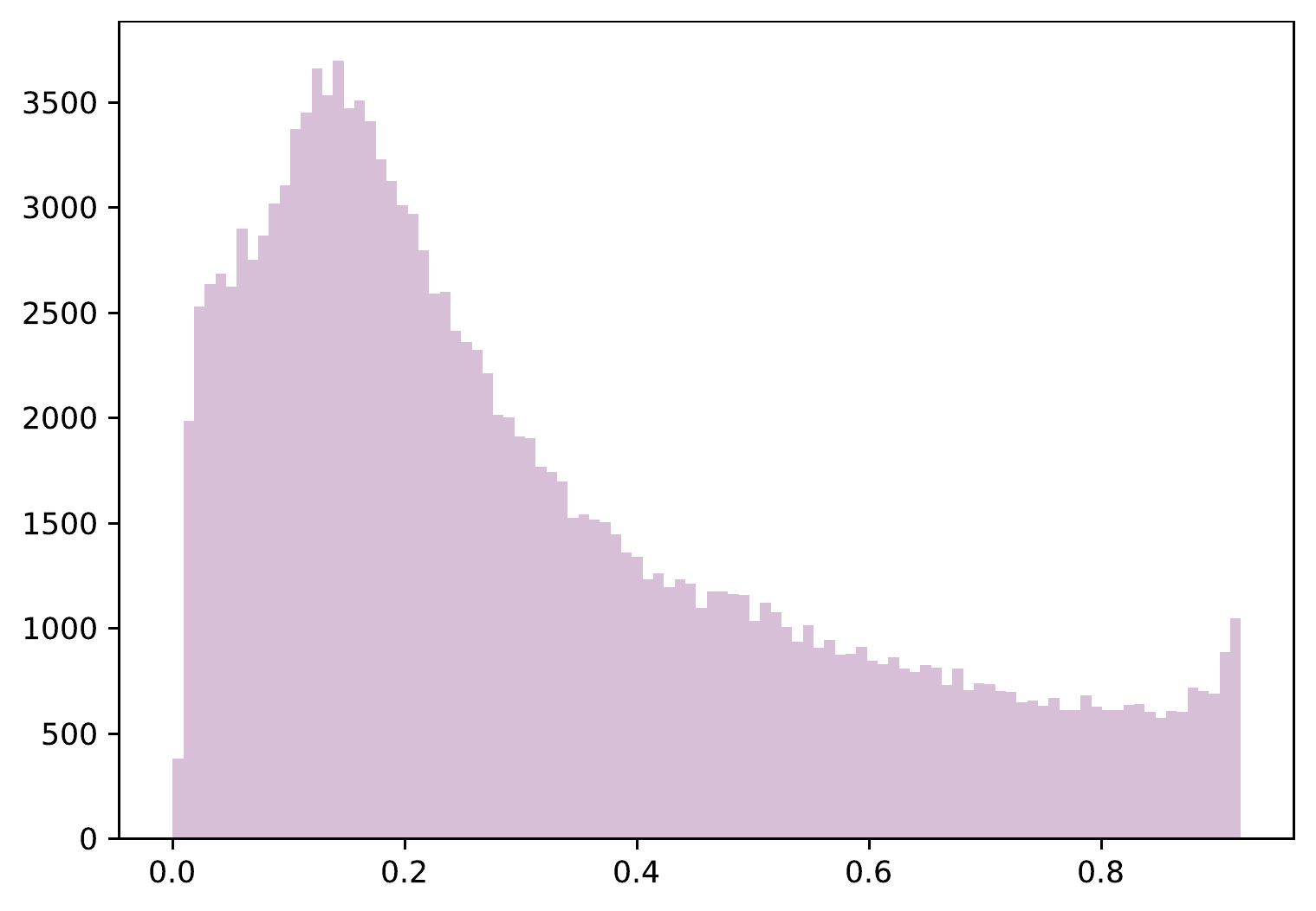}
    \end{minipage}
    \hfill
    \begin{minipage}{0.4\textwidth}
        \centering
        \includegraphics[width=\textwidth]{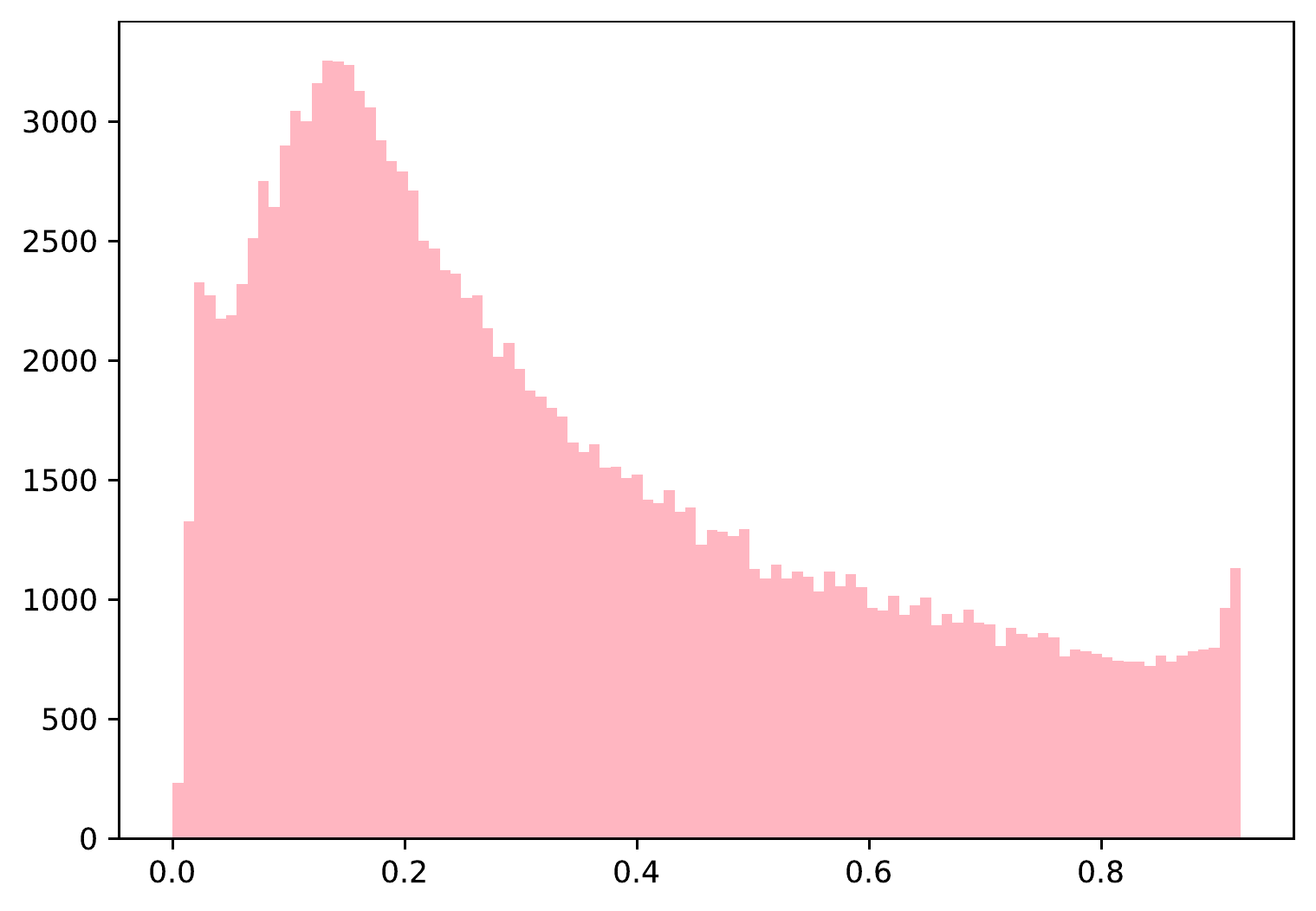}
    \end{minipage}
    \caption{Distribution of normalized object box width (left) and height (right).}
    \label{fig:box_size}
    \end{center}
\end{figure}

In this work, we mainly study the recognition and reasoning of attribute and affordance for robotics and embodied AI, hence we remove the two arcs corresponding to image generation $\alpha \rightarrow I$, $\beta \rightarrow I$.
Due to the deterministic relation between $O$, $A$, and $B$, we can simplify the three nodes to a single node $O'$ (Fig.~\ref{fig:reduce-graph}).

Different \textit{methods} can exploit different causal paths. We propose diverse baselines to implement different causal subgraphs, including the subgraphs with $\alpha\rightarrow\beta$, and $\alpha\leftarrow\beta$. The causal graphs of some baselines are shown in Fig.~\ref{fig:baseline-graph}.

The ablation experiment with arc ${\alpha}\rightarrow{\beta}$ and ${\alpha}\leftarrow{\beta}$ shows that the causal effect of ${\alpha}\rightarrow{\beta}$ is stronger than the alternative in our datasets.
Besides, from the aspect of embodied AI and robotics, affordance is more important in practical applications like object manipulation, so we focus more on affordance recognition and regard $\beta$ inference as our main goal. Therefore, in OCRN and some other baselines, we keep the arc ${\alpha}\rightarrow{\beta}$. And in causal reasoning, we focus on the evaluation of ${\alpha}\rightarrow{\beta}$ too.
The causal graph of OCRN is shown in Fig.~\ref{fig:ocrn-graph}.

\section{OCL Characteristics}
\label{sec:dataset}

\subsection{Object Box Size}
We visualize the distribution of normalized object box size in Fig.~\ref{fig:box_size}, where the box width and height are normalized by the width and height of the whole image. It shows that most objects in our knowledge base are \textit{small objects}, providing abundant regional information.

\subsection{Annotator Information} 
\begin{figure*}[ht]
\begin{minipage}{0.32\textwidth}
    \centering
    \includegraphics[width=0.9\linewidth]{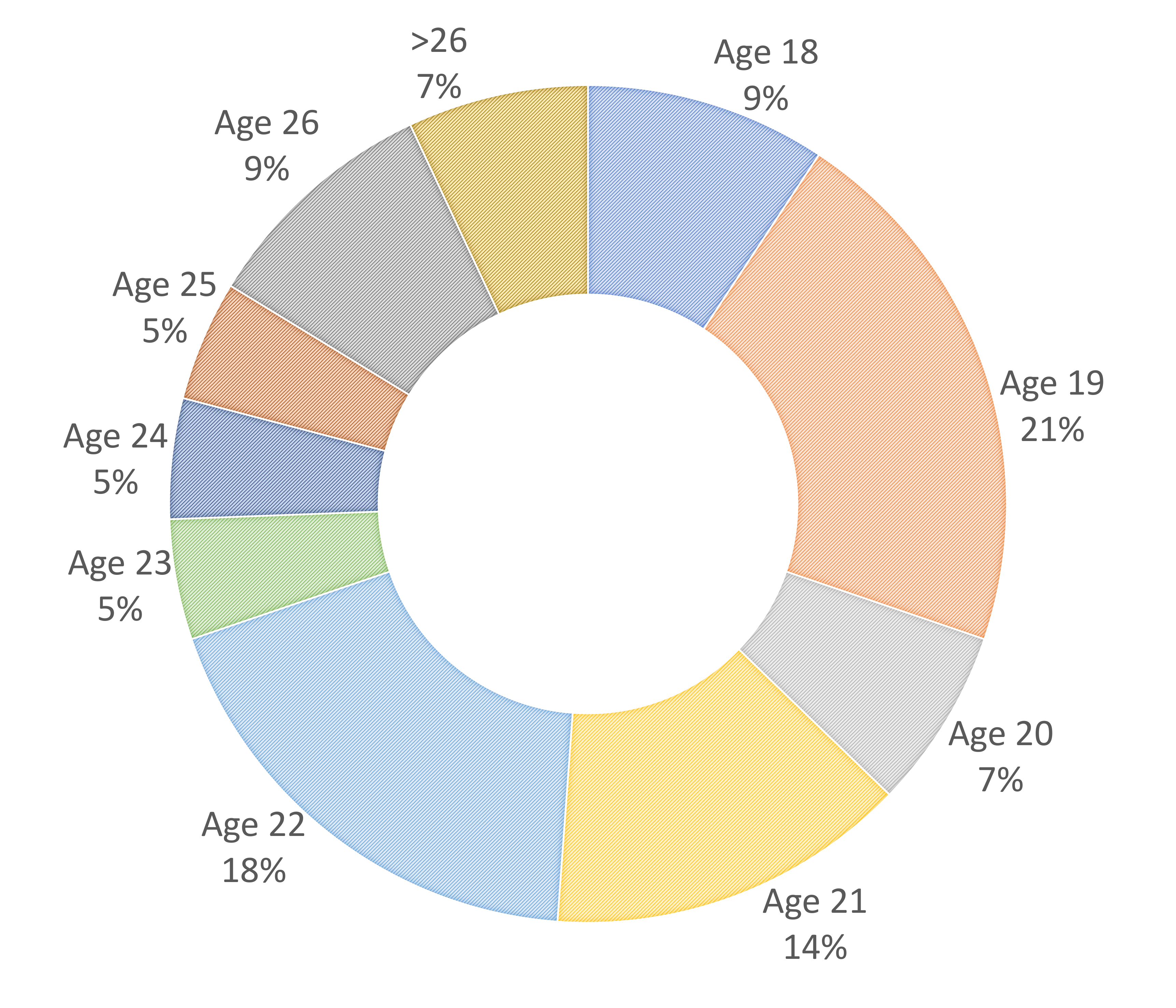}
    \caption{Age information of annotators.}
    \label{fig:age}
\end{minipage}
\hfill
\begin{minipage}{0.32\textwidth}
    \centering
    \includegraphics[width=0.9\linewidth]{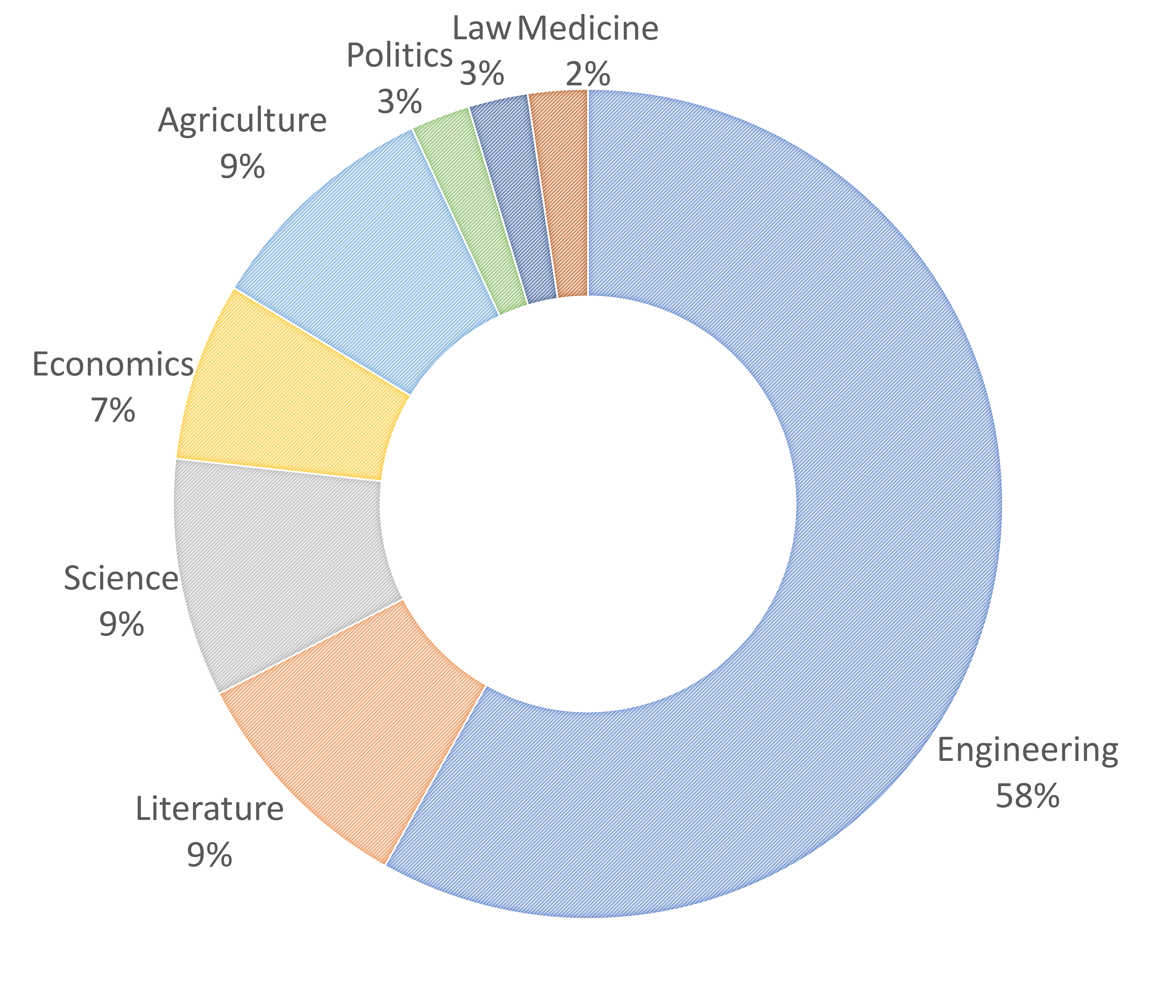}
    \caption{Major information of annotators.} 
    \label{fig:major}
\end{minipage}
\hfill
\begin{minipage}{0.32\textwidth}
    \centering
    \includegraphics[width=0.9\linewidth]{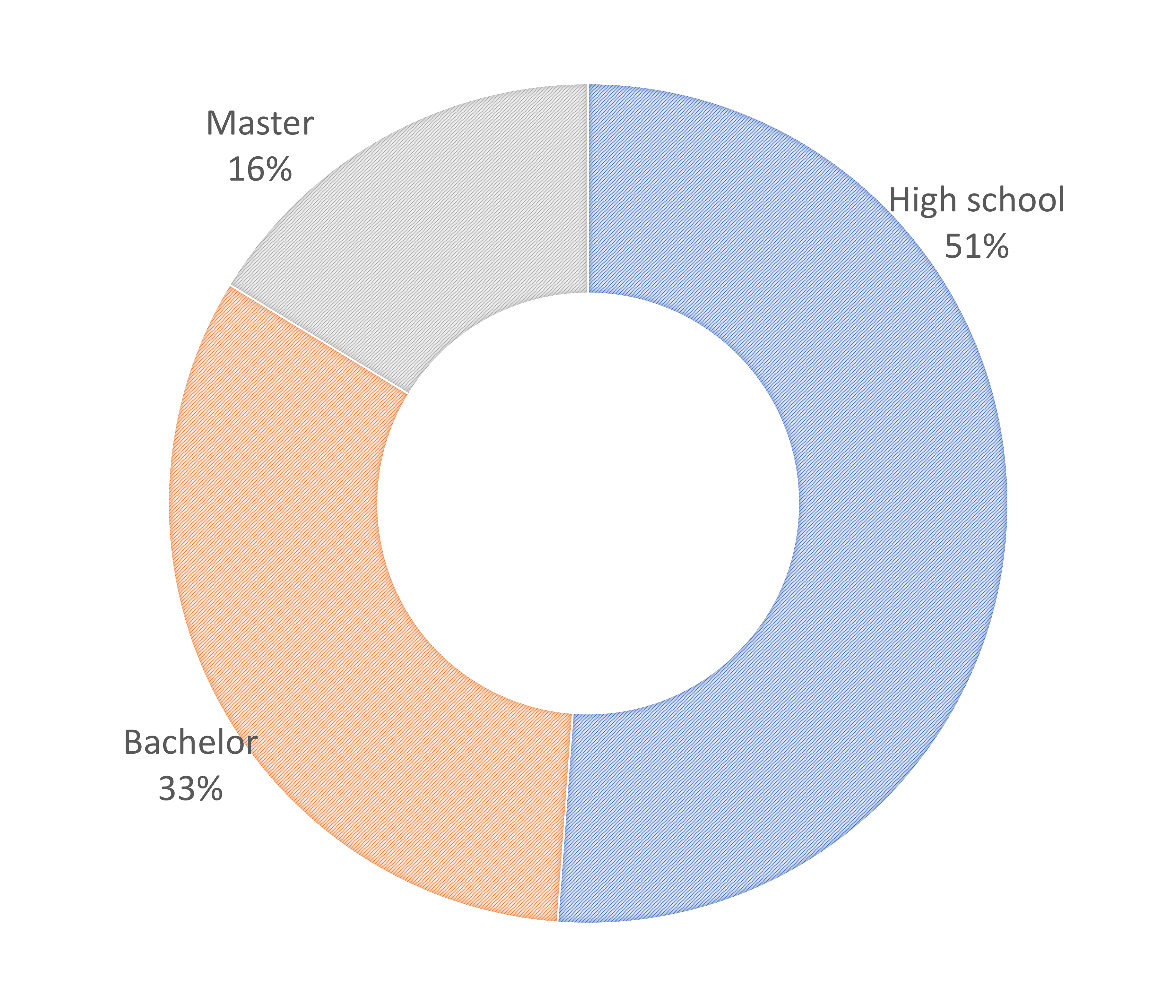}
    \caption{Degree information of annotators.} 
    \label{fig:degree}
\end{minipage}
\end{figure*}

Annotators' age, major, and education degree are presented in Fig.~\ref{fig:age},~\ref{fig:major}, and~\ref{fig:degree}.

\begin{figure*}[ht]
\begin{center}
\begin{minipage}{0.4\textwidth}
    \centering
    \includegraphics[width=\textwidth]{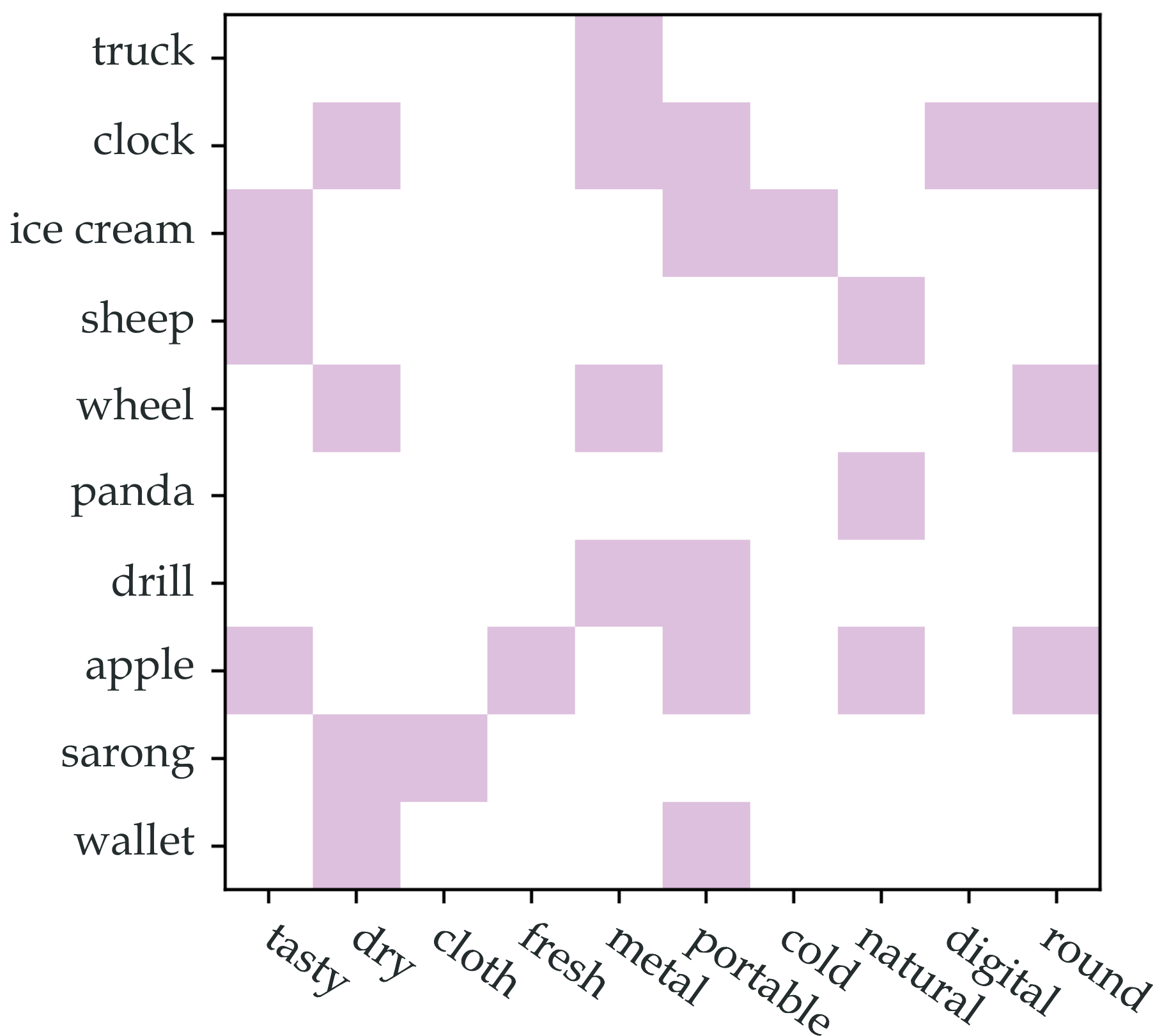}
    \caption{Category-level attribute ($A$) matrix.} 
    \label{fig:cat}
\end{minipage}
\begin{minipage}{0.4\textwidth}
    \centering
    \includegraphics[width=\textwidth]{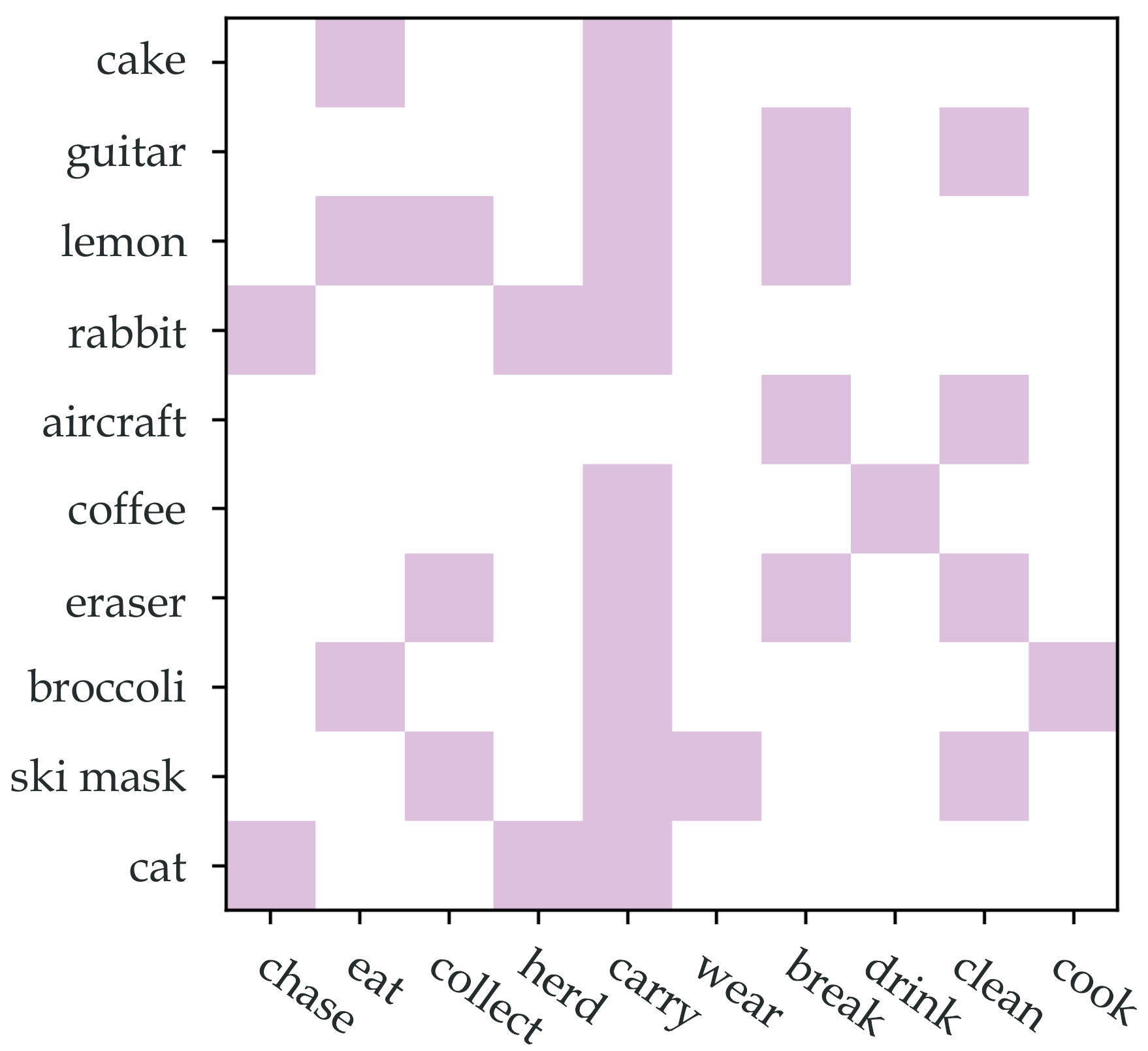}
    \caption{Category-level affordance ($B$) matrix.} 
    \label{fig:caf}
\end{minipage}
\end{center}
\end{figure*}

\begin{figure*}[ht]
    \centering
    \includegraphics[width=0.5\textwidth]{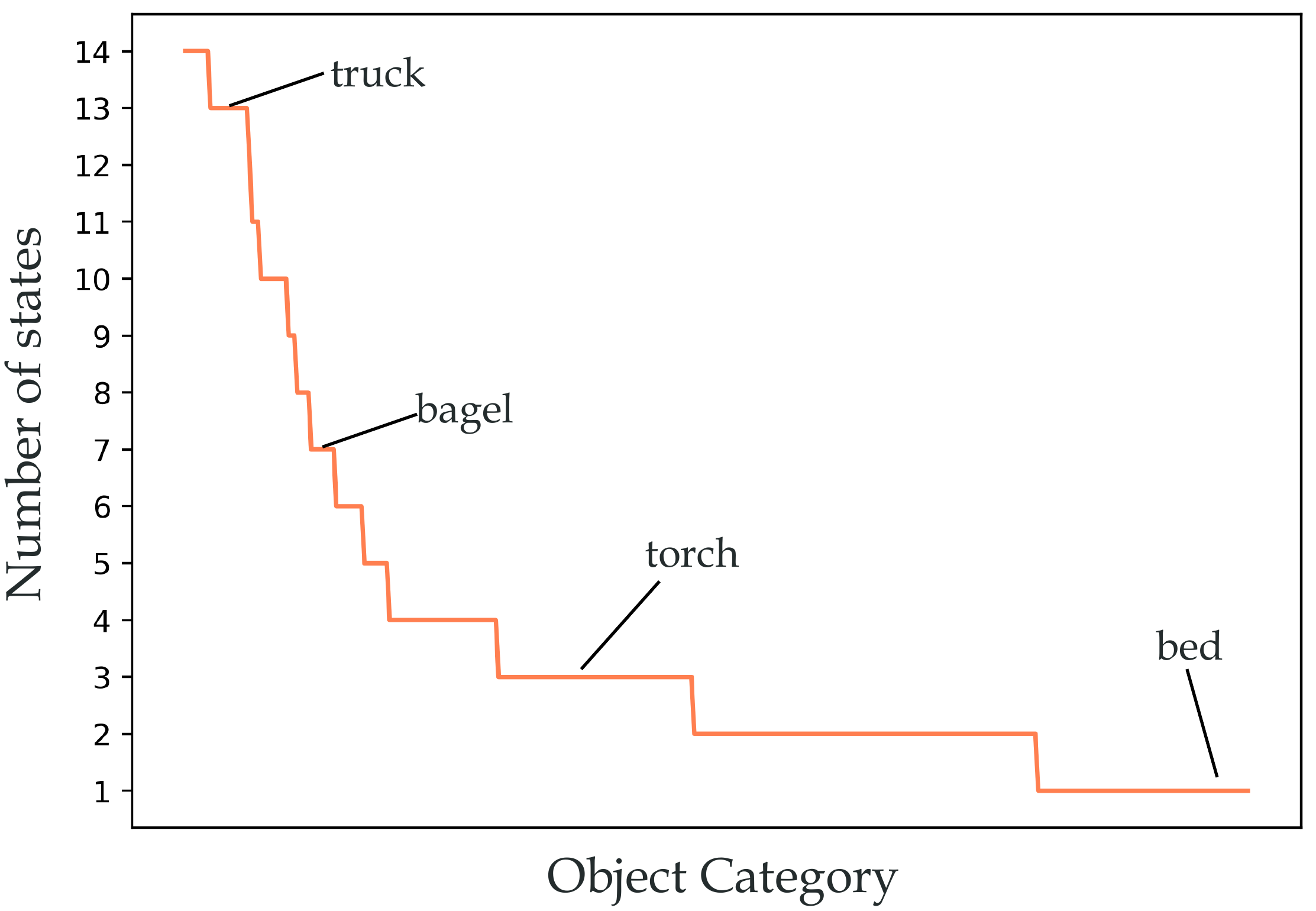}
    \caption{State distributions of different object categories.}
    \label{fig:state-distribution}
\end{figure*}

\subsection{Matrix Samples}
The category-level attribute and affordance ($A, B$) matrices are detailed in Fig.~\ref{fig:cat},~\ref{fig:caf} as heatmaps, and the cells with dark color indicate positive samples.
For example, \texttt{ice cream} is \texttt{cold} while \texttt{clock} is not \texttt{natural}, \texttt{cake} can be \texttt{eaten} while \texttt{eraser} can not be \texttt{cooked}. These are in line with our common sense.

\subsection{State Distribution}
Before annotating the affordances, we first define the object states for all object categories and annotate the state affordances. In total, we define 1,376 states for 381 object categories. 
And Fig.~\ref{fig:state-distribution} shows the state distribution per object category.

\subsection{Attribute-Affordance Relation}
We analyze the instance-level attribute-affordance relations in our knowledge base under three criteria. 
(1) \textbf{Attribute Conditioned Affordance Probability.} It is computed as $P(\beta|\alpha)$ to estimate affordance probability given an attribute. The range is [0,1]. 
(2) \textbf{Attribute-Affordance Correlation.} For all instances in our dataset, we evaluate the label correlation of each attribute-affordance pair, whose scale is in [-1,1]. 
(3) \textbf{Attribute-Affordance Causality}. Starting with the annotated cause-effect ($\alpha-\beta$) labels, we count for how many times each attribute-affordance pair appear in our dataset and normalize the value by the maximum occurrences, leading to a value in the range [0,1]. It should be mentioned that we only annotate whether an attribute-affordance pair \textbf{has} explicit and key causality, but the detailed effect (\textbf{positive} or \textbf{negative}) should be referred to instance labels.

\begin{figure*}[ht]
\begin{minipage}{0.3\textwidth}
    \centering
    \includegraphics[width=0.99\linewidth]{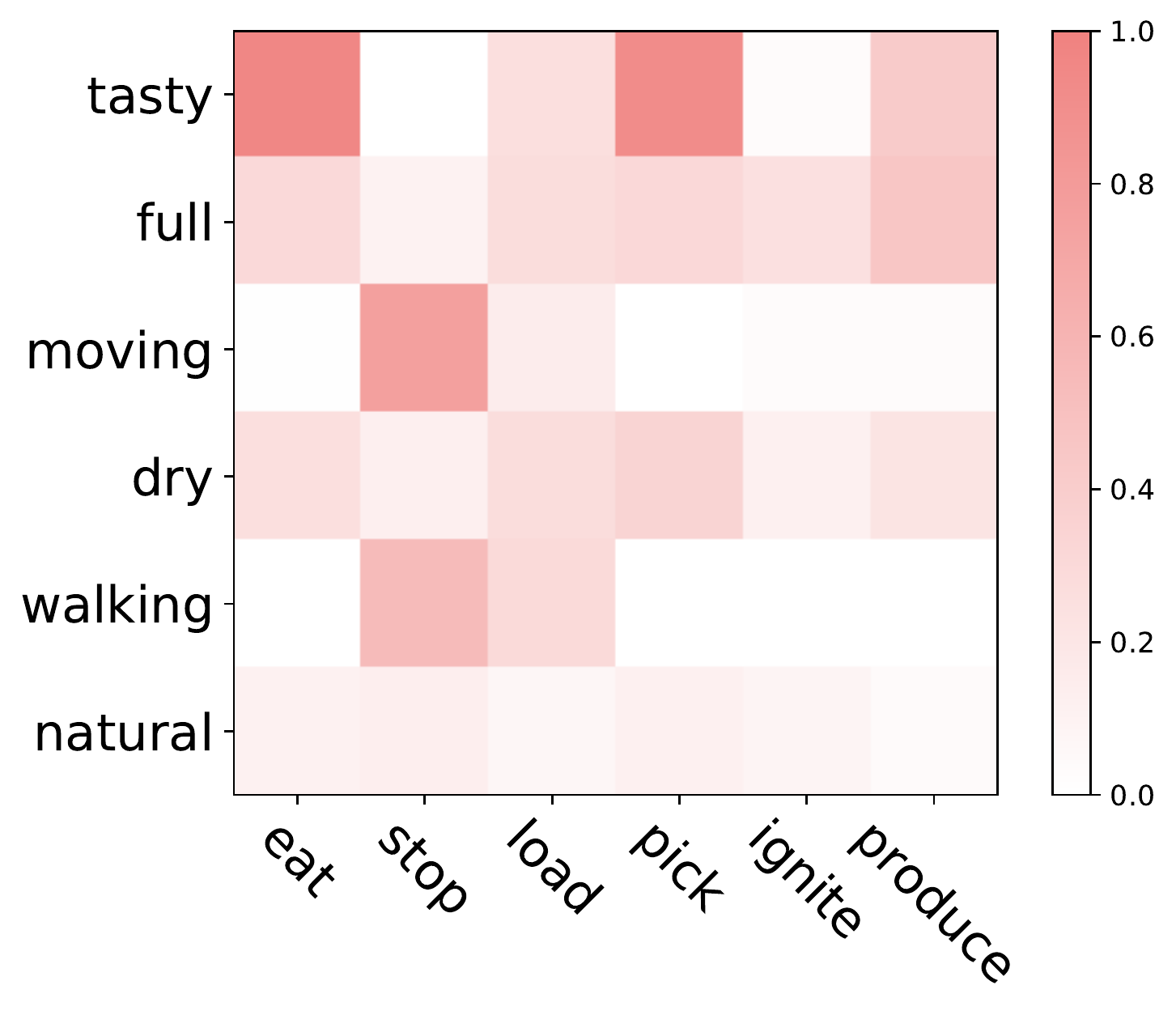}
    \caption{Attribute \textbf{conditioned} affordance matrix.}
    \label{fig:condition_matrix}
\end{minipage}
\hfill
\begin{minipage}{0.32\textwidth}
    \centering
    \includegraphics[width=0.99\linewidth]{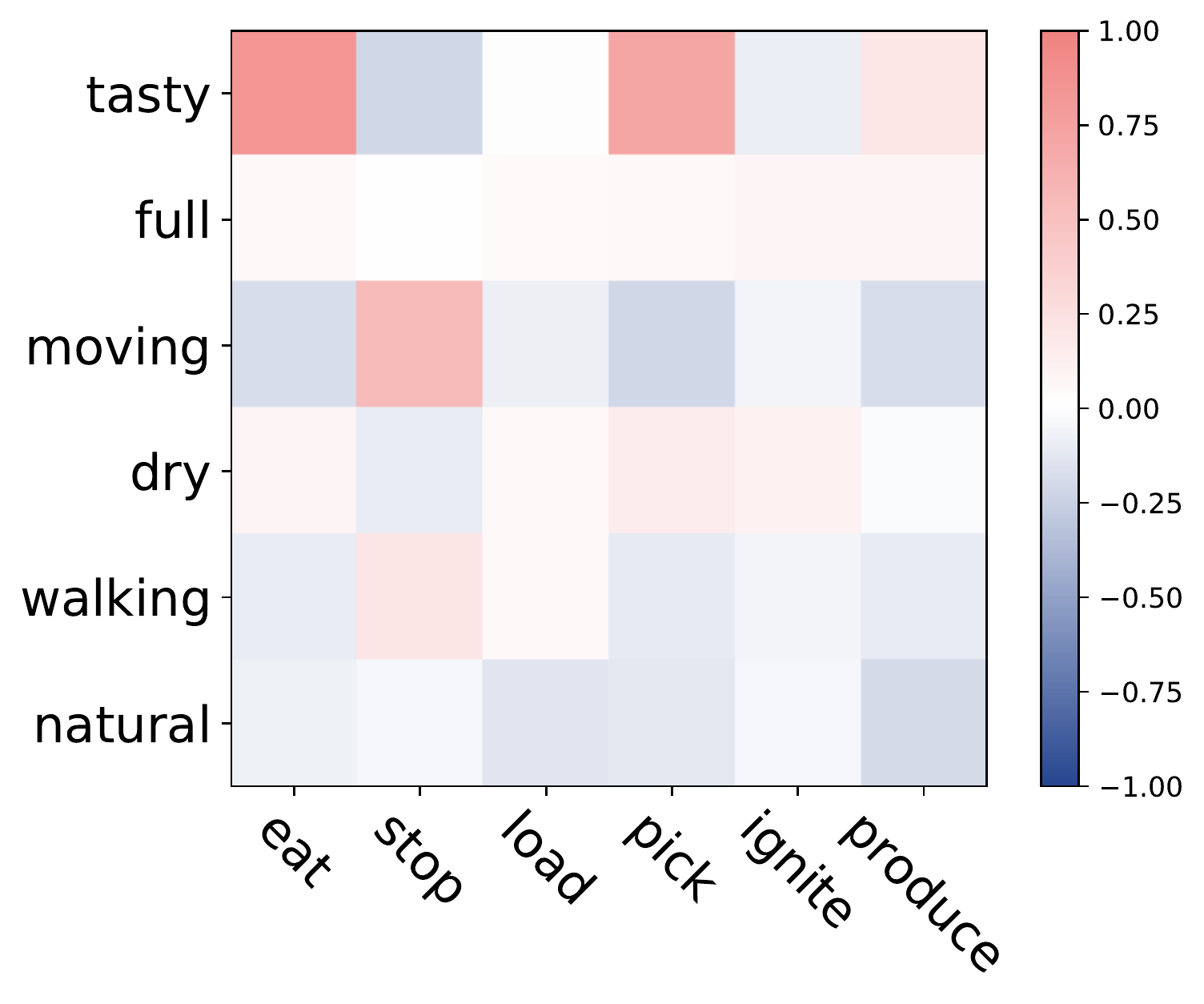}
    \caption{Attribute-affordance \textbf{correlation}.} 
    \label{fig:correlation_matrix}
\end{minipage}
\hfill
\begin{minipage}{0.31\textwidth}
    \centering
    \includegraphics[width=0.99\linewidth]{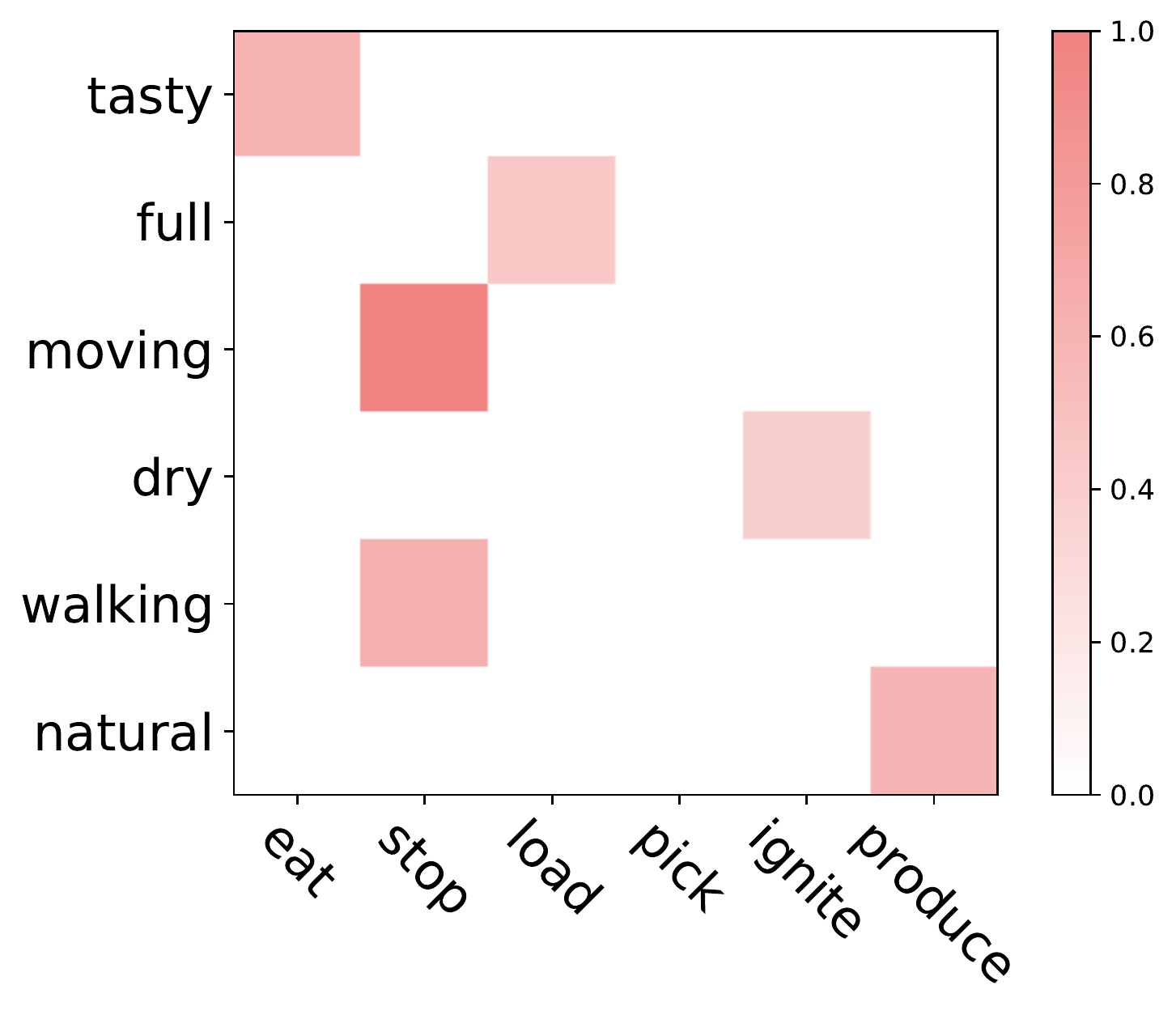}
    \caption{Attribute-affordance \textbf{causality}.} 
    \label{fig:causal_matrix}
\end{minipage}
\end{figure*}

We visualize the samples of attribute-affordance relation matrices in Fig.~\ref{fig:condition_matrix}, \ref{fig:correlation_matrix}, \ref{fig:causal_matrix} and observe some interesting properties of them. They reveal some common relations, such as what is between \textit{tasty} and \textit{eat}. However, some of the criteria suffer from data bias. 
For the condition matrix in Fig.~\ref{fig:condition_matrix}, it only cares about cases with \textit{positive} attribute labels, which is not good in highlighting the negative relations, \textit{e.g.}, the relation between \textit{natural} and \textit{produce}. For the former two matrices in Fig.~\ref{fig:condition_matrix}, \ref{fig:correlation_matrix}, they all point out the relation between \textit{tasty} and \textit{pick}, since most \textit{tasty} objects are \textit{pickable food}. This finding is simply misled by the data bias but violates the causal graph (inference from attribute to object category, then affordance). 
Last, the matrix obtained from our causal annotation in Fig.~\ref{fig:causal_matrix} is more sparse and clear of causality.

\begin{figure}[!ht]
	\begin{center}
        \begin{minipage}{.49\textwidth}
            \centerline{\includegraphics[width=\linewidth]{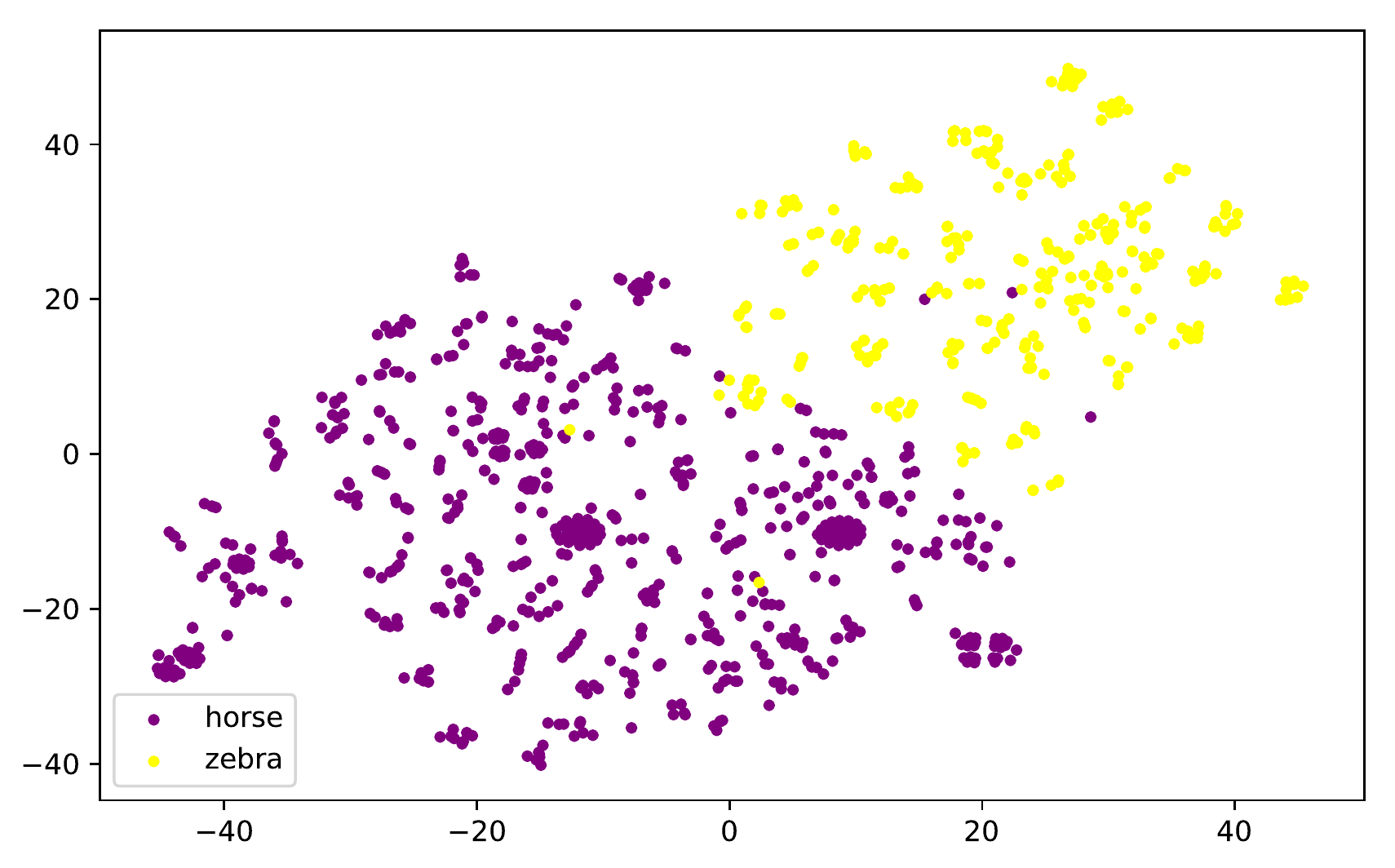}}
            \centerline{\small (a) Attribute Labels.} 
        \end{minipage}
        \hfill
        \begin{minipage}{.49\textwidth}
            \centerline{\includegraphics[width=\linewidth]{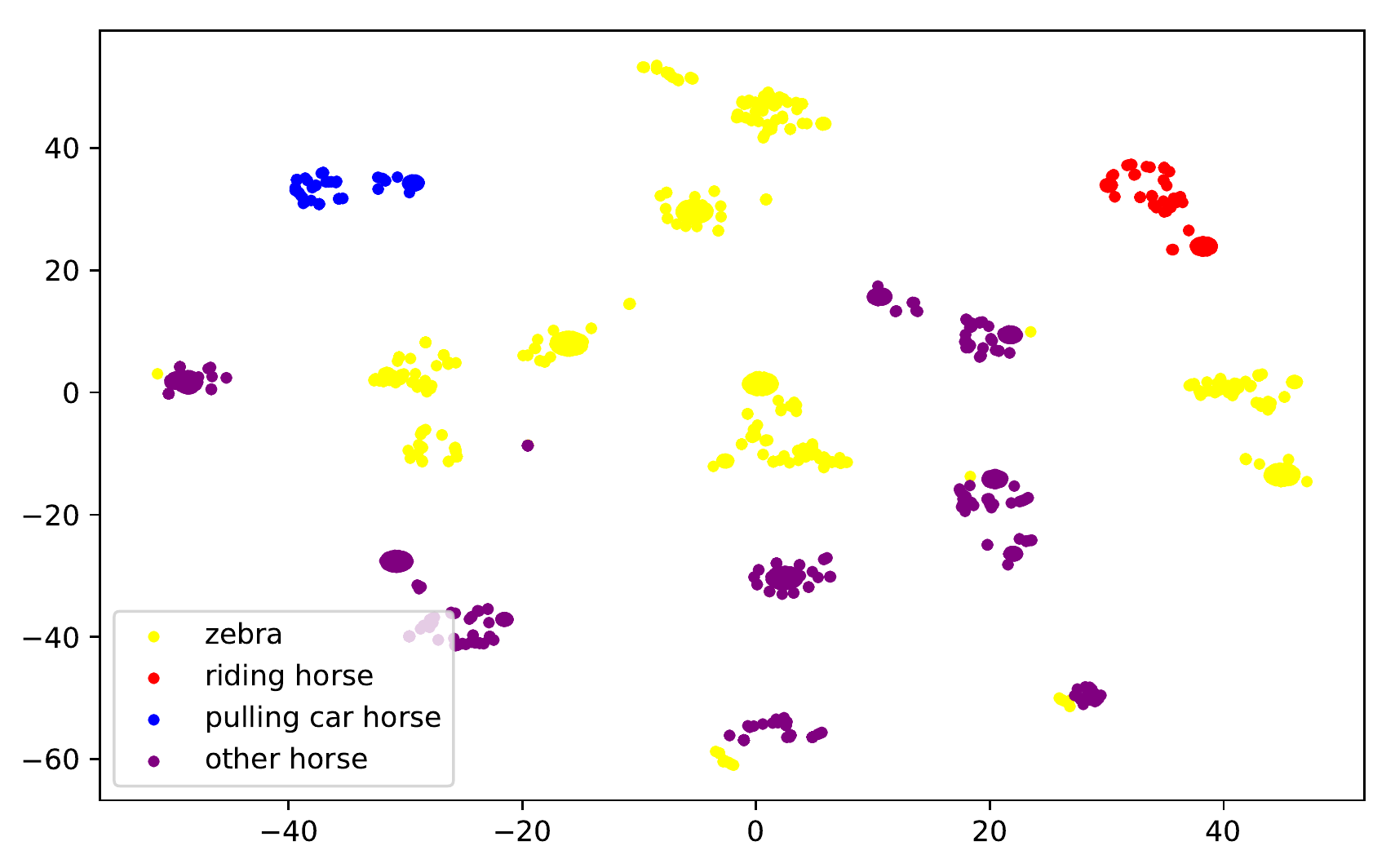}}
            \centerline{\small (b) Attribute-Affordance Labels.}
        \end{minipage}
	\end{center}
	\caption{Clustering using attribute and attribute-affordance labels.}
	\label{Figure:tsne_label}
\vspace{-5px}
\end{figure}

\subsection{Unified Object Representation} 
To compare the difference between attribute-only and attribute-affordance representations, we cluster the object instances of two similar animals (\texttt{zebra} and \texttt{horse}) with their attribute labels and attribute-affordance labels, respectively. 
The results are shown in Fig.~\ref{Figure:tsne_label} via t-SNE~\cite{tsne}. 
With both attribute and affordance labels, zebra and horse can be better separated than attribute only. And attribute and affordance together can differentiate specific \textbf{states} well, such as \texttt{riding}, \texttt{pulling car}, etc.

\subsection{Difference between Category- and Instance-Level Labels}
\begin{figure}[!ht]
	\begin{center}
        \begin{minipage}{.49\textwidth}
            \centerline{\includegraphics[width=\linewidth]{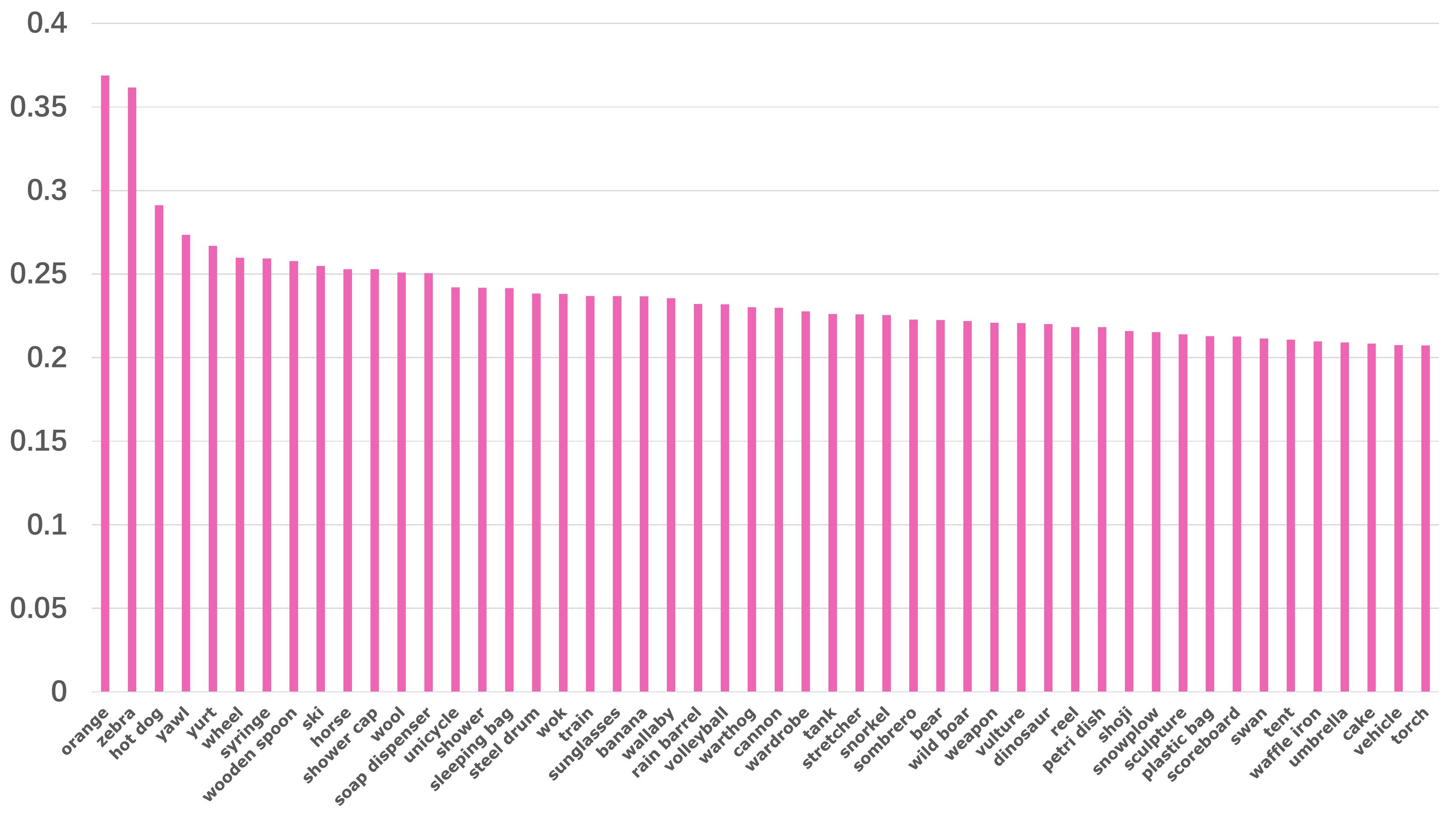}}
            \centerline{\small (a) Difference between $A$ and $\alpha$ labels.} 
        \end{minipage}
        \hfill
        \begin{minipage}{.49\textwidth}
            \centerline{\includegraphics[width=\linewidth]{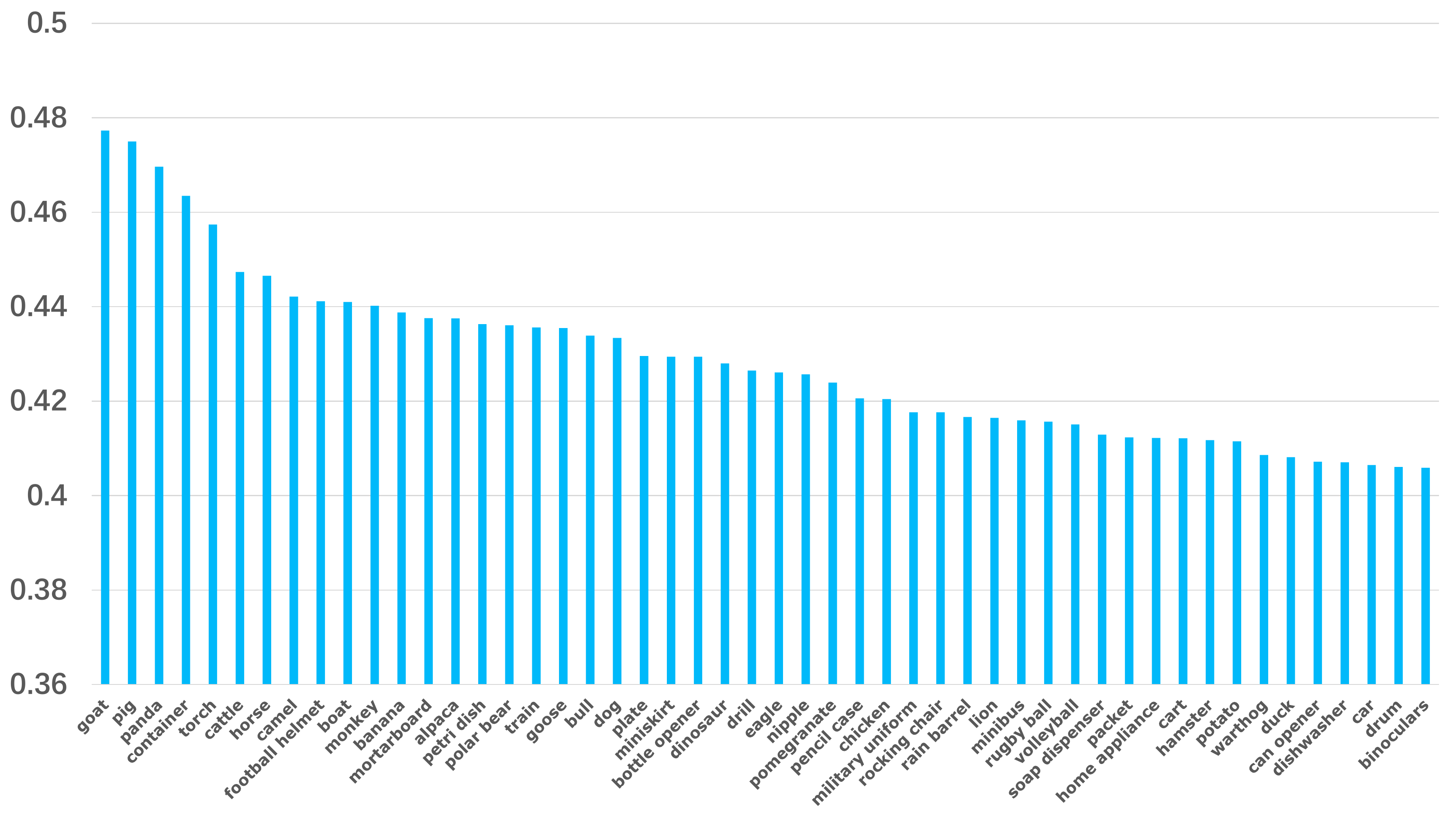}}
            \centerline{\small (b) Difference between $B$ and $\beta$ labels.} 
        \end{minipage}
	\end{center}
	\caption{Top-50 object categories with the largest ratio of the difference between category- and instance-level labels.} 
	\label{Figure:diff}
\end{figure}

We analyze the differences between category-level $A, B$ labels and instance-level $\alpha, \beta$ labels. 
For each object category, we compute the \textit{average ratio} of changed attribute/affordance classes during each instantiation from $A$ to $\alpha$ or from $B$ to $\beta$.
The top 50 categories with the most significant differences between $A$ and $\alpha$ as well as $B$ and $\beta$ are reported respectively in Fig.~\ref{Figure:diff}. 
We find that affordance labels change more dramatically than attribute labels during instantiations. 
This is because \textbf{each} attribute change may affect \textbf{several} affordances, \textit{e.g.}, when a common \texttt{book} becomes \texttt{burn}ing, we can neither \texttt{open} nor \texttt{read} it.

\subsection{Attribute-Affordance Causal Relations}
We annotate all object instances' causal relations of filtered [$\alpha_p, \beta_q$] pairs. 
In total, 1,085 [$\alpha_p, \beta_q$] pairs are chosen for the causality annotation, and over 2 M \textit{instance-$\alpha$-$\beta$} triplets are annotated.
In the ITE evaluation (main text Sec.~5), we report the mean AP of top-300 [$\alpha_p, \beta_q$] pairs to avoid the biased influence of very rare [$\alpha_p, \beta_q$] pairs that include less than 35 object instances.

\subsection{Data Partitioning}
For the OCL task, our knowledge base is split into the train, val, and test sets. The statistical details of the split are listed in Tab.~\ref{table:split}.
The image number ratio of the three sets is nearly 4:1:0.6, and the instance ratio is around 5:1:1.

\begin{table}[ht]
    \begin{center}
        \begin{tabular}{l|ccc}
        \hline 
        Set & Image & Object Instance & Object category  \\
        \hline
        Train    & 56,916 & 135,148 & 381   \\
        Val      & 14,446 & 25,176 & 221  \\
        Test     &  9,101 & 25,617 & 221  \\
        Val+Test & 23,547 & 50,793 & 221  \\
        \hline
        All      & 80,463 & 185,941 & 381 \\
        \hline
        \end{tabular}
        \caption{Detailed data split of our knowledge base.}
        \label{table:split}
    \end{center}
    \vspace{-10px}
\end{table}
    
\subsection{Images and Instances}
\begin{figure*}[ht]
    \centering
    \includegraphics[width=0.7\textwidth]{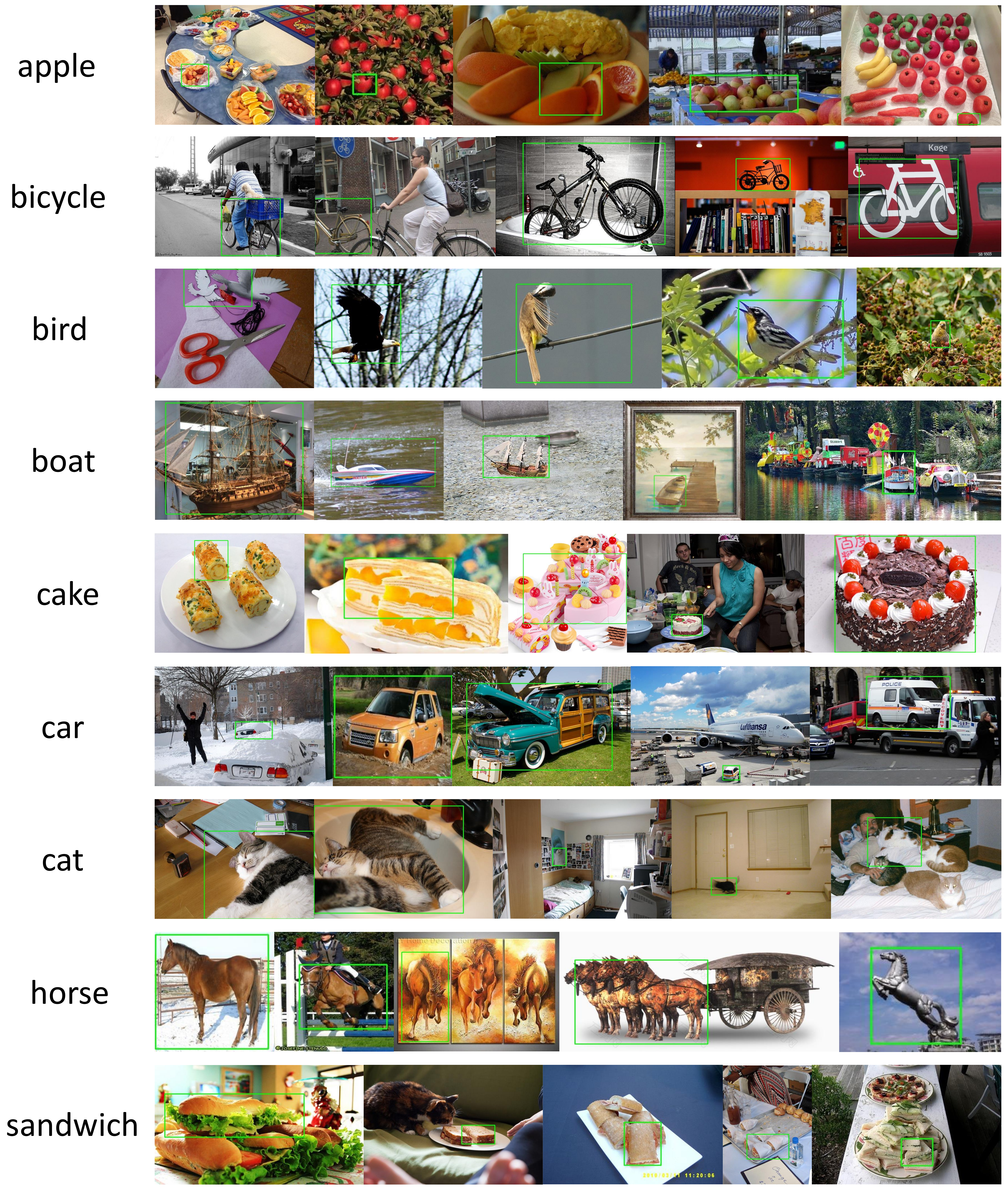}
    \vspace{-0px}
    \caption{More OCL samples of object categories.}
    \label{fig:OCL_obj_sample}
\end{figure*}

\begin{figure*}[ht]
    \centering
    \begin{subfigure} {0.45\textwidth}
        \includegraphics[width=0.95\textwidth]{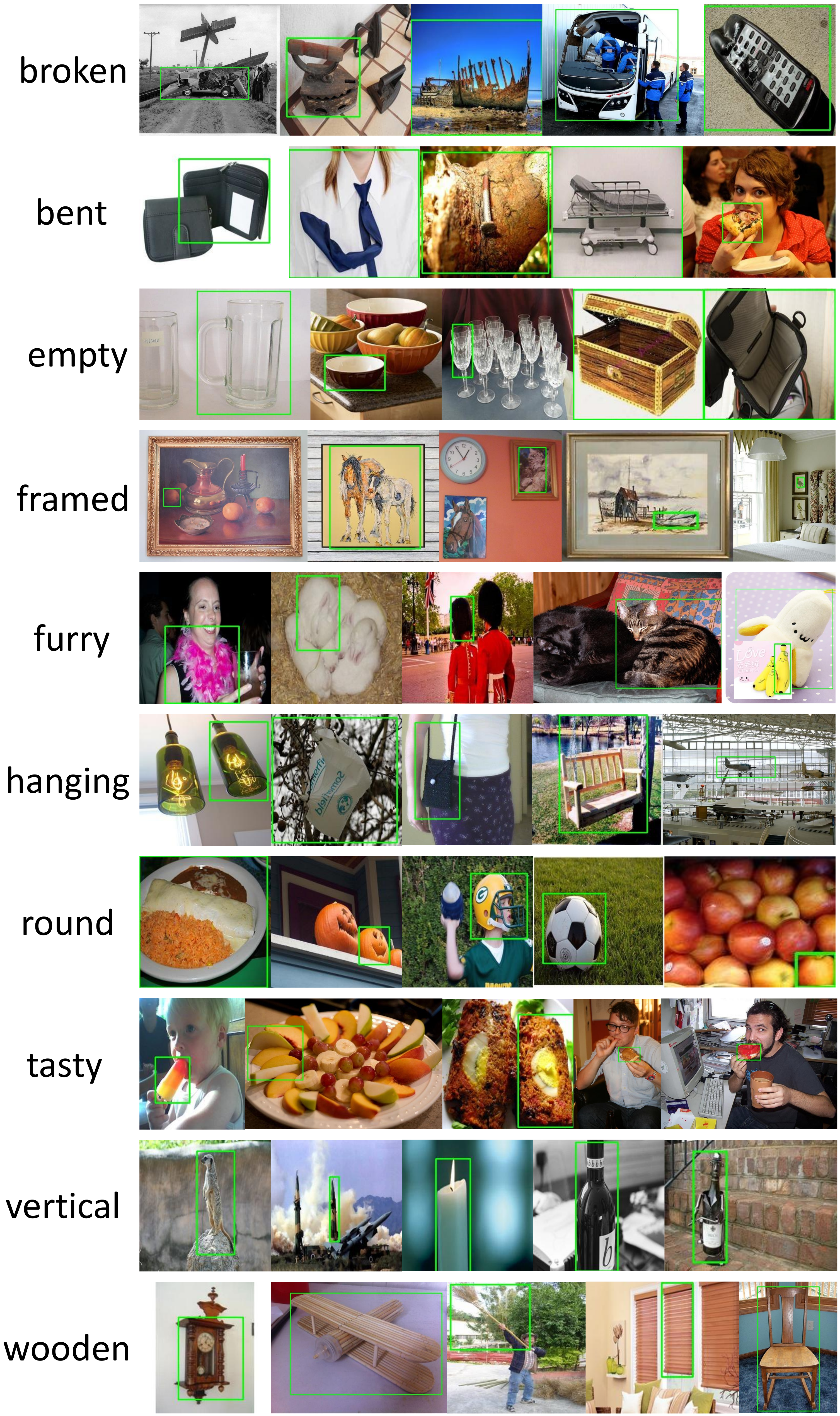}
        \caption{Attributes.}
        \label{fig:OCL_attr_sample}
    \end{subfigure}
    \begin{subfigure} {0.45\textwidth}
        \includegraphics[width=\textwidth]{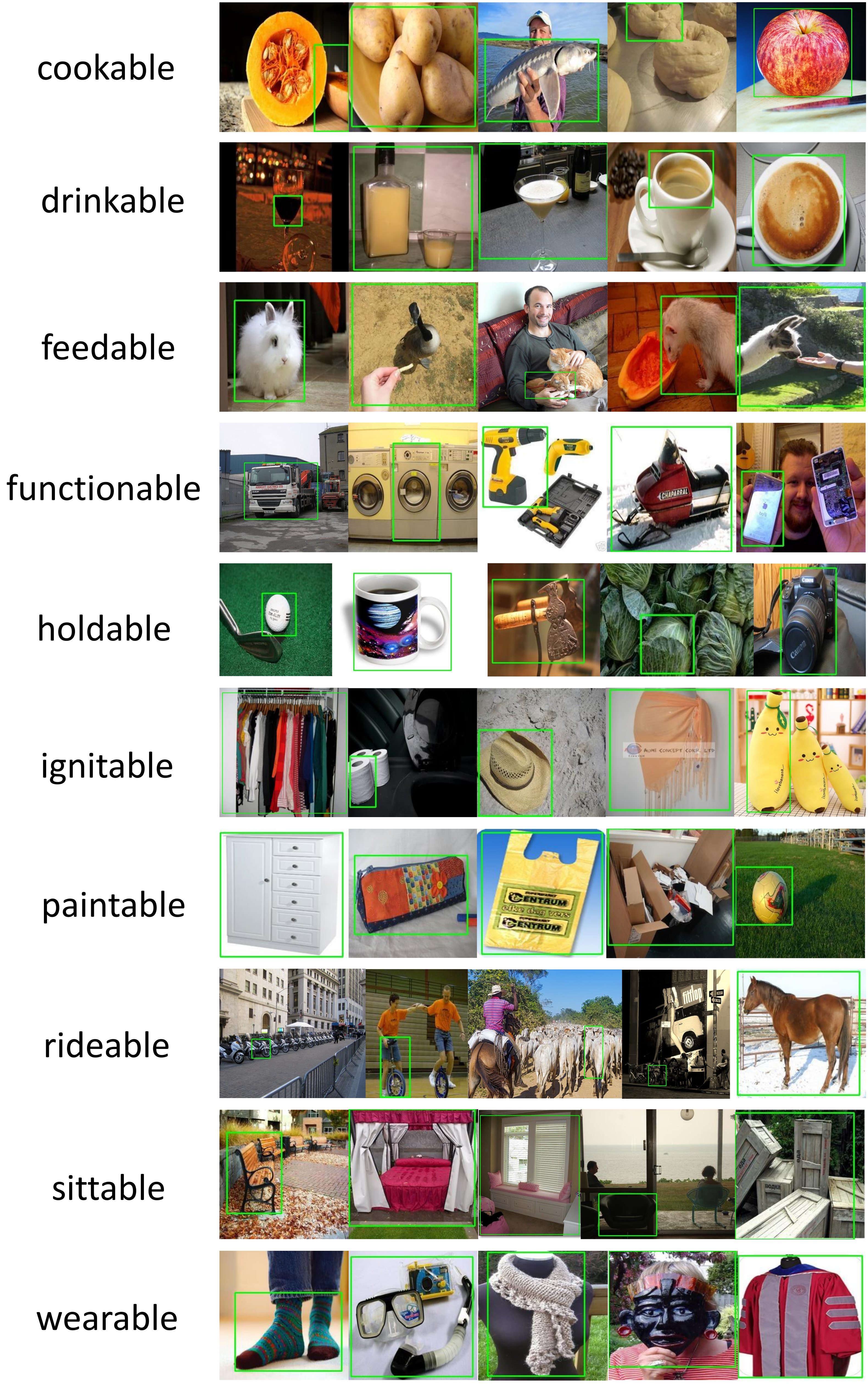}
        \caption{Affordances.}
        \label{fig:OCL_aff_sample}
    \end{subfigure}
    \vspace{-5px}
    \caption{More OCL samples of attributes and affordances.}
\end{figure*}

\begin{figure*}[ht]
    \centering
    \includegraphics[width=0.8\textwidth]{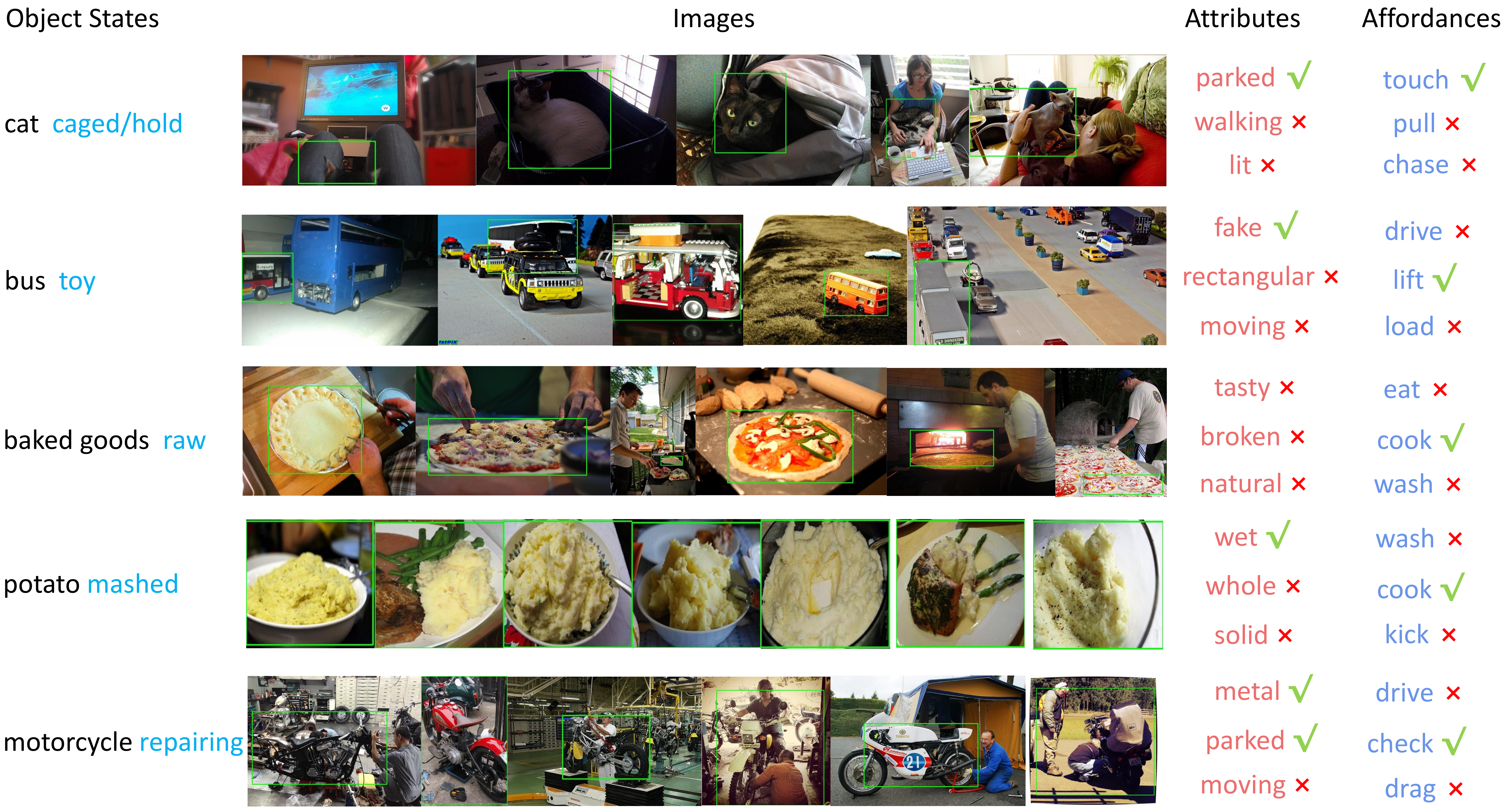}
    \vspace{-10px}
    \caption{More OCL samples. We present objects in different states, together with their key attributes and affordances.}
    \label{fig:OCL_state_sample}
\end{figure*}

\begin{figure*}[ht]
    \centering
    \includegraphics[width=0.6\textwidth]{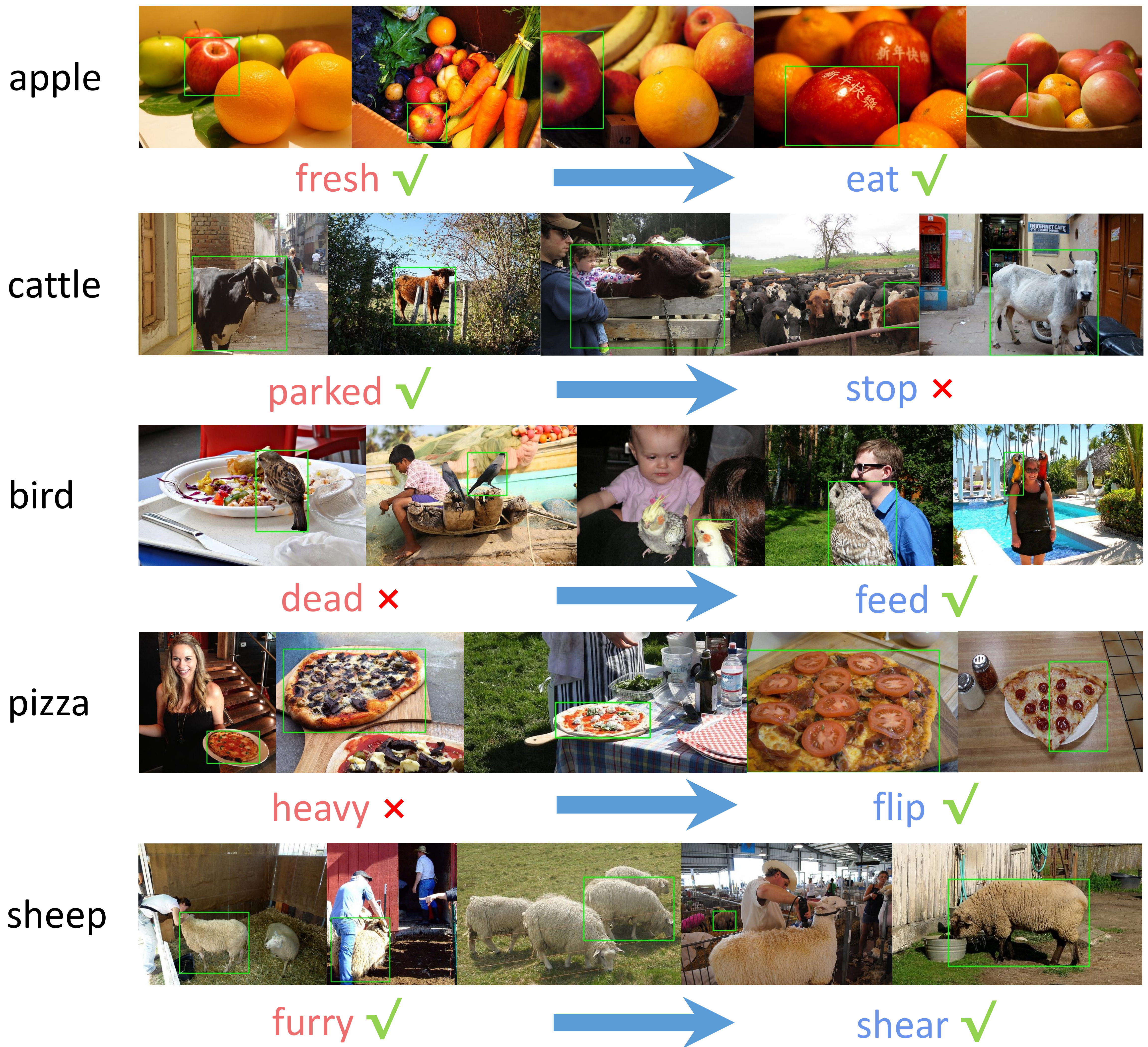}
    \vspace{-10px}
    \caption{More OCL samples of causal relations.}
    \label{fig:OCL_causal_sample}
\end{figure*}

\begin{figure*}[ht]
    \centering
    \includegraphics[width=0.8\textwidth]{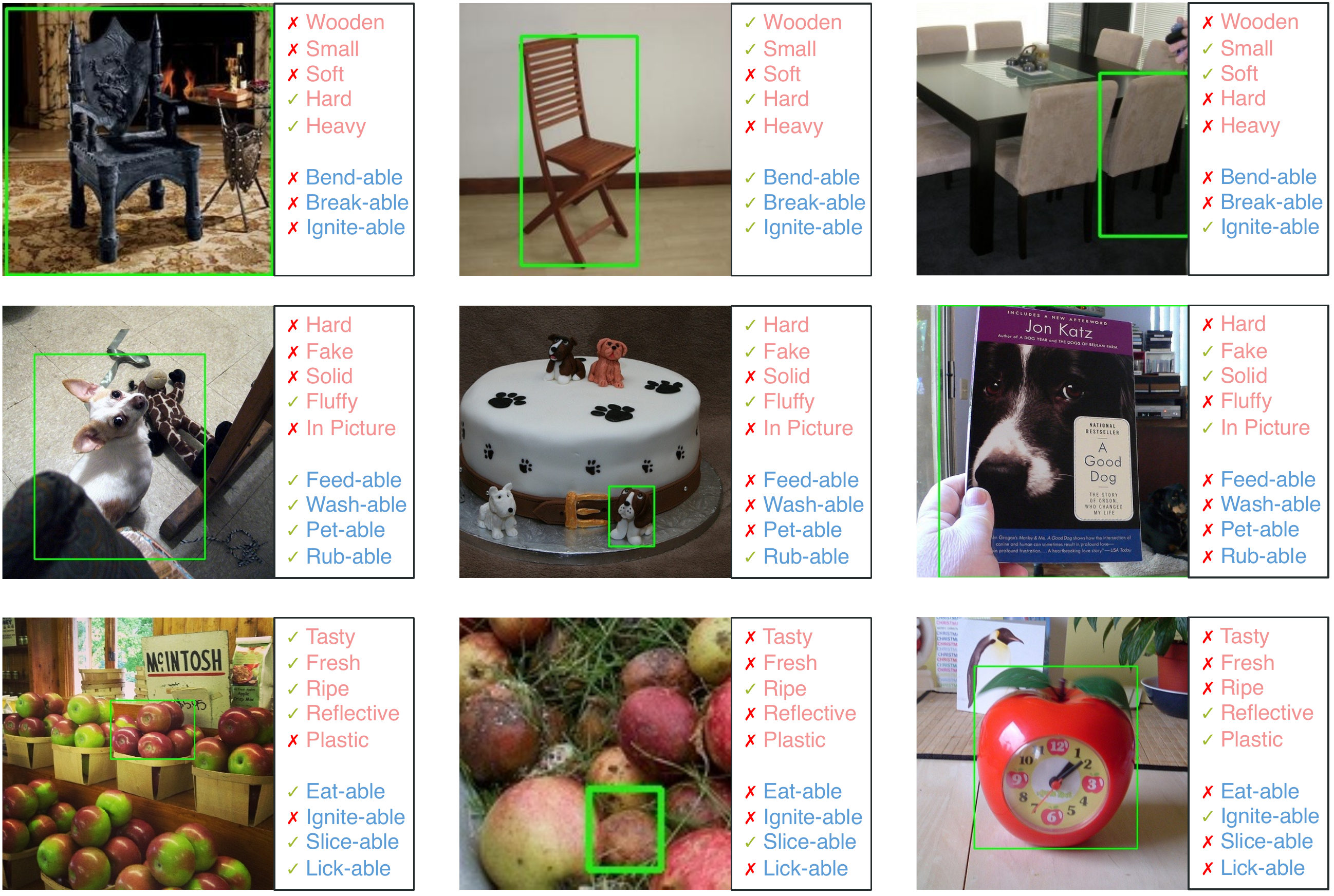}
    \vspace{-10px}
    \caption{More OCL samples in the same category but different states.}
    \label{fig:OCL_diff_state_sample}
\end{figure*}

\begin{figure*}[!ht]
    \centering
    \includegraphics[width=0.95\textwidth]{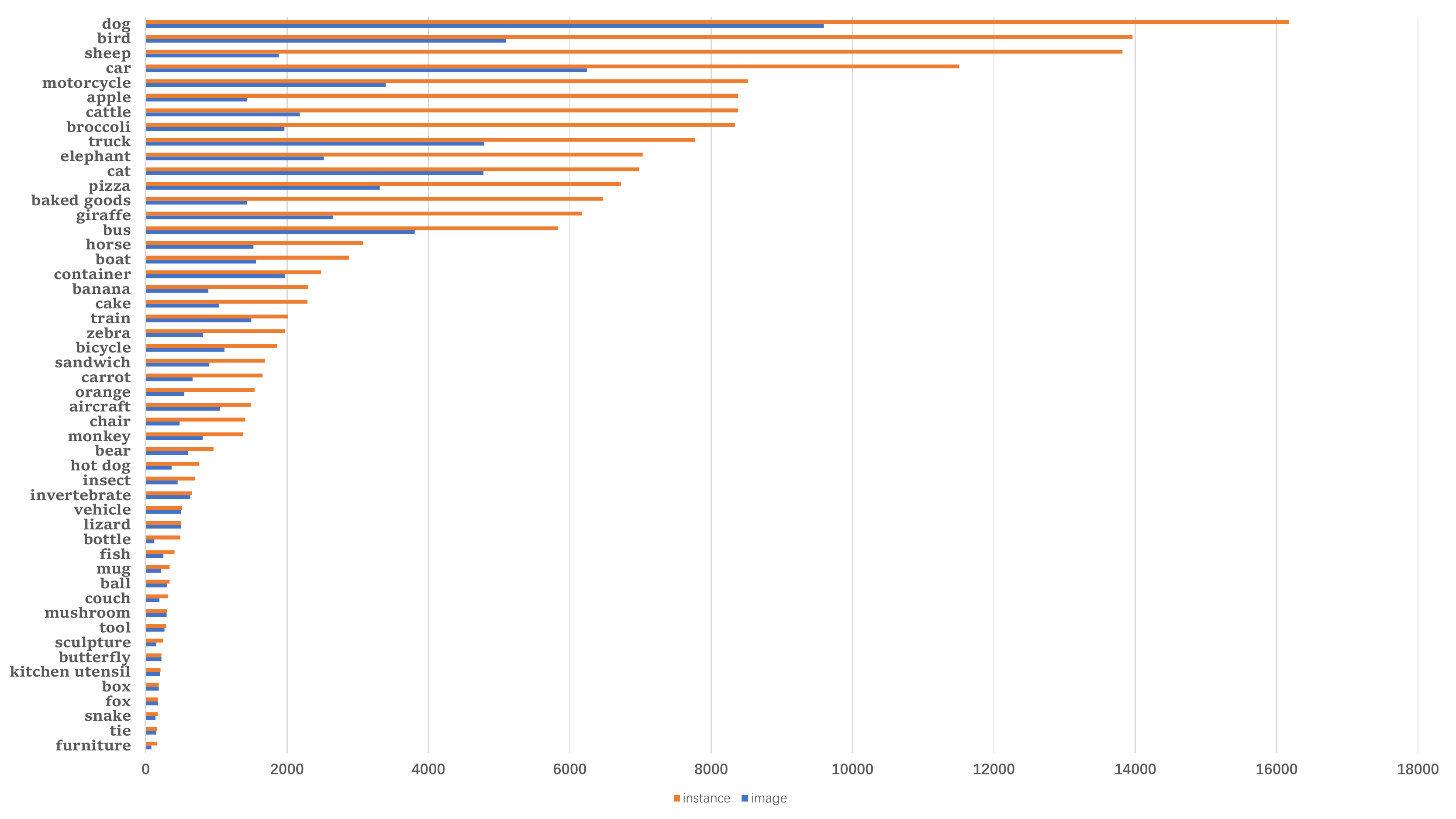}
    \vspace{-13px}
    \caption{Counts of object categories.}
    \label{fig:obj-counts}
    \vspace{-10px}
\end{figure*}

\begin{figure*}[!ht]
    \centering
    \includegraphics[width=0.95\textwidth]{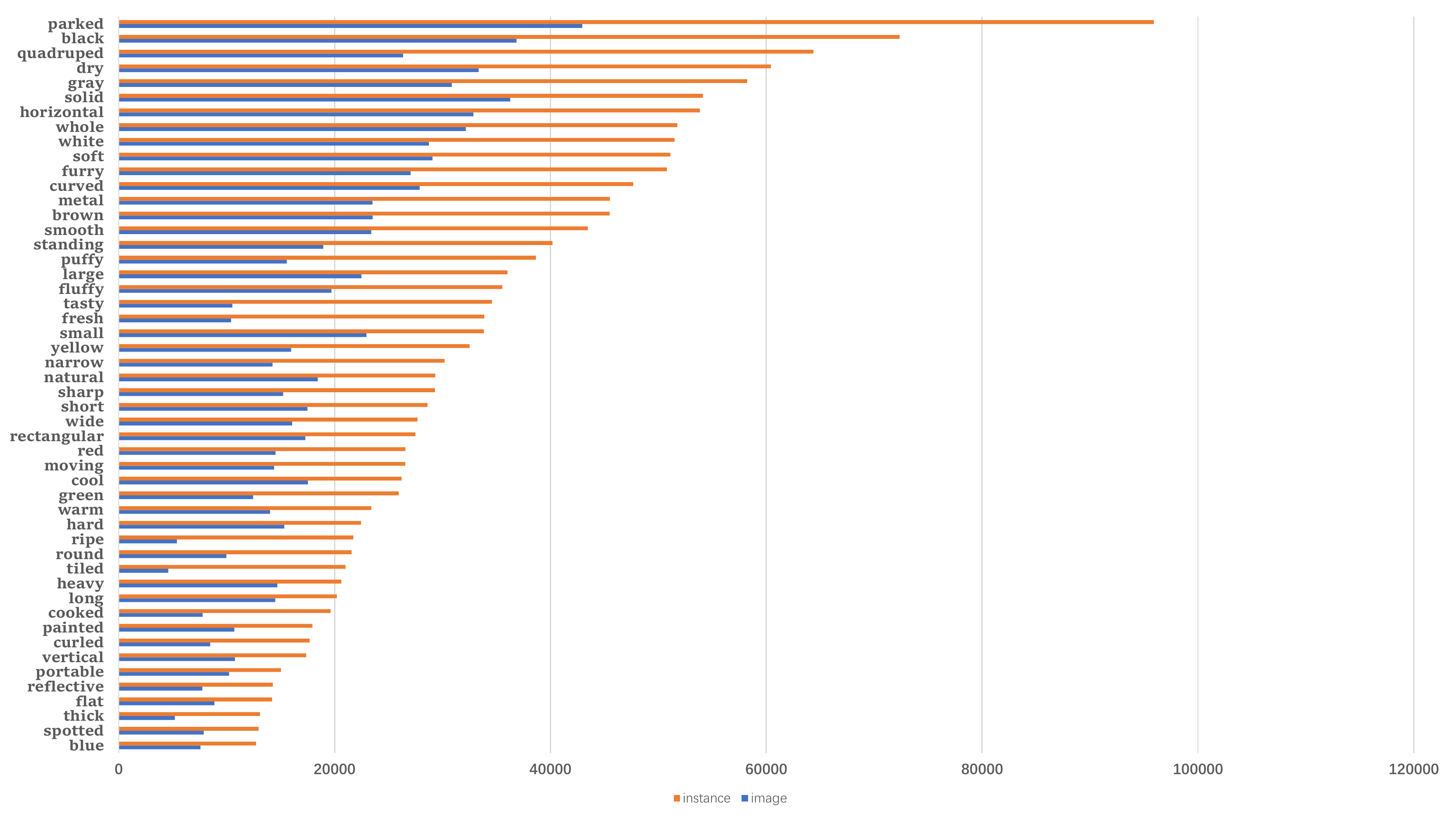}
    \vspace{-13px}
    \caption{Counts of attribute classes.}
    \label{fig:attr-counts}
    \vspace{-10px}
\end{figure*}

\begin{figure*}[!ht]
    \centering
    \includegraphics[width=0.95\textwidth]{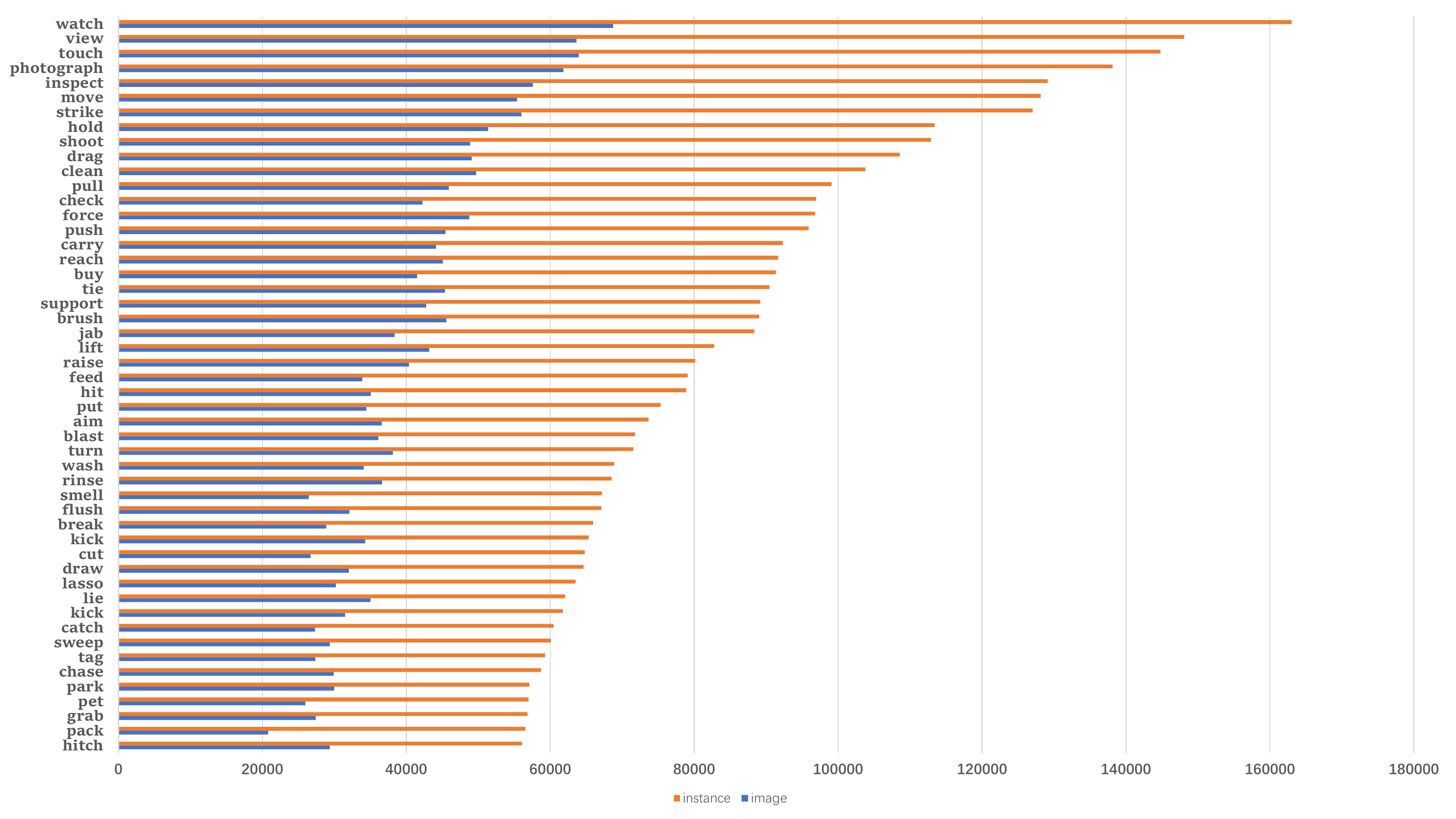}
    \vspace{-13px}
    \caption{Counts of affordance classes.}
    \label{fig:aff-counts}
    \vspace{-10px}
\end{figure*}

Some additional data samples of our knowledge base are shown in Fig.~\ref{fig:OCL_obj_sample}, \ref{fig:OCL_attr_sample}, \ref{fig:OCL_aff_sample}, \ref{fig:OCL_state_sample}, \ref{fig:OCL_causal_sample}, and \ref{fig:OCL_diff_state_sample}, including samples of diverse object categories with various bounding box distributions, different attributes and affordances, and human-labeled object states and obvious causal relations.
We also show the counts of object categories, attributes, and affordances in instance/image in Fig.~\ref{fig:obj-counts},~\ref{fig:attr-counts}, and~\ref{fig:aff-counts}.

\subsection{More Statistics of Annotation}
We divide $A, B, \alpha, \beta,$ causality annotation into multiple finer-grained small sets in our pipeline. Generally, we have 13, 19, 124, 140, and 85 annotator sets (381 total) for $A, B, \alpha, \beta$, and causality annotation respectively. We assign each small set to 2 annotators. However, considering the controversial situations introduced, part of the annotation are confused cases based on their results. In the whole process, 9.6\% of $A$, 7.7\% of $B$, 5.2\% of $\alpha$, 7.9\% of $\beta$, and 13.7\% of causality are confusing and re-assigned to additional annotators. These indeterminable ones will be sent to two extra annotators until agreement. The quality of the dataset is guaranteed by a low confusion ratio and multiple refining stages.

\subsection{Potential Bias}
We have considered the bias issue in the construction of our dataset.
(1) In our dataset, the existing datasets (ImageNet~\cite{imagenet}, COCO~\cite{coco}, aPY~\cite{apy}, SUN~\cite{sun}) are open-sourced datasets and the images collected from the Internet are publicly accessible too. The dataset is constructed for only non-commercial purposes. We will only provide the URLs of these images to avoid copyright infringement.
(2) During image collection, we choose images with general objects and are particularly careful with the image selection to avoid unsuitable content, private images, or implicit biases.
(3) During annotation, the annotators cover different genders, ages, and fields of expertise to avoid potential annotation biases. And they are all informed on how we will use the annotations in our research.

\section{ITE Metric Details}
\label{sec:detailed-tde-metric}
ITE (\textbf{Individual Treatment Effect (ITE)}~\cite{rubin2005causal}) is to measure whether a model infers affordance with proper attention to the causality-related attribute. That said, when removing the attribute, the model is expected to have \textit{large prediction difference further away from the ground truth}.

We detail some settings in our ITE metric. For the ITE score:
\begin{equation}
    \begin{aligned}
        \mathcal{S}_\text{ITE}=\left\{
            \begin{aligned}
                & \max(\Delta\hat{\beta_q},~0), & \beta_q=1, \\
                & \max(-\Delta\hat{\beta_q},~0), & \beta_q=0,
    \end{aligned}
    \right.
    \end{aligned}
\end{equation}
where
\begin{equation}
    \Delta\hat{\beta_q}=\hat{\beta_q}|_{do(\alpha_p)}-\hat{\beta_q}|_{do(\bcancel{\alpha_p})}==\hat{\beta_q}-\hat{\beta_q}|_{do(\bcancel{\alpha_p})},
\end{equation}
we want the moving direction of affordance prediction after the intervention to be correct according to the GT affordance labels ($\beta_q$). 
Concretely, for an instance with the labeled causal relation between [$\alpha_p, \beta_q$],
if the label $\beta_q=1$, we expect the prediction change $\Delta\hat{\beta_q}$ to be larger, indicating the elimination of $\alpha_p$ leads to a drop of predicted probability. Because without the effect of $\alpha_p$, the probability of $\beta_q$ should be \textbf{contrary} to the fact (${\beta_q}=1$).
Similarly, if $\beta_q=0$, we expect $\Delta\hat{\beta_q}$ to be smaller, i.e. the elimination of $\alpha_p$ leads to an increase of predicted probability.
The design of the ITE loss also follows the setting of this ITE score.

In $\alpha$-$\beta$-ITE, the ITE score is multiplied by two factors of recognition performance:

\begin{equation}
    \begin{aligned}
        P(\hat{\alpha_p}=\alpha_p) &= \left\{
            \begin{aligned}
                & \hat{\alpha_p}, & \alpha_p=1, \\
                & 1-\hat{\alpha_p}, & \alpha_p=0,
            \end{aligned}
        \right. \\
        P(\hat{\beta_q}=\beta_q) &= \left\{
            \begin{aligned}
                & \hat{\beta_q}, & \beta_q=1, \\
                & 1-\hat{\beta_q}, & \beta_q=0.
            \end{aligned}
        \right.
    \end{aligned}
\end{equation}
And the overall metric is:
\begin{equation}
    \mathcal{S}_{\alpha\text{-}\beta\text{-ITE}}=\mathcal{S}_\text{ITE}P(\hat{\alpha_p}=\alpha_p)P(\hat{\beta_q}=\beta_q)
\end{equation}
The factors measure the correctness of attributes and affordances. Hence a model achieves a high $\mathcal{S}_{\alpha\text{-}\beta\text{-ITE}}$ only if it correctly predicts attribute and affordance and learns the causal relation between them.

In our experiments, for attribute/affordance recognition only, all methods adopt labels to learn knowledge from the data. In the evaluation of causal relation, only the ``w/ $L_{ITE}$'' models adopt the causal relation labels. We hope the models can automatically learn to mine and learn the intrinsic causalities. Thus, we design the ITE to evaluate this ability. 
Similar to our OCRN, some works~\cite{vcrcnn,tang2020long,tang2020unbiased} also try to marry supervised deep learning and causal inference.

\section{Baseline Details} 
\label{sec:baseline-detail}

We introduce the details of all baselines here:

{\bf Fold \uppercase\expandafter{\romannumeral 1}.} 
No arc between $\alpha$ and $\beta$.

{\bf (1) Direct Mapping from Visual Feature (DM-V)}: feeding $f_I$ into MLP-Sigmoids to predict $P_{\alpha}, P_{\beta}$. Each $\alpha$ and $\beta$ class owns customized MLP followed by LayerNorm~\cite{layernorm} to generate class-specific features and share the same MLP-Sigmoid in classification.

{\bf (2) DM from Linguistic Representation (DM-L)}: replacing the input representation $f_I$ of DM-V with linguistic feature $f_L$, which is the expectation of Bert~\cite{bert} of category names w.r.t $P(O_i|I)$.

{\bf (3) Multi-Modality (MM)}: mapping $f_I$ to the semantic space via minimizing the distance to its $f_L$. The multi-modal aligned $f_I$ is fed to an MLP-Sigmoids to predict $P_{\alpha}, P_{\beta}$.

{\bf (4) Linguistic Correlation (LingCorr)}: measuring the correlation between object and $\alpha$/$\beta$ classes via their Bert~\cite{bert} cosine similarity. $P_{\alpha}$, $P_{\beta}$ are given by multiplying $P(O|I)$ to correlation matrices.

{\bf (5) Kernelized Probabilistic Matrix Factorization (KPMF)~\cite{kpmf}}: calculating the Softmax normalized cosine similarity between each testing instance and all training samples as weights. Then $P_{\alpha}$ or $P_{\beta}$ is generated as the weighted sum of GT $\alpha$ or $\beta$ of training samples.

{\bf (6) $\mathbf{A\&B}$ Lookup}: returning the expectation of category-level attribute or affordance vectors $A_i,B_i$ w.r.t $P(O_i|I)$.
In detail, seen category probabilities are obtained from GT prior $M_A, M_B$. 
Unseen category probabilities are voted by the top 3 most similar seen categories according to the cosine similarity of category Word2Vec~\cite{word2vec} vectors. 
Then, we generate category-level attribute and affordance matrices $M'_{A}, M'_{B}$ given the GT prior (seen) and similarity-based probabilities (unseen).
Finally, we multiply $P(O|I)$ with $M'_{A}, M'_{B}$ to predict $P_{A}, P_{B}$ and assign them to $P_{\alpha},P_{\beta}$ respectively. 

{\bf (7) Hierarchical Mapping (HMa)}: first mapping $f_I$ to category-level attribute or affordance space by an MLP supervised by GT $A$ or $B$. Then the mapped features are fed to an MLP-Sigmoids to predict $P_{\alpha}$ or $P_{\beta}$.

{\bf Fold \uppercase\expandafter{\romannumeral 2}.}
Directed arc from $\beta$ to $\alpha$.

{\bf (8) DM from $\beta$ to $\alpha$ (DM-$\beta\rightarrow\alpha$)}: training a $\beta$ classifier with $f_I$ same with DM-V, but using the concatenated representation of affordance as $f_{\beta}$ to train the $\alpha$ classifier.

{\bf (9) DM from $\beta$ and $I$ to $\alpha$ (DM-$\beta I \rightarrow\alpha$)}: training a $\beta$ classifier with $f_I$ same with DM-V, but using the concatenated representation of attributes $f_{\beta}$ and objects $f_{I}$ to train the $\alpha$ classifier.

{\bf Fold \uppercase\expandafter{\romannumeral 3}.} 
Directed arc from $\alpha$ to $\beta$.

{\bf (10) DM from $\alpha$ to $\beta$ (DM-$\alpha\rightarrow\beta$)}: training an $\alpha$ classifier with $f_I$ same with DM-V, but using the concatenated representation of attributes as $f_{\alpha}$ to train the $\beta$ classifier. 

{\bf (11) DM from $\alpha$ and $I$ to $\beta$ (DM-$\alpha I \rightarrow\beta$)}: training an $\alpha$ classifier with $f_I$ same with DM-V, but using the concatenated representation of attributes $f_{\alpha}$ and objects $f_{I}$ to train the $\beta$ classifier.

{\bf (12) Ngram~\cite{ngram}}: adopting Ngram to retrieve the relevance between $\alpha$ and $\beta$ and generating an association matrix $M_{\alpha-\beta}$. Then we multiply DM predicted $P_{\alpha}$ with $M_{\alpha-\beta}$ to estimate $P_{\beta}$.

{\bf (13) Markov Logic Network (MLN-GT)}~\cite{mln}: adopting MLN to model the $\alpha-\beta$ relations following \cite{yuke}. After training on OCL, we infer $\beta$ with \textbf{GT} $\alpha$ to estimate its \textit{performance upper bound}.

{\bf (14) Instantiation with attention (Attention)}: feeding $[f_\alpha, f_I]$ to an MLP-Sigmoid to generate attentions and predicting $P_\beta$ by multiplying the attentions with $P_B$.

We operate baselines with a directed arc from $\alpha$ to $\beta$ (Fold \uppercase\expandafter{\romannumeral 3}) to perform ITE. The ITE calculation needs \textbf{feature zero-masking} to eliminate the effect of specific attributes~\cite{tang2020unbiased}. These methods (DM-At, DM-AtO, Attention, OCRN) follow the same ITE calculation (feature masking).
Two unique cases are Ngram and MLN-GT. 
Ngram uses attribute probabilities to infer affordance. Thus, we randomize the specific attribute probabilities for Ngram to operate the ITE calculation. 
And MLN-GT must use GT attribute labels to distinguish the ``positive'' and ``negative'' causes and then reason out the effect affordance. Thus, in ITE, we directly eliminate its corresponding attribute input.

\section{Detailed Result Analysis}
\label{sec:result-analysis}

\subsection{Detailed Attribute and Affordance Performances}
\begin{figure}[ht]
\centering
\begin{minipage}{0.49\textwidth}
    \centering
    \includegraphics[width=\textwidth]{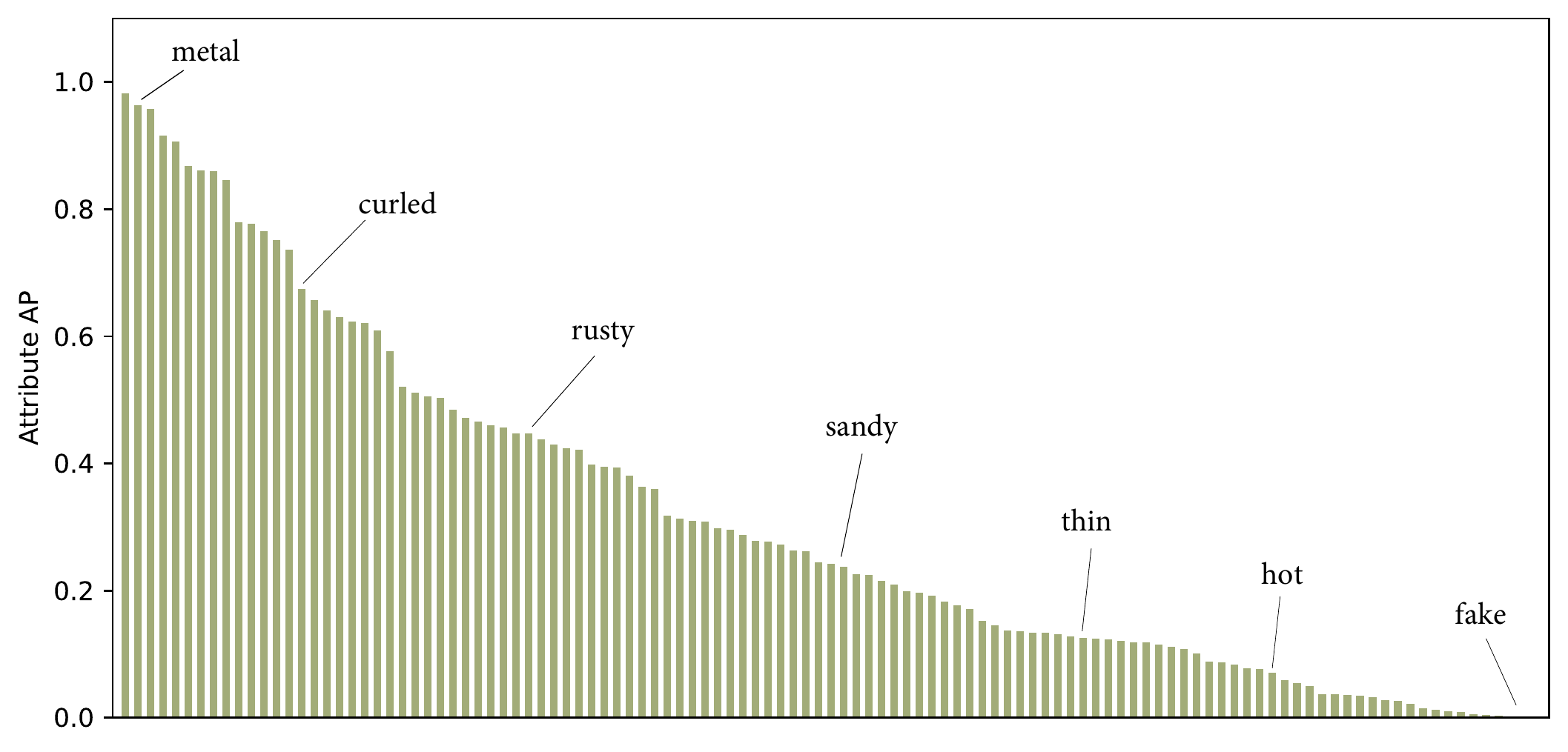}
    \caption{AP of attribute classes.}
    \label{fig:attr-ap}
\end{minipage}
\begin{minipage}{0.49\textwidth}
    \centering
    \includegraphics[width=\textwidth]{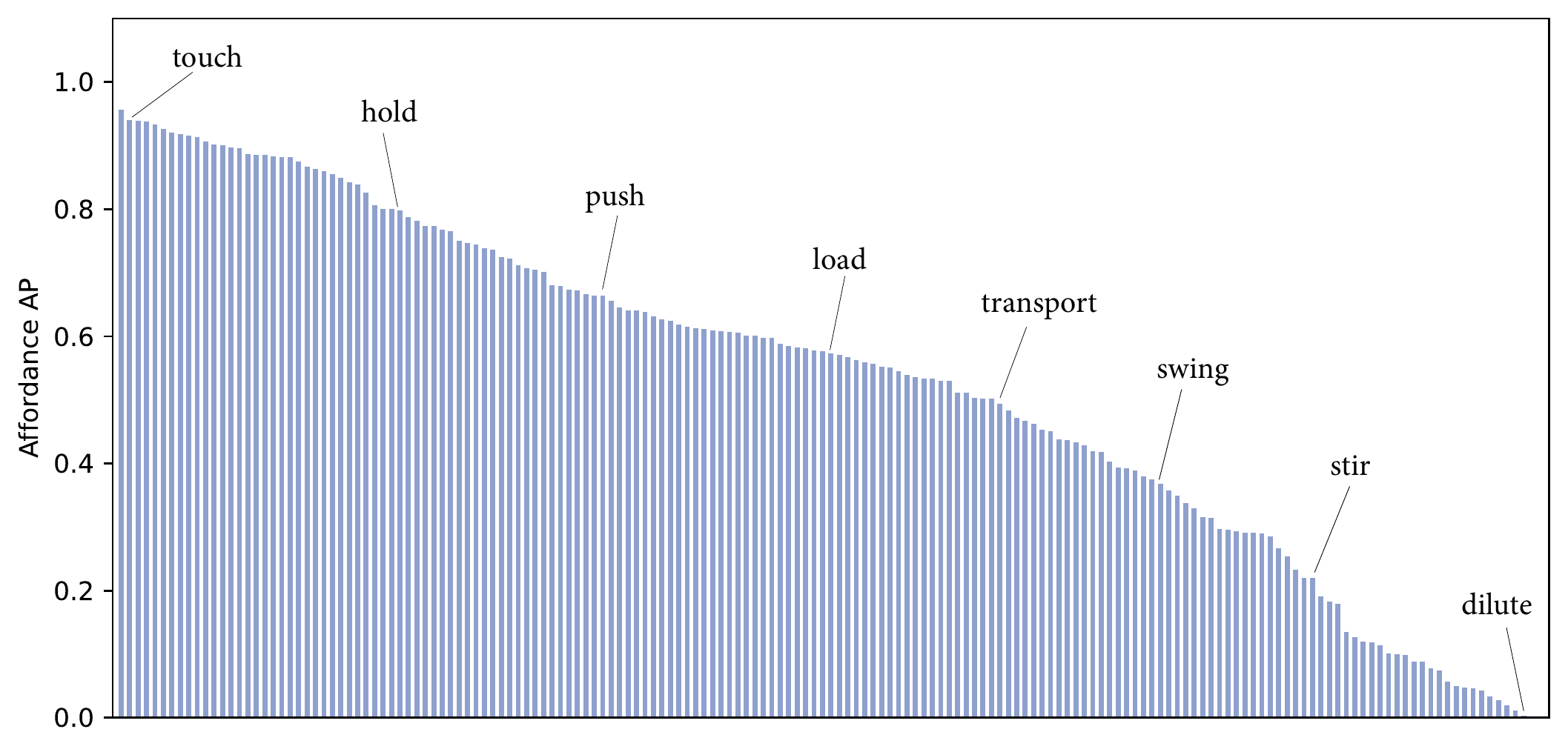}
    \caption{AP of affordance classes.}
    \label{fig:aff-ap}
\end{minipage}
\end{figure}

We compute and analyze the performance (AP) of OCRN on each attribute or affordance class in Fig.~\ref{fig:attr-ap} and Fig.~\ref{fig:aff-ap}, which suggest that visually abstract concepts like \texttt{fake} are more difficult to model than concrete ones like \texttt{metal}, \texttt{breakable}.
The performance of attribute classes is lower than affordance classes. This is mainly because the attributes have more diversity. Thus the \textit{positive} instances of each attribute class are \textbf{less} than the affordance class.

\subsection{Visualization of ITE Result}
\begin{figure}[t]
  \begin{minipage}{0.49\textwidth}
      \centering
      \begin{subfigure}[b]{\textwidth}
          \centering
          \includegraphics[width=\textwidth]{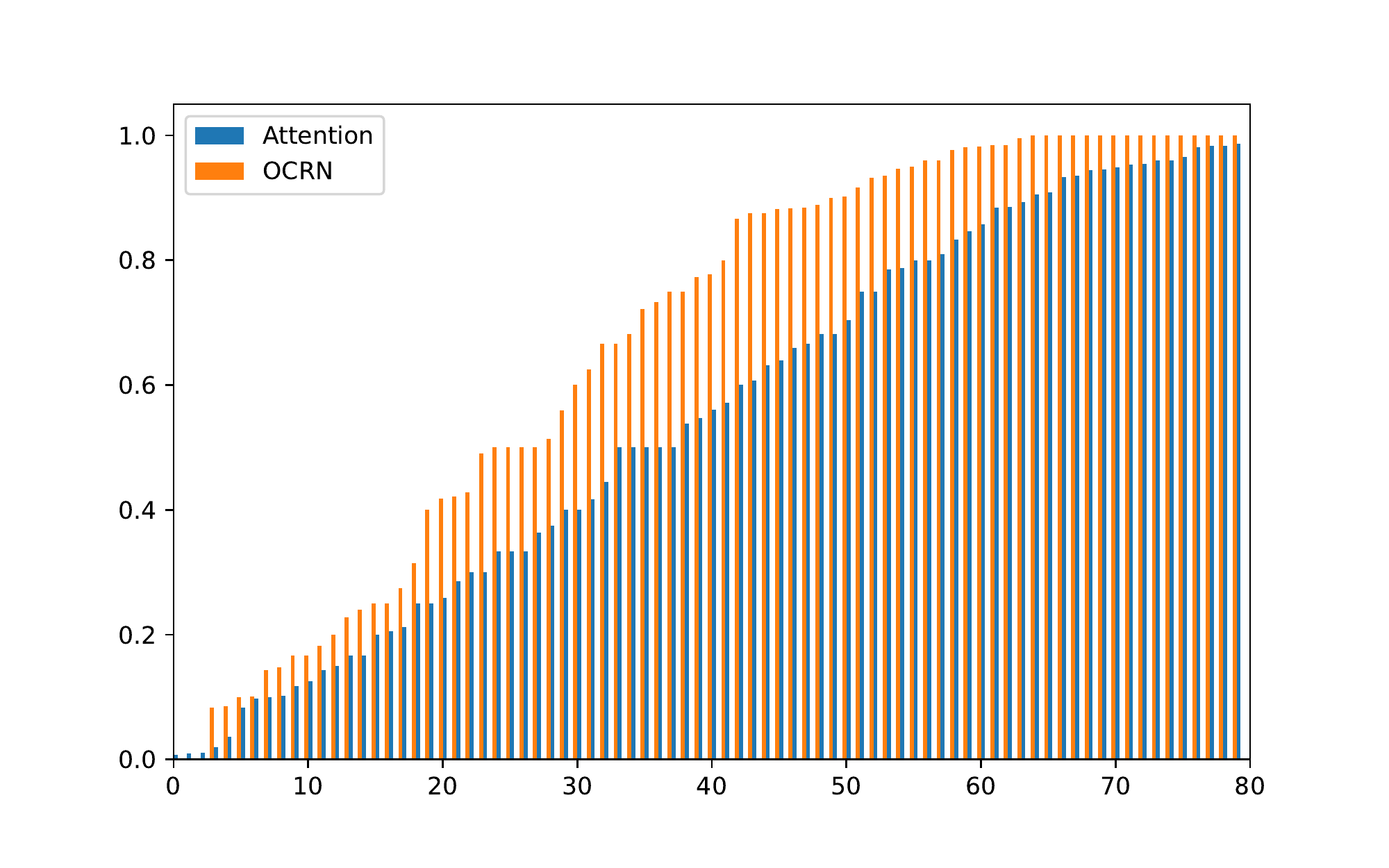}
          \caption{Proportion of correct ITE (ground truth $\beta_q=0$).}
      \end{subfigure}
      \begin{subfigure}[b]{\textwidth}
          \centering
          \includegraphics[width=\textwidth]{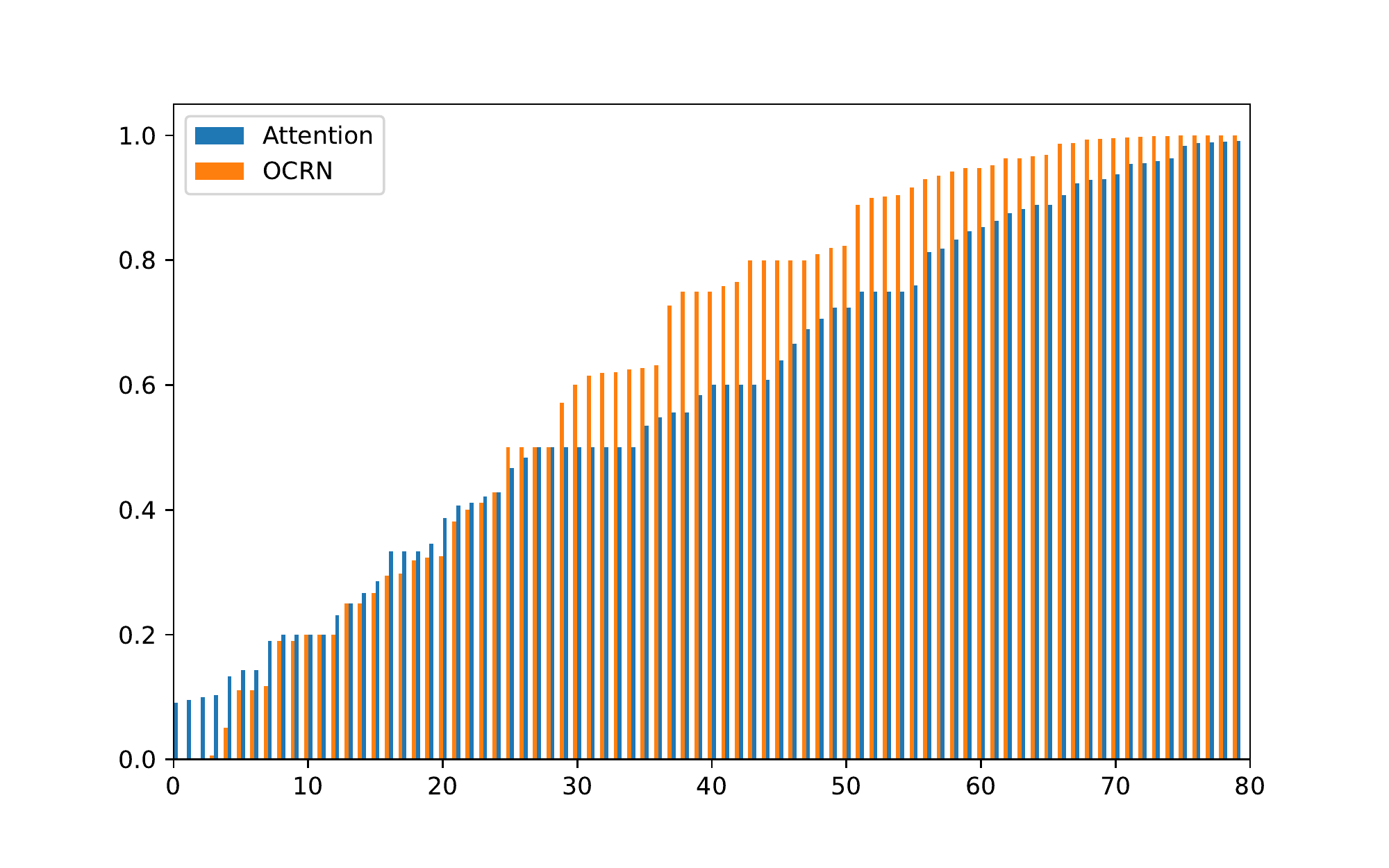}
          \caption{Proportion of correct ITE (ground truth $\beta_q=1$).}
      \end{subfigure}
      \caption{ITE given different [$\alpha_p, \beta_q$].}
      \label{fig:tde-plot}
  \end{minipage}
\end{figure}

In Fig.~\ref{fig:tde-plot}, we show the correct instance proportions ($\%$) of OCRN and Attention after ITE.
(a) randomly chosen causal pairs $[\alpha_p,\beta_q]$ with ground truth $\beta_q=1$, expecting $\hat{\beta_q} > \hat{\beta_q}|_{do(\bcancel{\alpha_p})}$.
(b) randomly chosen causal pairs $[\alpha_p,\beta_q]$ with ground truth $\beta_q=0$, expecting $\hat{\beta_q} < \hat{\beta_q}|_{do(\bcancel{\alpha_p})}$.
The higher proportions indicate that OCRN performs better on ITE.

\subsection{Attribute and Affordance Recognition Given Detected Boxes}
Though OCL is a high-level concept learning task with object boxes as inputs, we can also consider object detection in evaluation for practical applications.
We adopt Swin Transformer (Swin)~\cite{swin} as the detector.
It is pretrained on COCO~\cite{coco} and finetuned on the OCL train set with GT boxes of 381 categories.
On the OCL test set, it achieves 22.9 $AP_{50}$ on object detection.
Subsequently, it will provide detected box $b_o$ for all models in inference.
We can consider the detection effect in the attribute and affordance recognition metric to build a more strict criterion.
Namely, all \textit{false positive} detections (IoU$<$0.3 with referring to GT boxes) as the \textit{false positives} of $\alpha$ and $\beta$ recognition too.
Moreover, ITE calculation needs to construct the counterfactual of an object instance. If the inaccurately detected object box shifts according to the GT box, it is difficult to know whether the counterfactual comes from the attribute masking or visual content change, using the corresponding attribute-affordance causal relation labels of this GT box.
Thus, considering the unique property of causal inference different from common recognition, here we do not report the ITE score.
Tab.~\ref{table:res-main-det} shows the results given detected boxes.
Due to the more strict criterion and detection quality, the performances of all methods degrade greatly. 
But OCRN still holds the superiority on two tracks. 

\begin{table}[ht]
\centering
    \resizebox{0.25\textwidth}{!}{
        \begin{tabular}{l|cc}
        \hline
        Method & $\alpha$ & $\beta$ \\
        \hline
        DM-V            &  \underline{7.4} & \underline{11.0} \\ 
        DM-L            &  4.6 &  9.1 \\ 
        MM              &  5.4 &  9.9 \\ 
        LingCorr        &  1.7 &  5.6 \\ 
        KPMF            &  6.4 & 10.5 \\ 
        $A\&B$-Lookup   &  4.1 &  5.8 \\ 
        HMa             &  6.5 & 10.9 \\ 
        \hline
        DM-At           &  6.8 & 10.5 \\ 
        DM-AtO          &  6.6 & 10.8 \\
        Ngram           &  5.1 & 10.2 \\ 
        MLN-GT          & - & - \\ 
        Attention       &  5.5 & 10.1 \\ 
        OCRN            & \textbf{7.9} & \textbf{11.3} \\
        \hline
        \end{tabular}
    } 
    \caption{Attribute and affordance recognition results given detected boxes from Swin Transformer~\cite{swin}.}
    \label{table:res-main-det}
\end{table}

\subsection{OCL-Based Image Retrieval}
\begin{figure}[ht]
\centering
\begin{minipage}{0.49\textwidth}
    \centering
    \includegraphics[width=\textwidth]{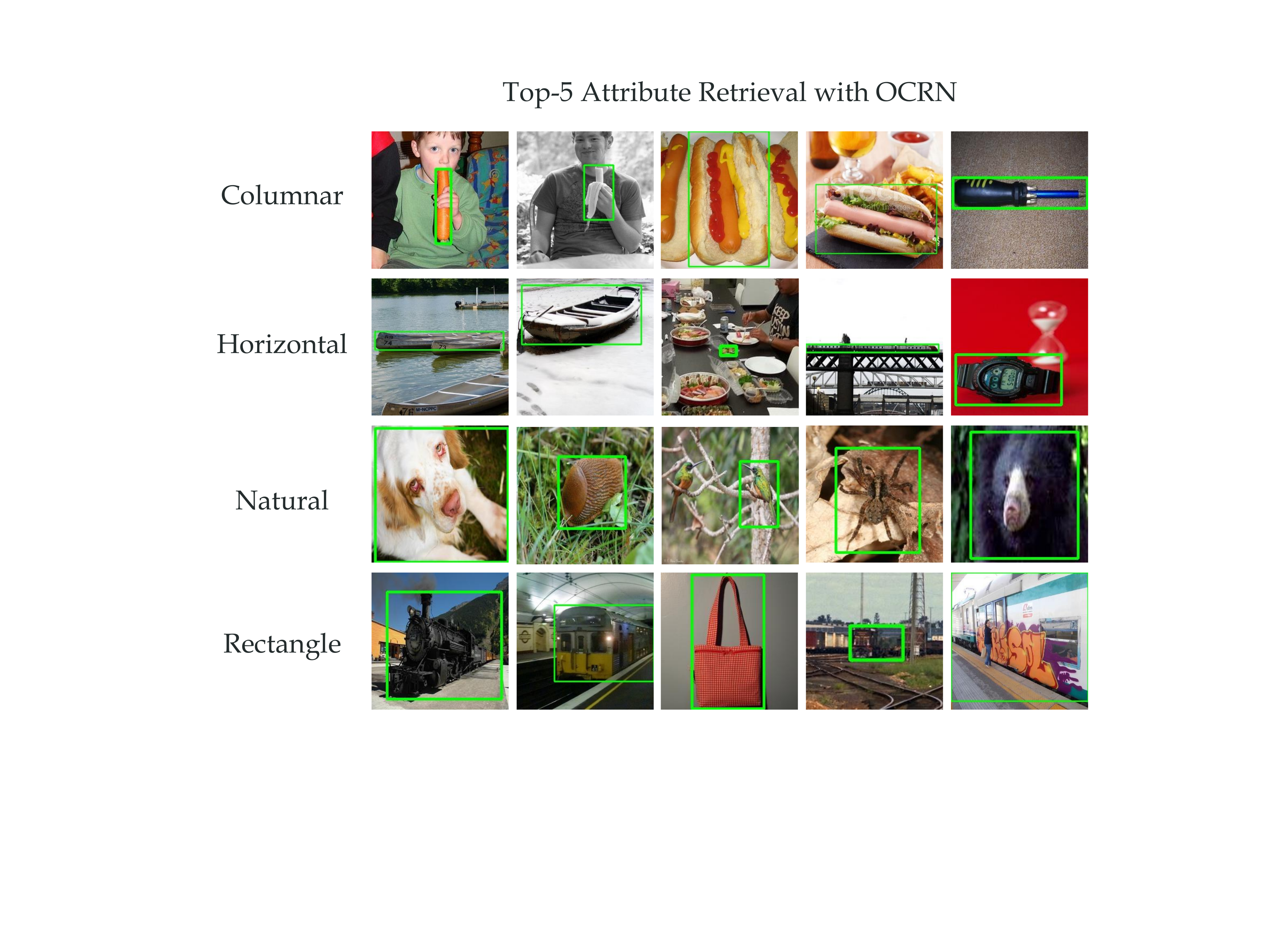}
    \caption{Top-5 attribute retrievals on the OCL test set.}
    \label{fig:det-ablation-attr}
\end{minipage}
\begin{minipage}{0.49\textwidth}
    \centering
    \includegraphics[width=\textwidth]{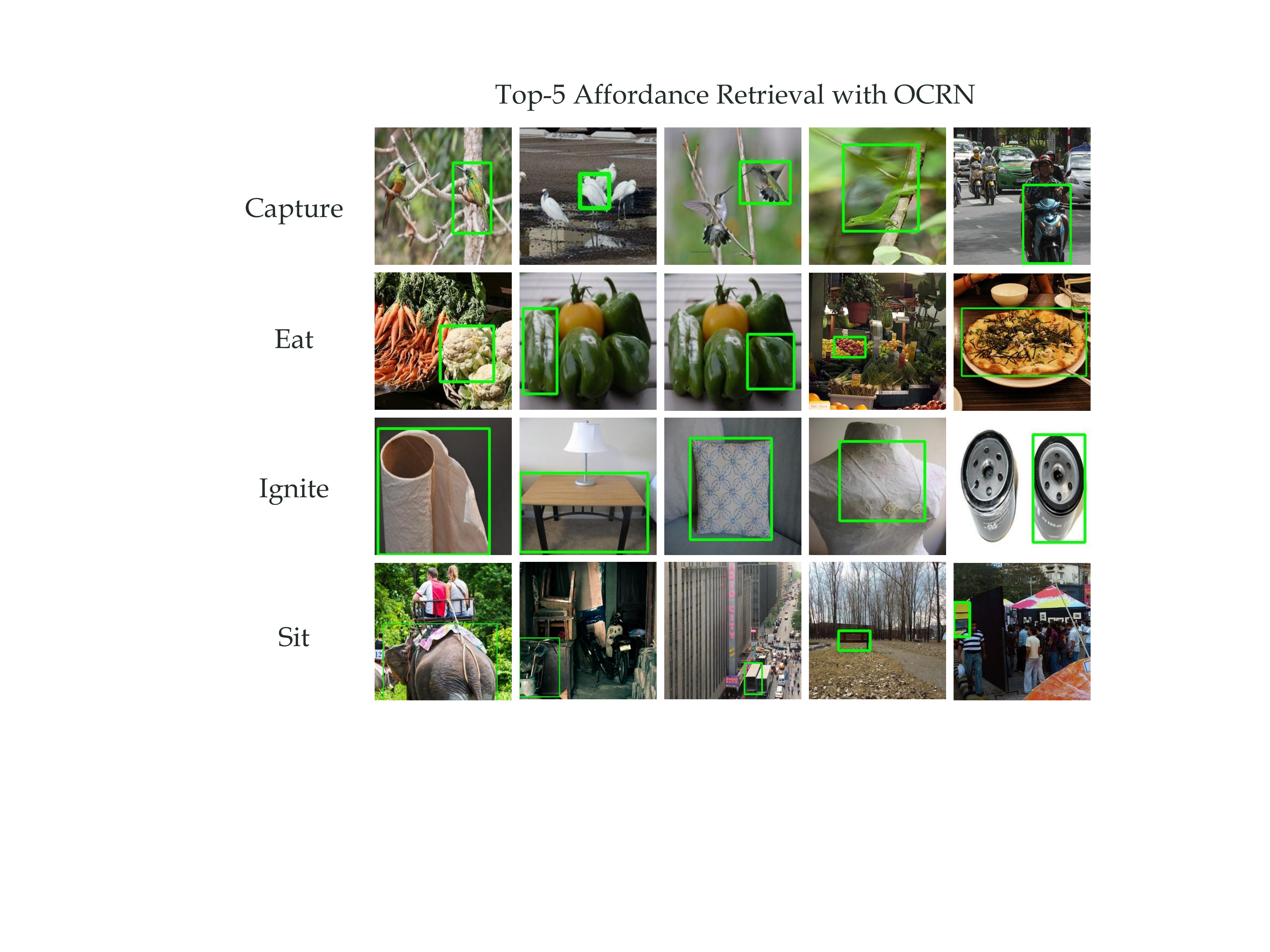}
    \caption{Top-5 affordance retrievals on the OCL test set.}
    \label{fig:visualize}
\end{minipage}
\end{figure}

We visualize the OCL reasoning performance by retrieving the top-score instances with OCRN. Some results are shown in Fig.~\ref{fig:det-ablation-attr} and Fig.~\ref{fig:visualize}. The model can correctly retrieve the related images, especially on some common concepts \textit{e.g.}, \texttt{columnar}, \texttt{sit}.

\section{Application on Human-Object Interaction (HOI) Detection} 
\label{sec:application-hoi}
To further verify the generalization ability, we apply OCL to Human-Object Interaction (HOI) detection~\cite{djrn,liu2022highlighting,pangea} and help HOI methods boost their performances.
HOI detection recently attracts a lot of attention and makes progresses~\cite{hicodet,liu2022interactiveness,tinpami,wu2022mining,dio,dcr,xu2022learning,decaug} thanks to the success of deep learning and large-scale HOI datasets~\cite{hicodet,vcoco,hakev1,hakev2}.

HOI depicts the actions performed upon objects by humans.
Usually, an object has multi-affordance, \textit{i.e.}, a person can perform different actions upon it. 
But in an image, just one or several actions/affordances are usually happening/\textbf{activated}. 
Without object knowledge, previous methods~\cite{tincvpr,ican,idn} can find the activated affordances from hundreds of actions~\cite{hicodet}.
For example, for each human-object pair in HICO-DET~\cite{hicodet}, a model has to select one or several actions from the defined 116 actions. 
With OCL, things are different. OCL covers many actions, so we can use OCRN to infer $P_{\beta}$ of an object to narrow the solution space.
Thus, we propose two ways:

(1) {\bf OCL Filtering}: We use $P_{\beta}$ to narrow the action space with a threshold $\gamma$ and generate $P^{\gamma}_{\beta}$. Affordances with probabilities higher than $\gamma$ are kept and others are set to \textit{zero} ($\gamma=0.5$). 
Then, the HOI model only needs to predict in a narrowed action space. 
In practice, we multiply the prediction $P_{HOI}$ from HOI model with $P^{\gamma}_{\beta}$ element-wisely to obtain the final prediction $P'_{HOI}=P_{HOI}*P^{\gamma}_{\beta}$.

(2) {\bf Human-as-Probe}: Another more straightforward way is to predict HOI via OCL directly. We treat the human paired with the object as a \textbf{probe}. 
Assuming the human feature is $f_h$ and human-object spatial configuration feature is $f_{sp}$ (from \cite{tincvpr,ican}). 
As $P_{\beta}$ indicates all possible affordances, the ongoing actions can be seen as the \textbf{instantiation} of $P_{\beta}$, \textit{i.e.}, they are activated by the ``probe'' $f_h$ and $f_{sp}$. So we use $f_h$ and $f_{sp}$ to generate attention $A_{h+sp}$ via MLP-Sigmoid. Then we operate $P_{\beta}*A_{h+sp}$ and late fusion to get the final prediction $P'_{HOI}=(P_{\beta}*A_{h+sp}+P_{HOI})/2$.

Concretely, we use OCRN to enhance HOI detection models (iCAN~\cite{ican}, TIN~\cite{tincvpr}, IDN~\cite{idn}) on HICO-DET~\cite{hicodet}.
As OCL merely contains 15 object categories in HICO-DET~\cite{hicodet}, the rest 65 object categories are \textbf{unseen}.
We embed OCRN into three HOI models according to OCL filtering and Human-as-Probe, and the public model checkpoints of \cite{ican,tincvpr,idn} are used. 

The results are shown in Tab.~\ref{table:res-hoi}. 
With OCL filtering, iCAN~\cite{ican}, TIN~\cite{tincvpr}, and IDN~\cite{idn} achieve a gain of mAP by 0.65\%, 0.90\%, and 0.77\% respectively. 
The Human-as-Probe is more suitable for HOI detection and contributes a performance boost of 1.50\%, 1.46\%, and 0.98\% to three models. 
These strongly verify the efficacy and generalization ability of OCL.

\begin{table}[ht]
\begin{center}
    \resizebox{0.35\textwidth}{!}{
    \begin{tabular}{l|ccc}
    \hline 
    Method & Full & Rare & Non-Rare \\
    \hline
    iCAN           & 14.84 & 10.45 & 16.15 \\
    iCAN+Filtering & 15.49 &  8.76 & 17.50 \\
    iCAN+Probe     & {\bf 16.34} & {\bf 11.66} & {\bf 17.74} \\
    \hline
    TIN           & 17.03 & 13.42 & 18.11 \\
    TIN+Filtering & 17.93 & 13.79 & 19.17 \\
    TIN+Probe     & {\bf 18.49} & {\bf 15.02} & {\bf 19.58} \\
    \hline
    IDN           & 23.36 & 22.47 & 23.63\\
    IDN+Filtering & 24.13 & 23.74 & 24.24 \\
    IDN+Probe     & {\bf 24.34} & {\bf 24.03} & {\bf 24.43} \\
    \hline
    \end{tabular}
    }
\end{center}
\caption{Results of HOI detection (using detected object boxes).}
\label{table:res-hoi} 
\end{table}

\begin{table}[ht]
\begin{center}
    \resizebox{\linewidth}{!}{
    \begin{tabular}{l|l|cc}
    \hline 
    Model & Test Inference & $\alpha$ Amp. & $\beta$ Amp. \\
    \hline
     OCRN                               & $\text{argmax} _y P(y | x)$                 & \textbf{0.127} &  \textbf{0.112} \\
    \hline
     DM-V + Joint ND-way Softmax        & $\text{argmax} _y \max _d P _{te}(y,d | x)$ &     0.151     &    0.158       \\
     DM-V + Joint ND-way Softmax        & $\text{argmax} _y \sum _d P _{te}(y,d | x)$ &     0.148     &    0.154       \\
     DM-V + N-way classifier per domain & $\text{argmax} _y P _{te}(y | d^*,x)$       &     0.135     & \textbf{0.112} \\
     DM-V + N-way classifier per domain & $\text{argmax} _y \sum _d s (y,d,x)$        &     0.147     &    0.145       \\
    \hline
    \end{tabular}
    }
\end{center}
\caption{Comparison with debiasing models.}
\label{table:debias} 
\end{table}

\section{Comparison on Imbalance Learning}
\label{app:sec:debias}

\subsection{Debiasing Learning}

\begin{figure*}[ht]
    \begin{minipage}{0.45\textwidth}
        \centering
        \includegraphics[width=\linewidth]{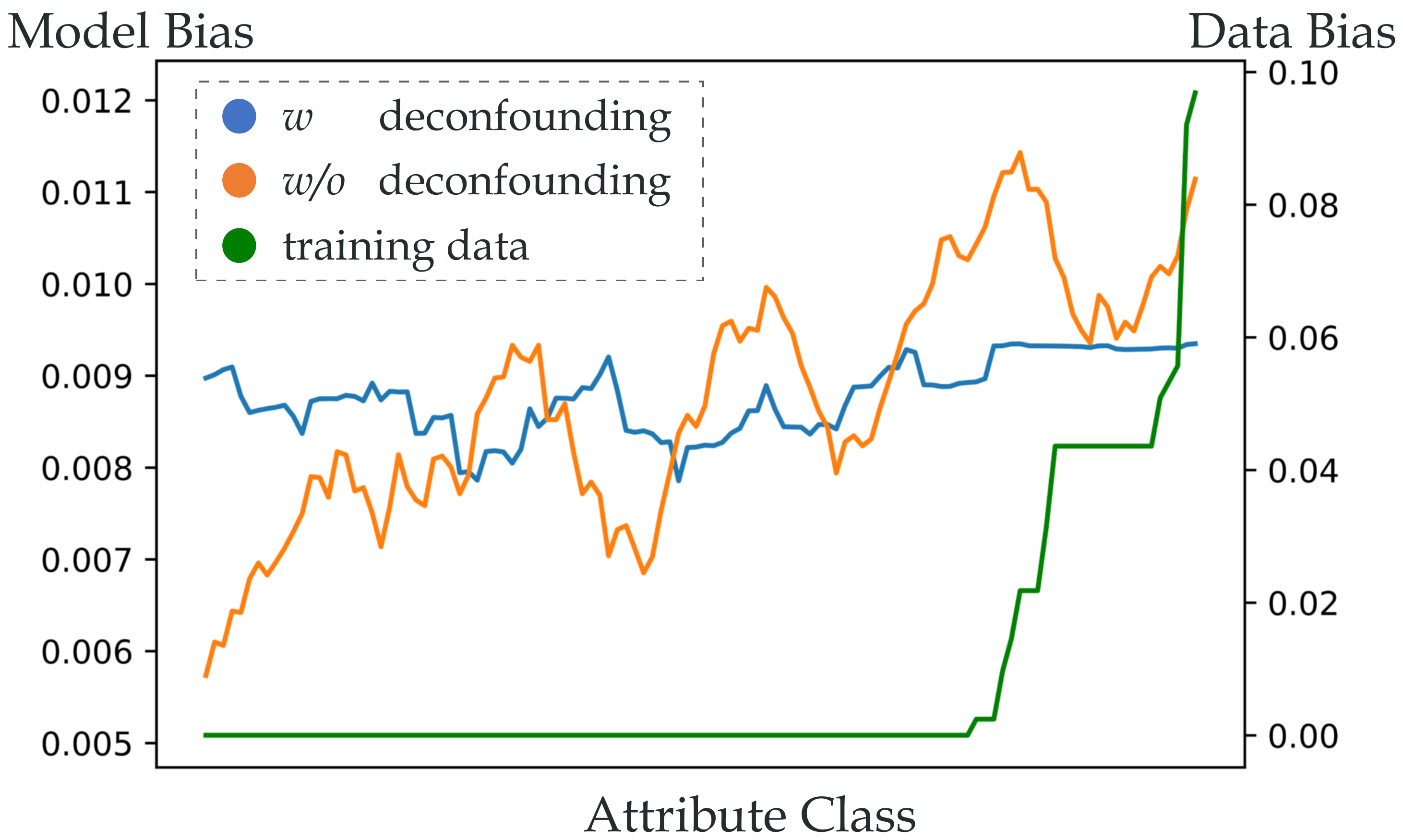}
        \vspace{-12px}
        \caption{Attribute bias (w/ and w/o deconfounding) for category \texttt{frying pan}.
        } 
        \label{fig:attr-bias-comparison}
    \end{minipage}
    \hfill
    \begin{minipage}{0.45\textwidth}
        \centering
        \includegraphics[width=\linewidth]{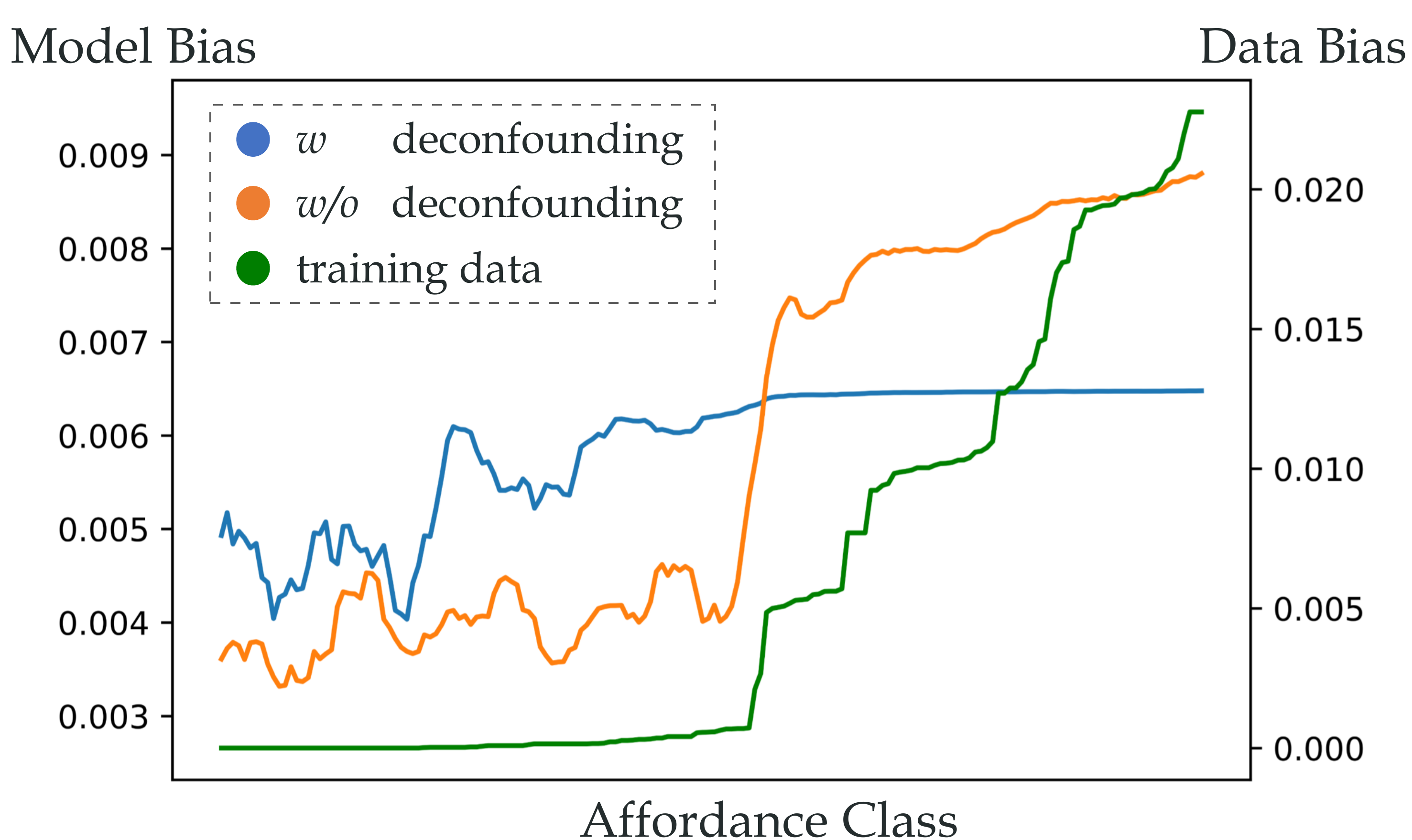}
        \vspace{-12px}
        \caption{Affordance bias (w/ and w/o deconfounding) for category \texttt{giraffe}.
        } 
        \label{fig:aff-bias-comparison}
    \end{minipage}
\end{figure*}

The motivation of the OCRN is to follow the prior knowledge of the three levels of objects with a deep learning-based causal graph model, to pursue the object understanding beyond the common direct mapping from pixels to labels, and to avoid the bias estimation such as in the \textbf{Simpson's paradox}~\cite{pearl2016causal}. Thus, we use intervention to deconfound the confounder \textit{category} and exclusive the possible spurious bias and correlation imported bias from imbalanced object categories. Overall, we propose our OCRN in a causal inference perspective instead of the pure classification viewpoint, which also suits our causal graphical model well. Similar cases are also proposed in recent works like~\cite{vcrcnn,tang2020long,tang2020unbiased}.
Moreover, to better compare our method with the common debiasing methods, we further conduct the experiments as follows.

We regard $\alpha,\beta$ recognition as multiple independent binary classification tasks and implement some methods introduced in \cite{wang2020towards} on our strong baseline DM-V to reduce bias from object categories.
We use \textbf{mean bias amplification} (\textbf{Amp}) in \cite{zhao2017men} as bias evaluation metric: small Amp means model suffers less from data category bias. 
The test results are shown in Tab.~\ref{table:debias}. The proposed OCRN has comparable or smaller bias amplification than the variants of DM-V since our model follows the causal graph and exploits the tools of causal inference, while most methods for category bias are from the view of classification.

To verify the debiasing of OCRN, we compare the model bias of OCRN \textbf{w/ or w/o deconfounding}. The bias of category $O$ upon an attribute $\alpha$ is measured following \cite{zhao2017men}, by 
$b(O,\alpha)=c(O,\alpha)/\sum_{\alpha'}c(O,\alpha')$. When measuring \textbf{data} bias, $c(O,\alpha)$ is the number of co-occurrence of $O$ and $\alpha$ in OCL, and when it comes to \textbf{model} bias, $c(O,\alpha)$ is the sum of probabilities that $O$ are predicted positive with $\alpha$. 
The bias of $\beta$ is measured in the same manner. 
Fig.~\ref{fig:attr-bias-comparison} and \ref{fig:aff-bias-comparison} show some examples of the biases of training data and models, indicating that OCRN deconfounding effectively prevents the model from bias toward the train set.

\subsection{Long-tailed Learning}

Besides the debiasing learning techniques, we also apply longtailed learning methods on our baseline method DM-$\alpha\rightarrow\beta$ for comparison, including class-balanced sampling~\cite{buda2018systematic} and focal loss~\cite{lin2017focal}. The models with additional re-balancing modules suffer from minor accuracy degradation, mainly for OCL is long-tailed on object class while we infer $\alpha,~\beta$, so the gap minimizes the effect of long-tailed learning.

\begin{table}[ht]
\begin{center}
    \resizebox{\linewidth}{!}{
    \begin{tabular}{l|cc}
    \hline 
    Method & $\alpha$ & $\beta$ \\
    \hline
 DM-$\alpha\rightarrow\beta$                                                 & 28.7\% & 52.6\% \\
    \hline
 DM-$\alpha\rightarrow\beta$+Class balance sampling~\cite{buda2018systematic}& 27.3\% & 52.1\%  \\
 DM-$\alpha\rightarrow\beta$+Focal loss~\cite{lin2017focal}                  & 27.6\% & 51.2\%  \\
    \hline
    \end{tabular}
    }
\end{center}
\caption{Comparison with debiasing models.}
\label{table:debias} 
\end{table}

\section{Discussion about States}
\label{app:sec:state}
We did not use object states in our model because there is also a \textbf{compositional zero-shot problem} and object-state pairs, \textit{i.e.}, there can be \textbf{unseen} states in real-world data. Differently, affordances are more general. The models explicitly incorporating object states will fail to generalize to these zero-shot states and it adds to the object category bias. 
In experiments, the state supervision during training would indeed \textit{slightly improve} the affordance recognition performance, since instances in the \textbf{same state} lie in a tight cluster in affordance label space. But this will hurt the ITE performance greatly.

\section{Discussion about Causality and Causal Graph}
\label{app:sec:causal-graph}
Annotating causality in the real world is extremely difficult. In data annotation, we have met numerous ambiguities and difficulties to confirm the ``right'' causal relations. To address these challenges, we follow the following principles:
(1) Firstly, we only emphasize \textbf{clear} and \textbf{strong} causal relations via crowdsourcing, but omit the vague ones. 
(2) Second, we take an object \textbf{affordance-centric} viewpoint to look at the possible causal relations.
(3) We would rather \textbf{discard} than condone the controversial situation in the annotation.
(4) We only focus on the simple relations between \textbf{one} attribute and \textbf{one} affordance, instead of the very complex compositions of multiple attributes and affordances which are almost impossible to annotate.
Therefore, we finally find that we can label \textbf{a very small percentage} of all arcs with the whole causal graph consisting of so many nodes (category, attributes, affordances, contexts, etc.) while keeping the quality.

Our causal graph follows the human priors from our experts and crowdsourcing annotators. Some previous works also follow this before designing the method, such as~\cite{yuke}. 
From the viewpoint of causal discovery~\cite{pearl2016causal,tang2020long,tang2020unbiased,vcrcnn}, the above arcs (\textit{e.g.}, the inverted arc from attribute to category in the causal graph directed acyclic graph, DAG) are indeed possible. However, here, we mainly study the object concept learning problem, especially attribute and affordance learning for intelligent robots and embodied AI. Thus, from the perspective of affordance learning, we think the arcs from category to attribute and affordance are more vital and meaningful to us. 

Causality can also be confused with \textbf{enabling condition}.
In OCL, the affordance of an object indicates what human \textit{can do} to/with it. 
In this case, ``fresh'' causes ``\textbf{eat-able}'' (\textbf{rather than causes ``eat'' action}).
As causality is discussed in the view of \textbf{embodied agents}, this rule can hold.
In modern causal inference models like structured causal models (SCM), causality and enabling conditions are not strictly distinguished.
As stated by Cheng \etal~\cite{cheng1991causes}, 
causes and enabling conditions hold the same logical relation to the effect in those terms and the methods that explain their distinction come from the subject judgment of humans. 
The distinction can be explained based on the \textit{normality} of potential factors, or considering the existing assumption of the inquirer. They proposed an approach by measuring the covariation between potential factors to the effect over a set of questions.
So in SCM, both will be represented as nodes and involved in causal mechanisms.
OCL follows the ``open'' setting: affordance is a subjective property of the object, so all reasons given by humans/robots (including enabling conditions) are regarded as causal factors.

\section{Detailed Lists}
\label{sec:classes-list}
The detailed object categories, attributes, and affordances are listed on our website: \url{https://mvig-rhos.com/ocl}.

\end{document}